\documentclass[10pt,twocolumn,letterpaper]{article}

\usepackage{titling}
\pretitle{\begin{center}\huge}

\usepackage{iccv}
\usepackage{amsmath}
\usepackage{amssymb}
\usepackage{booktabs}
\usepackage{animate}
\usepackage{enumitem}
\usepackage{multirow}
\usepackage{diagbox}

\usepackage{enumitem}
\usepackage{tabularx}
\usepackage{microtype}
\usepackage[accsupp]{axessibility}

\usepackage{times}
\usepackage{epsfig}
\usepackage{graphicx}
\usepackage{times}
\usepackage{epsfig}
\usepackage{algorithm, algorithmic}
\usepackage{xcolor}

\usepackage{diagbox}
\usepackage{float}
\usepackage{afterpage}
\usepackage{bm}
\usepackage{times}
\usepackage{epsfig}
\usepackage{graphicx}
\usepackage{subfig}
\usepackage{tabu}
\usepackage{multirow}
\usepackage{color}
\usepackage{tablefootnote}
\usepackage{adjustbox}
\usepackage{bbding}
\usepackage{etoolbox}
\usepackage{booktabs}
\usepackage{soul}
\usepackage{subfiles}
\usepackage{xcolor,colortbl}
\usepackage[pagebackref=true,breaklinks=true,letterpaper=true,colorlinks,bookmarks=false]{hyperref}

\def\etal{\textit{et al}.}
\def\ie{\textit{i.e.}}
\def\eg{\textit{e.g.}}

\usepackage[capitalize]{cleveref}
\crefname{section}{Sec.}{Secs.}
\Crefname{section}{Section}{Sections}
\Crefname{table}{Table}{Tables}
\crefname{table}{Tab.}{Tabs.}

\newcommand{\paraheading}[1]{\vspace{-1.1em}\paragraph{#1.}}
\newcommand{\topparaheading}[1]{\vspace{-0.2em}\paragraph{#1.}}

\definecolor{Gray}{gray}{0.5}
\definecolor{LightCyan}{rgb}{0.88,1,1}

\newcolumntype{a}{>{\columncolor{Gray}}c}
\newcolumntype{b}{>{\columncolor{white}}c}

\def\etal{\textit{et al}.}
\def\ie{\textit{i.e.}}
\def\eg{\textit{e.g.}}

\iccvfinalcopy 

\def\iccvPaperID{****} 

\ificcvfinal\fi

\begin{document}


\title{RIGID: Recurrent GAN Inversion and Editing of Real Face Videos}

\author{Yangyang Xu$^1$,
       Shengfeng He$^2$,
       Kwan-Yee K. Wong$^1$,
       and Ping Luo$^{1,3}$\thanks{Corresponding author: pingluo@hku.hk.}
       \\
       {$^1$Deapartment of Computer Science, The University of Hong Kong} \\
       {$^2$School of Computing and Information Systems, Singapore Management University}\\
       {$^3$Shanghai AI Laboratory}\\
      }

\teaser{
    \centering

    \subfloat[\footnotesize{Original}]
    {\begin{minipage}{.16\linewidth}
    \centering
        \animategraphics[autoplay,loop,width=\textwidth]{24}{figure/teaser_bb/bb_ori/}{0100}{0124}\\
        \animategraphics[autoplay,loop,width=\textwidth]{24}{figure/teaser_bb/bb_ori/}{0100}{0124}
    \end{minipage}}
    \hspace{-1.5mm}
    \subfloat[\footnotesize{IGCI / 120,000s}]{\label{fig_igci}
    {\begin{minipage}{.16\linewidth}
    \centering
        \animategraphics[autoplay,loop,width=\textwidth]{24}{figure/teaser_bb/bb_rec_xu/}{0100}{0124}\\
        \animategraphics[autoplay,loop,width=\textwidth]{24}{figure/teaser_bb/bb_Chubby_xu/}{0100}{0124}
    \end{minipage}}}
    \hspace{-1.5mm}
    \subfloat[\footnotesize{Latent-T. / 49.2s}]{\label{fig_lt}
    {\begin{minipage}{.16\linewidth}
    \centering
        \animategraphics[autoplay,loop,width=\textwidth]{24}{figure/teaser_bb/bb_rec_lt/}{0100}{0124}\\
        \animategraphics[autoplay,loop,width=\textwidth]{24}{figure/teaser_bb/bb_Chubby_lt/}{0100}{0124}
    \end{minipage}}}
    \hspace{-1.5mm}
    \subfloat[\footnotesize{STIT / 1,600s}]{\label{fig_stit}
    {\begin{minipage}{.16\linewidth}
    \centering
        \animategraphics[autoplay,loop,width=\textwidth]{24}{figure/teaser_bb/bb_rec_stit/}{0100}{0124}\\
        \animategraphics[autoplay,loop,width=\textwidth]{24}{figure/teaser_bb/bb_Chubby_stit/}{0100}{0124}
    \end{minipage}}}
    \hspace{-1.5mm}
    \subfloat[\footnotesize{TCSVE / 3,400s}]{\label{fig_veg}
    {\begin{minipage}{.16\linewidth}
    \centering
        \animategraphics[autoplay,loop,width=\textwidth]{24}{figure/teaser_bb/bb_rec_stit/}{0100}{0124}\\
        \animategraphics[autoplay,loop,width=\textwidth]{24}{figure/teaser_bb/bb_Chubby_tcsve/}{0100}{0124}
    \end{minipage}}}
    \hspace{-1.5mm}
    \subfloat[\footnotesize{RIGID / 54.5s}]{\label{fig_rigid}
    {\begin{minipage}{.16\linewidth}
    \centering
        \animategraphics[autoplay,loop,width=\textwidth]{24}{figure/teaser_bb/bb_rec_ours/}{0100}{0124}\\
        \animategraphics[autoplay,loop,width=\textwidth]{24}{figure/teaser_bb/bb_Chubby_ours/}{0100}{0124}
    \end{minipage}}}
\rotatebox[origin=c]{270}{\hspace{-2.5mm} \small{\texttt{Inversion}} \hspace{12.5mm} \small{\texttt{+Chubby}}}\\
\caption{Comparisons of video inversion and editing with existing methods. Number in each column denotes the average editing time over 100 frames. Our RIGID achieves temporal coherent inversion and editing performances with much less time cost. This figure contains \emph{animated videos}, which are best viewed using Adobe Acrobat.}
\label{fig:teaser}
}


\maketitle

\begin{abstract}
GAN inversion is indispensable for applying the powerful editability of GAN to real images. However, existing methods invert video frames individually often leading to undesired inconsistent results over time. In this paper, we propose a unified recurrent framework, named \textbf{R}ecurrent v\textbf{I}deo \textbf{G}AN \textbf{I}nversion and e\textbf{D}iting (RIGID), to explicitly and simultaneously enforce temporally coherent GAN inversion and facial editing of real videos. Our approach models the temporal relations between current and previous frames from three aspects. To enable a faithful real video reconstruction, we first maximize the inversion fidelity and consistency by learning a temporal compensated latent code. Second, we observe incoherent noises lie in the high-frequency domain that can be disentangled from the latent space. Third, to remove the inconsistency after attribute manipulation, we propose an \textit{in-between frame composition constraint} such that the arbitrary frame must be a direct composite of its neighboring frames. Our unified framework learns the inherent coherence between input frames in an end-to-end manner, and therefore it is agnostic to a specific attribute and can be applied to arbitrary editing of the same video without re-training. Extensive experiments demonstrate that RIGID outperforms state-of-the-art methods qualitatively and quantitatively in both inversion and editing tasks. The deliverables can be found in \url{https://cnnlstm.github.io/RIGID}.
\end{abstract}

\section{Introduction}

Generative adversarial networks (GANs) have demonstrated powerful generative ability in synthesizing high-quality faces from a latent code~\cite{karras2017progressive,karras2019style,karras2020analyzing,zheng2023my}. It is evidenced that the latent space of a well-trained GAN is semantically organized, and shifting the latent code along with a specific direction results in the manipulation of a corresponding attribute~\cite{shen2020interfacegan,shen2021closed,harkonen2020ganspace,Patashnik_2021_ICCV,yuksel2021latentclr,voynov2020unsupervised}. Hence, many works migrate this power to real face processing by inverting a real face image to a latent code~\cite{zhou2022pro,li2023parsing,Xu2021ICCV,zhong2022faithful,xu2022self,xu2022high}. Although this two-combo strategy becomes a standard for editing high-resolution images, applying it to real videos has less been explored. A naive inversion and editing for each frame can undoubtedly produce incoherence in the resulted video.

Different from processing images, maintaining temporal coherence is the core issue for video editing~\cite{logacheva2020deeplandscape,tian2021a,stylegan_v}. Specifically, both the GAN inversion and attribute manipulation may introduce discontinuity across frames. IGCI~\cite{Xu2021ICCV} proposes the first attempt to invert consecutive images simultaneously. They leverage the continuity of the inputs to optimize both the reconstruction fidelity and editability of the outputs, but they fail to consider the temporal correlation between results (see flickering in Fig.~\ref{fig_igci}). Recent work STIT~\cite{tzaban2022stitch} implicitly recovers the original temporal correlations by the faithful inversion of each frame. It fine-tunes an individual generator for every input video such that the generator can capture all the reconstruction details, and TCSVE~\cite{xu2022temporally} extends this idea by proposing a temporal consistency loss that applies on the edited videos. Although they works well for most cases, they are video- and attribute-specific (needs to retrain the model for a new video or a new target attribute), and thus suffers from the expensive training cost and poor generalization ability. 

In this paper, we aim to design a unified approach that learns the temporal correlations between successive frames for both inversion and editing, and it can be generalized to other target attributes without re-training. To this end, we propose a \textbf{R}ecurrent v\textbf{I}deo \textbf{G}AN \textbf{I}nversion and e\textbf{D}iting (RIGID) framework, which evolves and enables the image-based StyleGAN~\cite{karras2019style,karras2020analyzing} generator to output temporally coherent frames. The coherence is realized in both inversion and editing tasks. Given the current and previous frames, we formulate the inversion as the combination of an image-based inverted code and a temporal compensated code, while the latter amends the code with inter-frame similarity for an accurate and consistent inversion. On the other hand, we observe that the main sources of temporal incoherence, like ``flickering'', belong to high-frequency artifacts. This motivates us to disentangle the main video content from high-frequency artifacts in the latent space, and thus the ``incoherence'' can be shared with all the other frames. To build the temporal correlations after attribute manipulation, we propose a self-supervised ``\textit{in-between frame composition constraint}'' that applies to consecutive edited frames. It enforces any intermediate frame that can be composed by the warping results of their neighbors, which guarantees the smoothness of generated videos. RIGID is trained on the video episodes with several tailored losses. During the inference, it inverts video frames sequentially and therefore can handle videos with arbitrary lengths and support live stream editing. More importantly, once our model is trained, it is attribute-agnostic that can be reused for arbitrary attribute manipulations without re-training. As shown in Fig.~\ref{fig_rigid}, RIGID achieves temporal coherent inversion and editing with far less inference time (compared to those scene- and attribute-specific methods like STIT). Extensive experiments demonstrate the superiority over state-of-the-art methods in terms of quantitative and qualitative evaluations.

In summary, our contributions are three-fold:
\begin{itemize}
    \item We propose a recurrent video GAN inversion framework that unifies video inversion and editing. It learns the temporal correlation of generated videos explicitly.
    \item We model temporal coherence from both inversion and editing ends. For inversion, we discover the temporal compensated code and disentangle high-frequency artifacts in the latent space. For editing, we present a novel ``\textit{in-between frame composition constraint}'' to confine a continuous video transformation.
    \item We achieve attribute-agnostic editing that can vary editing attributes on the fly, avoiding expensive re-training of the model. Extensive experiments demonstrate the effectiveness of our method over state-of-the-arts.
\end{itemize}

\begin{figure*}
\centering
\includegraphics[width=\textwidth]{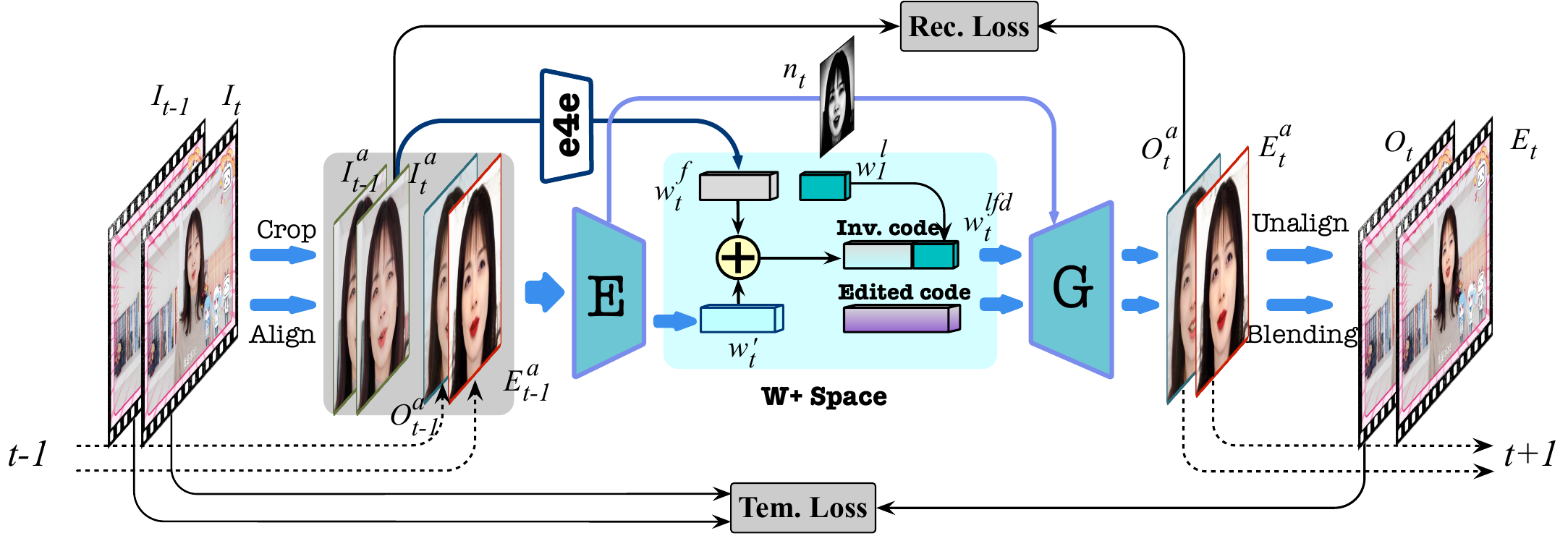}
\caption{Overview of our RIGID. We first align the faces on two neighbor frames ($I_{t-1}$ and $I_t$). Then we concatenate them with the inverted face $O_{t-1}$ and edited face $E_{t-1}$ in the previous step as the inputs of the recurrent encoder. The encoder learns the temporal compensated code $w_t^{\prime}$ and spatial noise map $n_t$, as a complement to the initial latent code $w_t$ (acquired by the ``e4e'' encoder). In addition, we share the latter part of latent codes across all frames using $w^l_1$ for eliminating the high-frequency temporal flickering. Both the final inverted latent code $w_t^{lfd}$ and the edited one are fed to the generator, producing the inverted and edited faces $O^a_{t}$ and $E^a_{t}$. The generated faces are unaligned and blended with the original frames. Dotted lines denote recurrent inputs from the last time step or outputs to the next step. In addition, a novel \textit{in-between frame composition constraint} is proposed that learns the temporal correlation during editing (details can be found in Fig.~\ref{fig:ifsc}).}
\label{fig:framework}
\end{figure*}

\section{Related Works}

\topparaheading{GAN Inversion} GAN inversion aims at inverting real images into the latent space of pre-trained generators for reconstruct and edit the real images~\cite{abdal2019image2stylegan,Xu2021ICCV,richardson2021encoding,wang2022HFGI,xia2021gan,tov2021designing,song2022editing}. Early works optimize the latent code directly for a specific image with expensive computational cost~\cite{creswell2018inverting,abdal2019image2stylegan}. Another class trains a general encoder that maps the real image to the latent code directly~\cite{richardson2021encoding,tov2021designing,wang2022HFGI}. Particularly, pSp~\cite{richardson2021encoding} proposes an encoder with pyramid architecture that inverts the real images into the~$\mathcal{W+}$~latent space of StyleGAN. Based on the same architecture, e4e~\cite{tov2021designing} analyzes the trade-offs between reconstruction and editability in StyleGAN's inversion. HFGI~\cite{wang2022HFGI} introduces the distortion map for improving the fidelity reconstitution. Existing works mainly focus on the image-based GAN inversion, inverting the real videos into GANs has not been well studied.

\paraheading{StyleGAN-based Video Generation and Editing} State-of-the-art video GANs still cannot generate high quality results as their image counterparts~\cite{tian2021a,hyun2021self,NEURIPS2018_d86ea612,tzaban2022stitch}. Opposite to design a video generator directly, many works use the pre-trained image generator (\eg, StyleGAN) for synthesizing high-quality videos~\cite{logacheva2020deeplandscape,tian2021a,yin2022styleheat,stylegan_v}. MoCoGAN-HD~\cite{tian2021a} decomposes the motion and content in videos and generates motion trajectory in StyleGAN's latent space. StyleHEAT~\cite{yin2022styleheat} shares the same decomposition idea, it exploits the flow field as the motion descriptor for talking face generation. Recently, StyleGAN-V~\cite{stylegan_v} injects the continuous motion representations into the StyleGAN. In this paper, we aim at inverting a real video into a pre-trained StyleGAN for temporal coherence reconstruction and editing. Few works concentrate on this task. Latent-Transformer~\cite{yao2021latent} presents a pipeline for facial video editing by inverting each frame individually, resulting in the temporal inconsistency in the edited videos. IGCI~\cite{Xu2021ICCV} introduces the consecutive frames into GAN inversion for improving the reconstruction quality and editability. STIT~\cite{tzaban2022stitch} edits the facial videos using StyleGAN2~\cite{karras2020analyzing} by fine-tuning the generator. {TCSVE~\cite{xu2022temporally} also optimizes generators with a temporal consistency loss that applies on the edited videos.} Above two optimization-based work need to re-optimize the generator for a new video or edit, which is time-consuming. In this paper, we propose a recurrent framework that learns the temporal correlations both in inverted and edited videos. Once it is properly trained, it supports various semantic editing methods with low computational costs.

\section{Approach}

\subsection{Formulation}


Given a real face video $\{I_t|t=1,...T\}$, our goal is to invert it to the latent space of pre-trained StyleGAN $G$ to obtain a set of latent codes, and output the inverted video $\{O_t|t=1,...T\}$, where $O_t$ is obtained by feeding the latent code to $G$. Meanwhile, we can also obtain an edited video $\{E_t|t=1,...T\}$ by manipulating the latent codes. The key is that both inverted and edited video should be temporal coherent.

To address that, we propose a recurrent video GAN inversion and editing framework to explicitly and simultaneously enforce temporally coherent GAN inversion and facial editing of real videos. For video inversion, as the original video of inborn temporal coherence, the best way to maintain inversion consistency is a faithful reconstruction. In addition, we propose a ``\textit{latent frequency disentanglement}'' strategy for eliminating the high-frequency temporal flickering in the latent space. We also propose an \textit{in-between frame composition constraint} that builds the temporal correlations of edited video. The overview of our RIGID can be seen in Fig.~\ref{fig:framework}.

\subsection{Coherent Video Inversion}

\subsubsection{Temporal Compensated Inversion}

Given a real video, we can deliver its temporal correlation to the generated video by a faithful reconstruction. Before inversion, face alignment is necessary since StyleGAN cannot handle the entire frame. After the alignment on each frame, we can obtain a set of aligned faces $\{I^a_t|t=1,...T\}$. We first encode them to the $\mathcal{W+}$ space using image-based inversion method (e4e~\cite{tov2021designing} in this paper), and obtain the initial latent codes $\{w_t|t=1,...T\},w_t\in \mathcal{W+}$. However, directly using the initial latent codes cannot reconstruct original faces accurately due to the missing of temporal context. We use a recurrent encoder that learns a temporal compensated code as a complement to the initialized one. Moreover, recovering a high-fidelity face solely in StyleGAN's $\mathcal{W+}$ latent space is too difficult due to the lack of spatial information~\cite{kim2021exploiting,wang2022HFGI}. Inspired by~\cite{abdal2020image2styleganpp}, the encoder also learns a noise map in StyleGAN's $\mathcal{N}$ space for injecting the spatial information.

Our goal is to unify inversion and editing in the same framework. As a result, both the inverted and edited faces from previous and current time steps are fed to the encoder to generate the temporal compensated code and the noise map. Here a ConvLSTM layer~\cite{shi2015convolutional} is integrated into the encoder for modeling the spatial-temporal correlations. Specifically, at time step $t$, we concatenate the aligned faces $I^a_{t-1}$, $I^a_{t}$, the last inverted face $O^a_{t-1}$, and last edited face $E^a_{t-1}$ as inputs. It outputs a temporal compensated code $w^{\prime}_t$ and spatial noise map $n_t$, we add $w^{\prime}_t$ with the initialized code $w_t$. Then both added code and noise map $n_t$ are sent to the generator for inversion, that is:
\begin{gather}
\{w^{\prime}_t,n_t\}=E(\texttt{cat}(I^a_{t-1},I^a_{t},O^a_{t-1},E^a_{t-1})),\\
O^a_t = G(w_t+w^{\prime}_t,n_t),
\label{eq2}
\end{gather}
where $E$ denotes the recurrent encoder and $\texttt{cat}(\cdot,\cdot,\cdot,\cdot)$ denotes the concatenation operator. 

\subsubsection{Latent Frequency Disentanglement}
The fidelity inversion delivers the temporal relations from the original video to the inverted one. However, frames are inverted one by one and may lead to subtle inconsistency, \ie, the unique high-frequency information in a single frame will be accumulated in a video and leads to temporal flickering. We notice that high-frequency temporal flickering mainly exists in the appearance of an image. Recent works evidenced that $\mathcal{W+}$ space is highly disentangled, and the appearance of the image is synthesized at the higher layer of StyleGAN, which is controlled by the latter part of the $w$ latent code~\cite{karras2020analyzing,richardson2021encoding}. Hence we propose a ``\textit{latent frequency disentanglement}'' strategy that eliminates the temporal flickering by sharing the latter part of $w$ latent code across all the frames. 


Specifically, we first decompose a latent code into the former part and the latter part, then we reuse the latter part (corresponding to high layers of StyleGAN) of the first frame in all the following frames that unify the high frequency information, that is:
\begin{gather}
w_t^{lfd} = \texttt{cat}((w_t+w^{\prime}_t)^f,w_1^l),
\label{eq8}
\end{gather}
where $w_t^{lfd}$ is the latent code after latent frequency disentanglement, $(w_t+w^{\prime}_t)^f$ is the former part of latent code $w_t+w^{\prime}_t$, and $w_1^l$ denotes the latter part of the first latent code. Now, we can get the final latent codes $\{c_t=\{w_t^{lfd},n_t\}|t=1,...T\}$ for reconstructing a temporal coherent video. That is, we replace Eq.~\ref{eq2} with following equation:
\begin{gather}
O^a_t = G(w_t^{lfd},n_t).
\label{eq9}
\end{gather}

The face can be edited by manipulating the latent code $w_t^{lfd}$:
\begin{gather}
E^a_t = G((w_t^{lfd}+\overrightarrow{\bold{n}}),n_t),
\label{eq3}
\end{gather}
where $\overrightarrow{\bold{n}}$ denotes the semantic direction acquired by arbitrary semantic editing techniques~\cite{shen2020interfacegan,shen2021closed,harkonen2020ganspace,yao2021latent,yang2021discovering}. It is noticed that the $n_t$ is also determined by edited face $E^a_{t-1}$, and it can be well cooperated with both inverted and edited code.

\subsection{Coherent Video Editing}

\subsubsection{Post Processing}
The generated faces are naturally aligned that lost the temporal coherence. We need to unaligned generated faces and blend them with the original frame~$I_t$. Here we follow~\cite{yao2021latent,tzaban2022stitch} that blends the face region only. In particular, we first segment the face region on an original frame by a pre-trained face parsing model~\cite{yu2018bisenet} to get the inner face mask $M_{I_t}$, then we blend the inverted face $O^a_t$ with the original frame according to mask $M_{I_t}$, which can be presented as:
\begin{equation}
O_t=\texttt{B}(\texttt{UA}(O^a_t),I_t,M_{I_t}),
\label{eq4}
\end{equation}
where $\texttt{B}$ and $\texttt{UA}$ denote the blending and unalignment respectively.

For the edited face $E^a_t$, its face region may be modified after editing, and the face mask of the original frame $M_{I_t}$ cannot well fit with the edited faces. Besides, directly using the face mask of the edited face also suffers a similar problem. Hence we use the union of two masks as the blending mask, that is:
\begin{equation}
E_t=\texttt{B}(\texttt{UA}(E^a_t),I_t,M_{E_t}\cup M_{I_t}).
\label{eq5}
\end{equation}

\subsubsection{\textit{In-between Frame Composition Constraint}}

After post processing, we can obtain the inverted and edited videos. Compared with the inverted videos, temporal correlation in an edited video is more difficult to learn, since there is no GT edited videos for supervision. We propose a self-supervised \textit{in-between frame composition constraint} that models the temporal correlation in an edited video.

\begin{figure}
\centering
\includegraphics[width=0.5\textwidth]{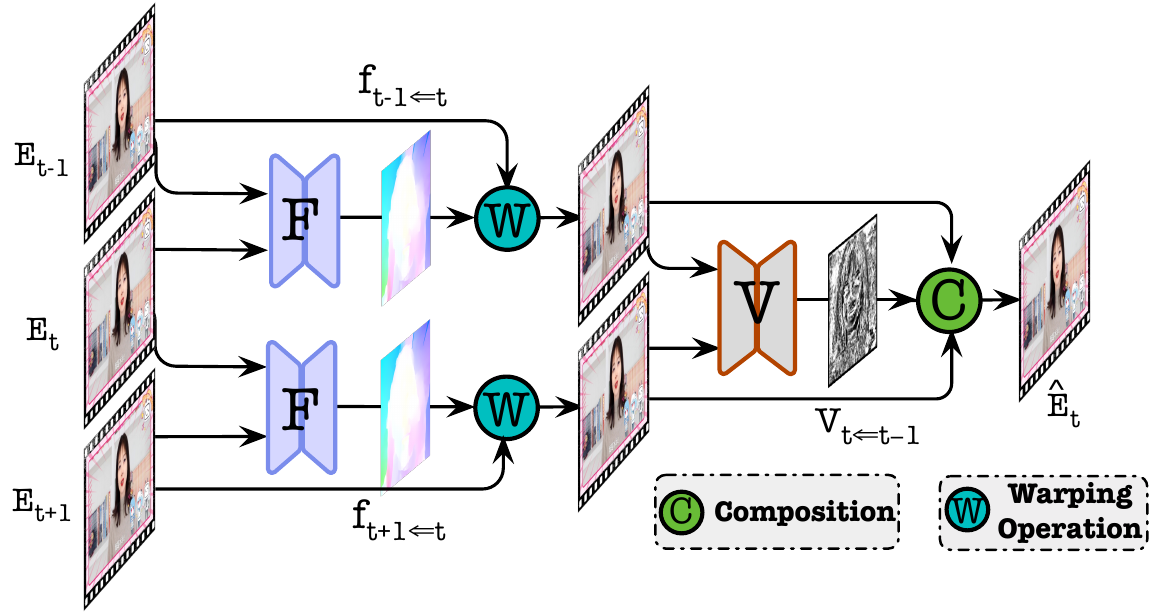}
\caption{The in-between frame can be composed by flow-based warping results of its neighboring frames according to a visibility map $V_{t \Leftarrow {t-1}}$.}
\label{fig:ifsc}
\end{figure}

Generally, for a triplet of consecutive frames $\{E_{t-1},E_{t},E_{t+1}\}$, the intermediate frame $E_{t}$ can be composed by flow-based warping results of its neighbor frames~\cite{jiang2018super,reda2019unsupervised,xu2019quadratic}. Specifically, as shown in Fig.~\ref{fig:ifsc}, let $f_{t \Rightarrow {t-1}}$  denotes the optical flow from $E_t$ to $E_{t-1}$ and $f_{t \Rightarrow {t+1}}$ is the flow from $E_t$ to $E_{t+1}$, then frames $E_{t-1}$ and $E_{t+1}$ are warped using different flow. And the intermediate frame can be composed using two warped results according to a visibility map, that is:
\begin{equation}
\begin{aligned}
\hat{E}_{t}=&V_{t \Leftarrow {t-1}} \odot \texttt{W}(E_{t-1}, f_{t \Rightarrow {t-1}}) +\\ &(1-V_{t \Leftarrow {t-1}}) \odot \texttt{W} (E_{t+1}, f_{t \Rightarrow {t+1}}),
\end{aligned}
\label{eq6}
\end{equation}
where $\hat{E}_{t}$ is the composed intermediate frame, $V_{t \Leftarrow {t-1}}$ is the visibility map from frame $E_{t}$ to $E_{t-1}$, it is a one channel mask with the same resolution with aligned face frames. It denotes whether a pixel remains visible when moving from frame $t-1$ to $t$ (0 is fully occluded and 1 is fully visible). $\texttt{W}$ is the warping operator, and $\odot$ denotes element-wise multiplication. This equation models the temporal correlations among three consecutive frames.

For the edited videos, both $f_{t \Rightarrow {t-1}}$ and $f_{t \Rightarrow {t+1}}$ are available, the remaining challenge is to estimate the visibility mask $V_{t \Leftarrow {t-1}}$. Since there is no GT for edited videos, we turn to train a visible net $V$ for estimating the mask. Particularly, we composite the in-between frame $\hat{I}_t$ on the real videos using Eq.~\ref{eq6}. Then we align the $l_1$ distance between $\hat{I}_t$ and $I_t$ for training visible net $V$. After training, we fix the $V$ and adopt it to the edited frames for estimating the visibility mask and composing the in-between frame $\hat{E}_{t}$. We minimize the distance between $\hat{E}_{t}$ and ${E}_{t}$ to form \textit{in-between frame composition constraint}:
\begin{equation}
\mathcal{L}_{ibfcc}=\sum_{t=2}^{T-1}||E_t-\hat{E}_{t}||_1.
\label{eq7}
\end{equation}
This constraint enforces the edited video as smooth as real video, which guarantees the temporal coherence effectively. Note that we train the encoder on the video episodes. Particularly, we first collect the outputs $\{E_{0},...,E_{T}\}$ in an episode via forward propagation, then we apply this constraint to train the encoder in backward propagation (we use differentiable unalignment).


\subsection{Loss Functions}

Our RIGID is trained under several tailored losses. Besides the \textit{in-between frame composition constraint} $\mathcal{L}_{ifsc}$ applied on the edited videos, it also includes reconstruction and temporal consistency losses on the inverted videos.

\textbf{Reconstruction Loss.}
We first introduce the reconstruction loss $\mathcal{L}_{rec}$ on the inverted faces. Following~\cite{richardson2021encoding,wang2022HFGI}, it includes a pixel-wise $\mathcal{L}_{2} $ loss for minimizing the reconstruction error, and LPIPS loss for a better image quality preservation~\cite{zhang2018unreasonable}, which can be represented as:
\begin{gather}
\label{eqx}
\mathcal{L}_{2} = \sum_{t=1}^T||I^a_t-O^a_t||_{2},\\
\mathcal{L}_{lpips} = \sum_{t=1}^T||P(I^a_t)-P(O^a_t)||_{2},\\
\mathcal{L}_{rec} = \mathcal{L}_{2}+\alpha\mathcal{L}_{lpips},
\label{eq7}
\end{gather}
where $T$ is the number of frame in each training episode, $P(\cdot)$ denotes the perceptual feature extractor, and $\alpha$ is the balance weight. Following~\cite{richardson2021encoding}, we set $\alpha=0.8$.

\textbf{Temporal Consistency Loss.}
We also introduce a warping-based temporal consistency loss $\mathcal{L}_{tc}$ for preserving the temporal consistency of inverted videos. In particular, we first calculate the optical flow between two real neighbor frames, then we warp a real frame according to the flow, meanwhile, we also warp the inverted frame by the same flow. Then we minimize the distance on two warped frames to form the temporal consistency loss, that is:
\begin{gather}
\hat{I}_{t-1} = \texttt{W}(I_{t-1},f_{t \Rightarrow t-1}),\\
\hat{O}_{t-1} = \texttt{W}(O_{t-1},f_{t \Rightarrow t-1}),\\
\mathcal{L}_{tc} = \sum_{t=2}^T ||\hat{I}_{t-1}-\hat{O}_{t-1}||_{1},
\label{eq8}
\end{gather}
where $f_{t \Rightarrow t-1}$ is the flow from frame $I_{t-1}$ to $I_{t}$. This loss enforces the temporal correlation in the videos to be the same as the input video, and improves the temporal smoothness.

\textbf{Final Loss.}
We get the final loss function for training the RIGID:
\begin{equation}
\mathcal{L}_{total}=\lambda_1\mathcal{L}_{rec}+\lambda_2\mathcal{L}_{tc}+\lambda_3\mathcal{L}_{ibfcc},
\label{eq10}
\end{equation}
where $\{\lambda_i\}$ denote the weight factors for balancing loss terms.

\begin{figure*}[t]
    \centering
    \captionsetup[subfloat]{labelformat=empty,justification=centering}
    \subfloat[]{
     \begin{minipage}{0.105\linewidth}
     \includegraphics[width=\linewidth]{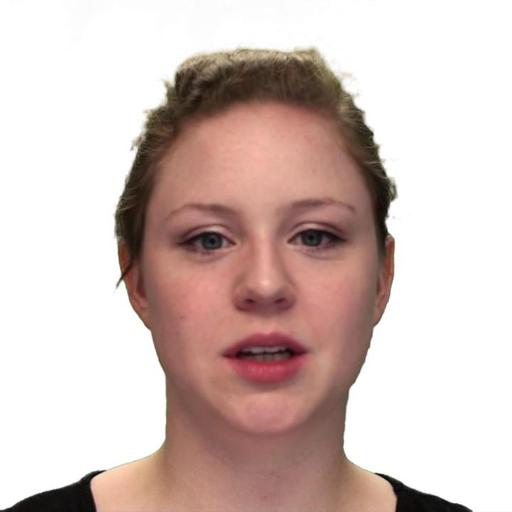}
     \includegraphics[width=\linewidth]{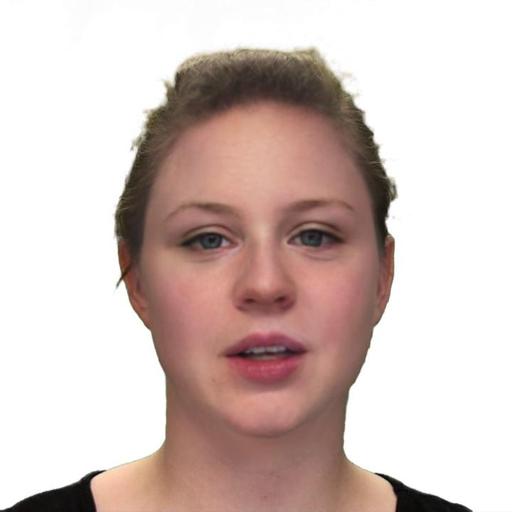}
     \includegraphics[width=\linewidth]{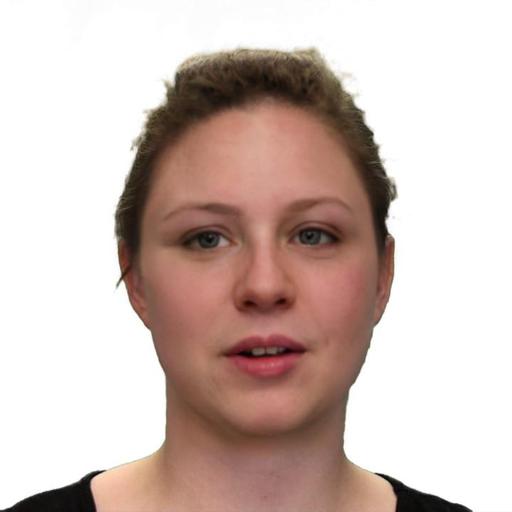}
     \includegraphics[width=\linewidth]{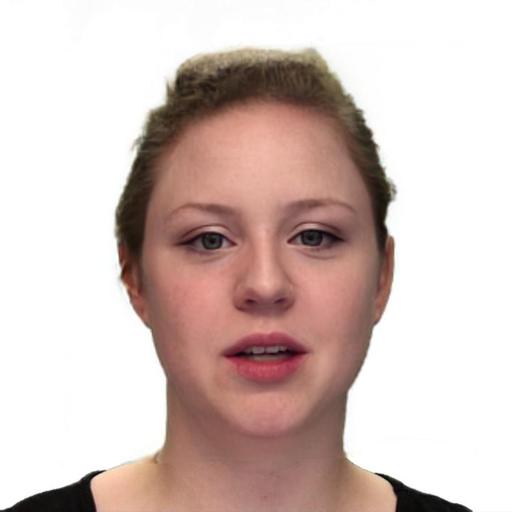}
     \includegraphics[width=\linewidth]{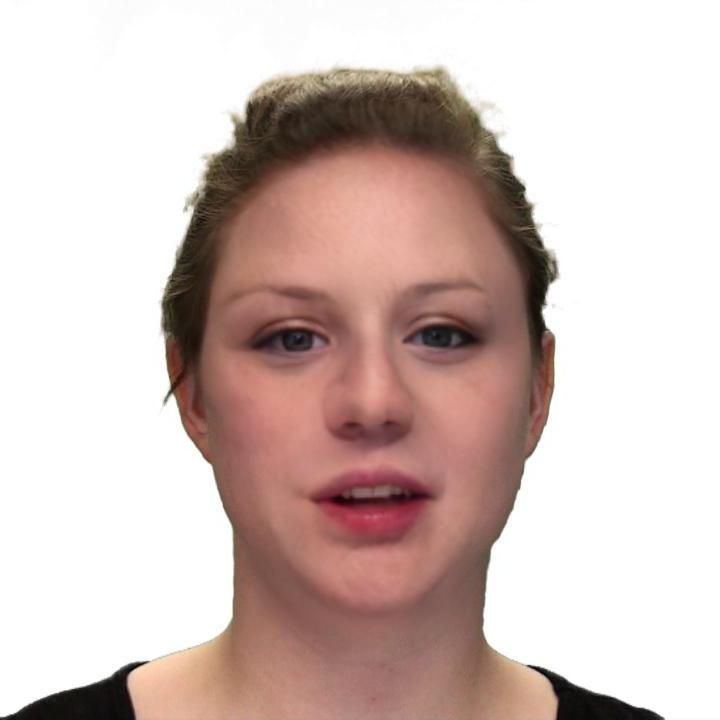}
     \end{minipage}
     }
    \hspace{-2.8mm}
    \subfloat[]{
     \begin{minipage}{0.105\linewidth}
     \includegraphics[width=\linewidth]{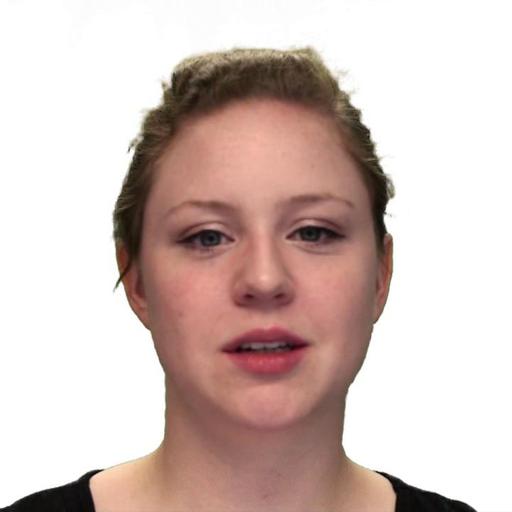}
     \includegraphics[width=\linewidth]{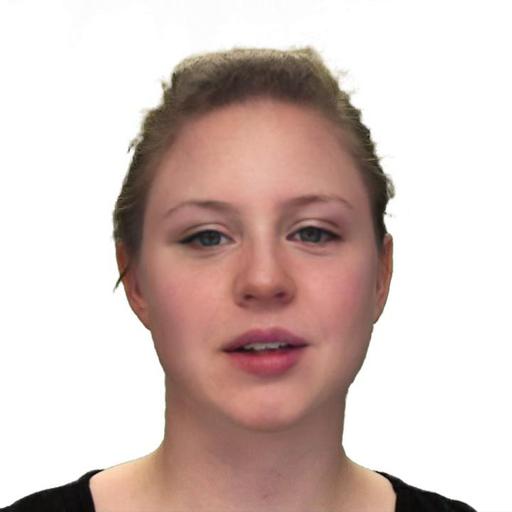}
     \includegraphics[width=\linewidth]{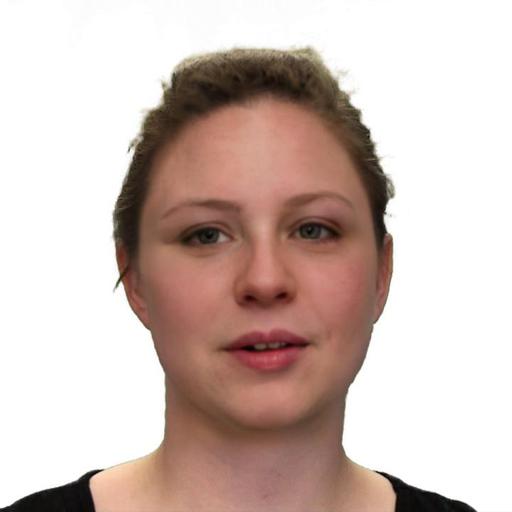}
     \includegraphics[width=\linewidth]{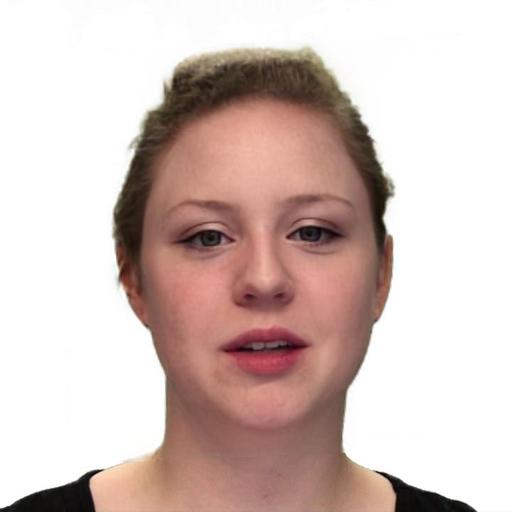}
     \includegraphics[width=\linewidth]{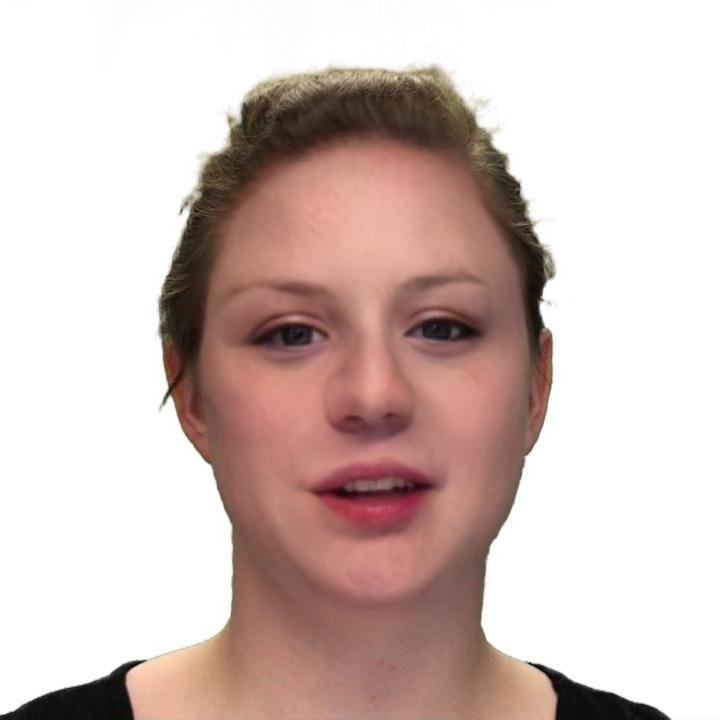}
     \end{minipage}
     }
    \hspace{-2.8mm}
    \subfloat[]{
     \begin{minipage}{0.105\linewidth}
     \includegraphics[width=\linewidth]{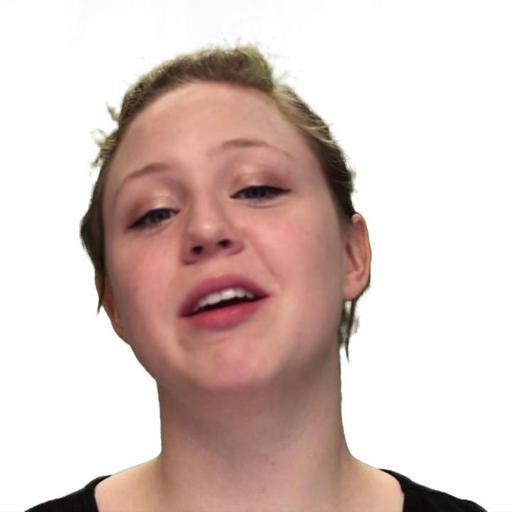}
     \includegraphics[width=\linewidth]{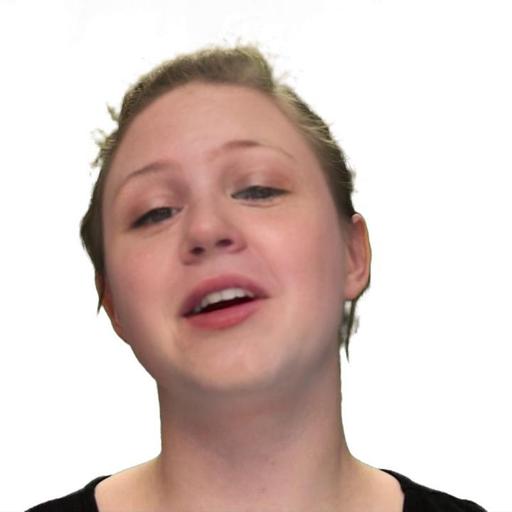}
     \includegraphics[width=\linewidth]{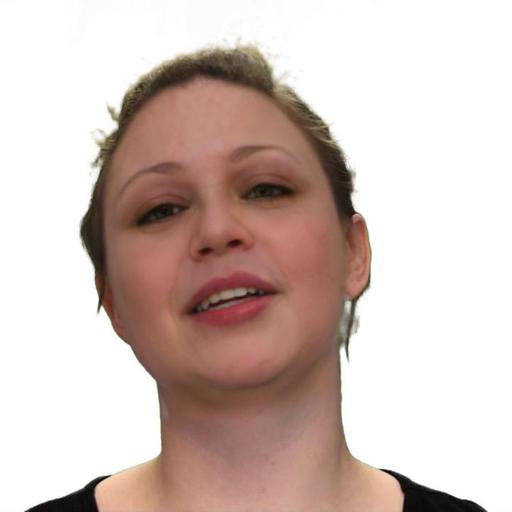}
     \includegraphics[width=\linewidth]{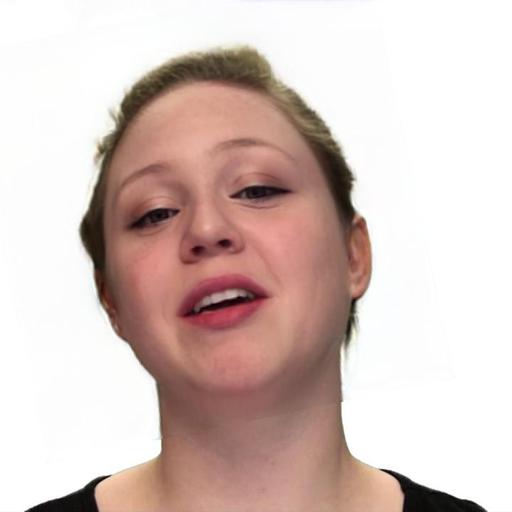}
     \includegraphics[width=\linewidth]{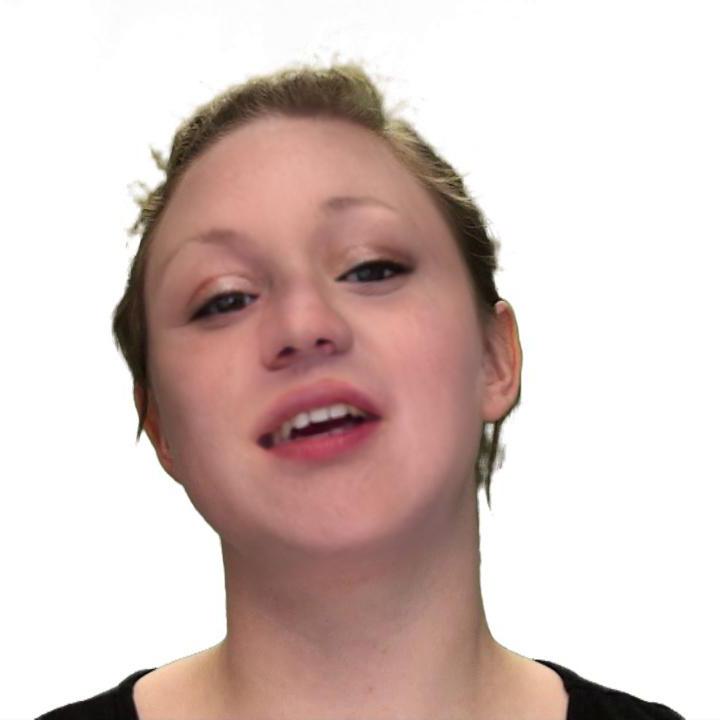}
     \end{minipage}
     }
    \hspace{-1.8mm}
    \subfloat[]{
     \begin{minipage}{0.105\linewidth}
     \includegraphics[width=\linewidth]{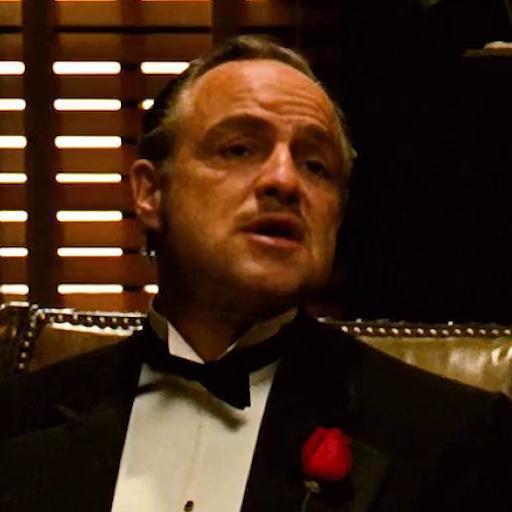}
     \includegraphics[width=\linewidth]{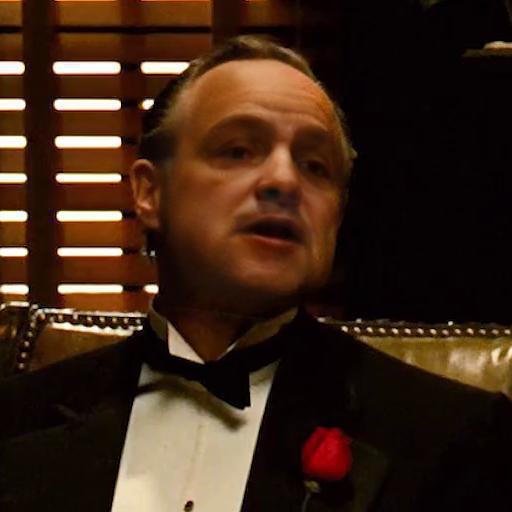}
     \includegraphics[width=\linewidth]{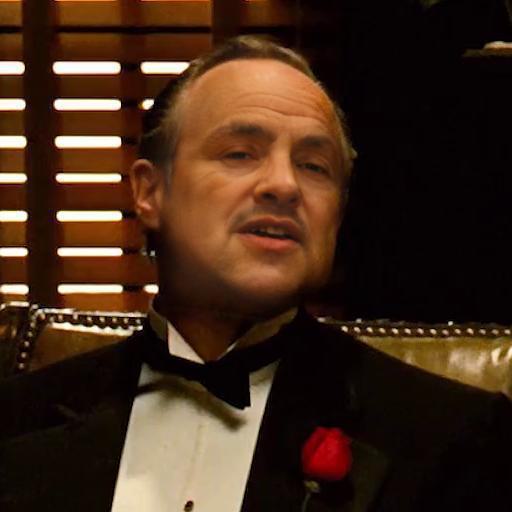}
     \includegraphics[width=\linewidth]{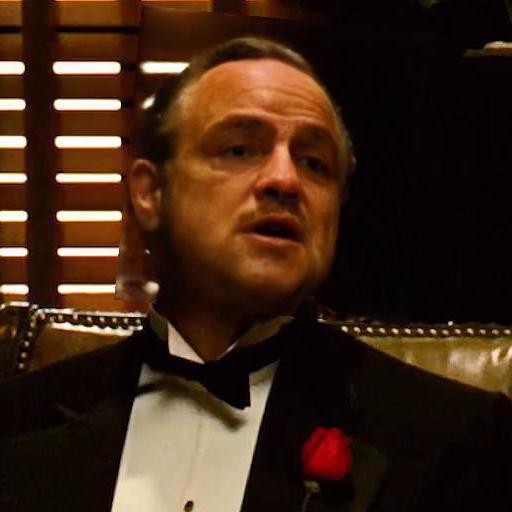}
     \includegraphics[width=\linewidth]{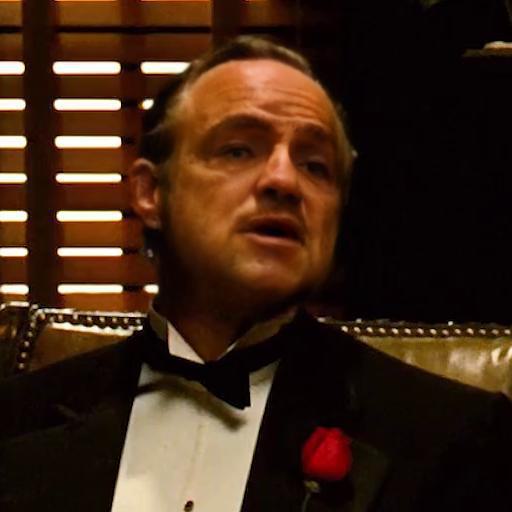}
     \end{minipage}
     }
    \hspace{-2.8mm}
    \subfloat[]{
     \begin{minipage}{0.105\linewidth}
     \includegraphics[width=\linewidth]{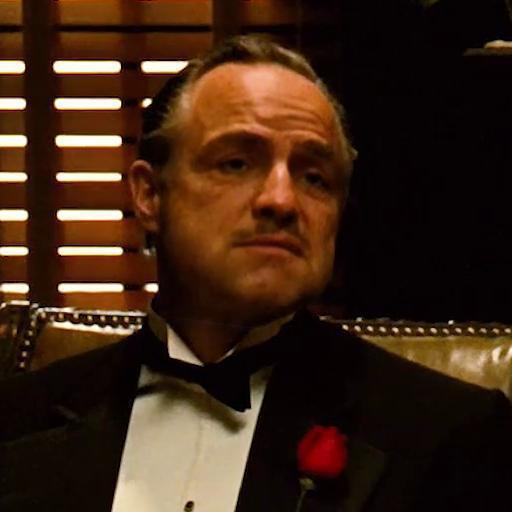}
     \includegraphics[width=\linewidth]{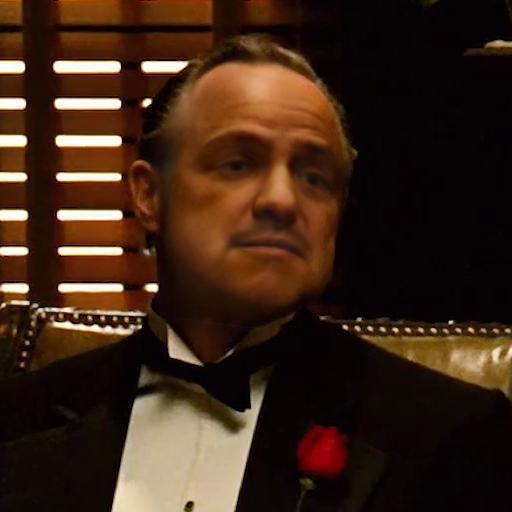}
     \includegraphics[width=\linewidth]{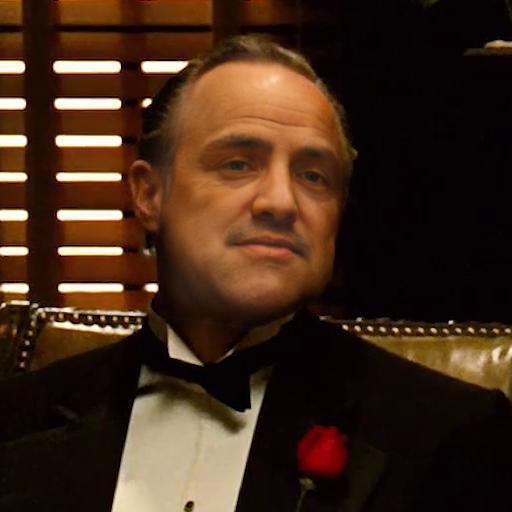}
     \includegraphics[width=\linewidth]{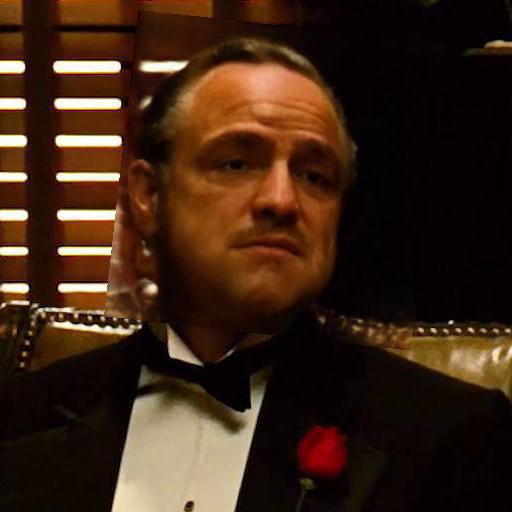}
     \includegraphics[width=\linewidth]{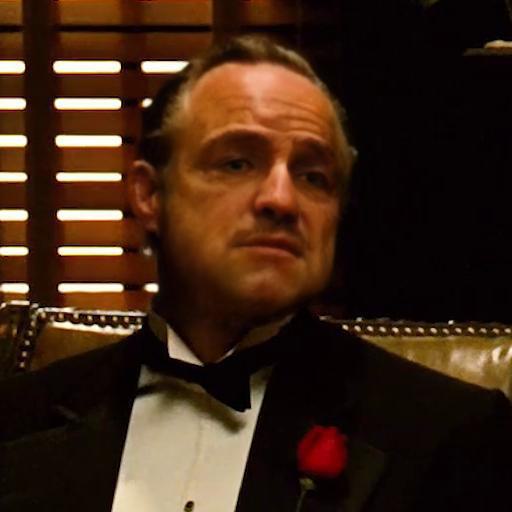}
     \end{minipage}
     }
    \hspace{-2.8mm}
    \subfloat[]{
     \begin{minipage}{0.105\linewidth}
     \includegraphics[width=\linewidth]{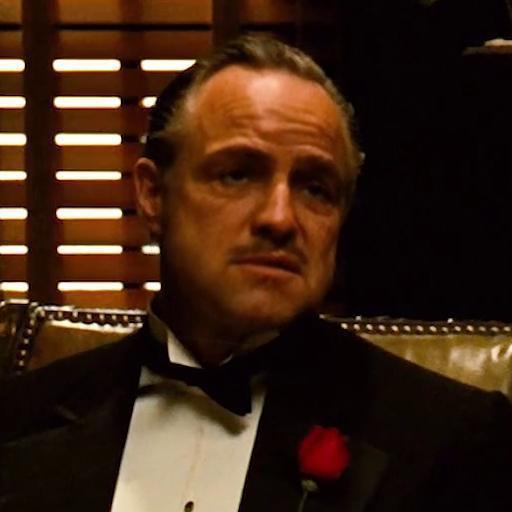}
     \includegraphics[width=\linewidth]{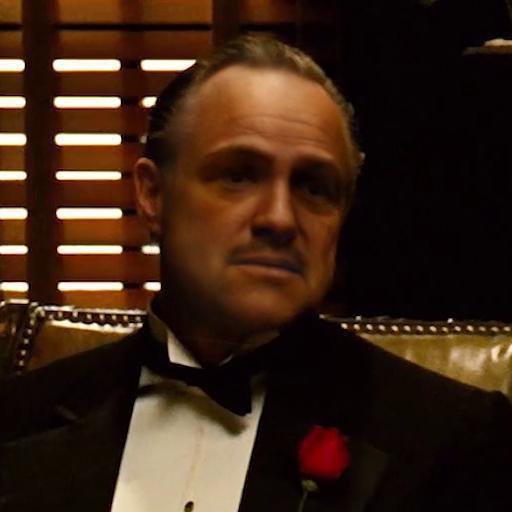}
     \includegraphics[width=\linewidth]{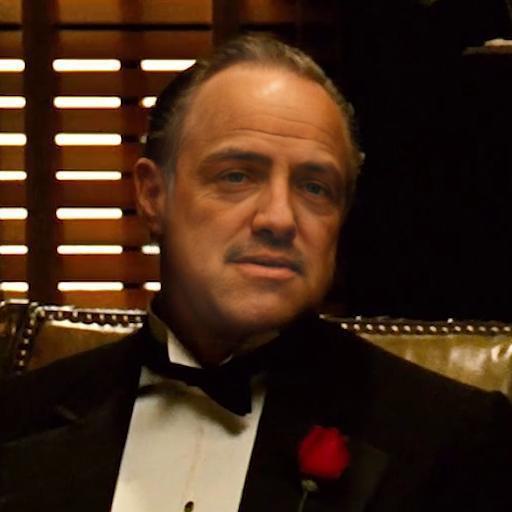}
     \includegraphics[width=\linewidth]{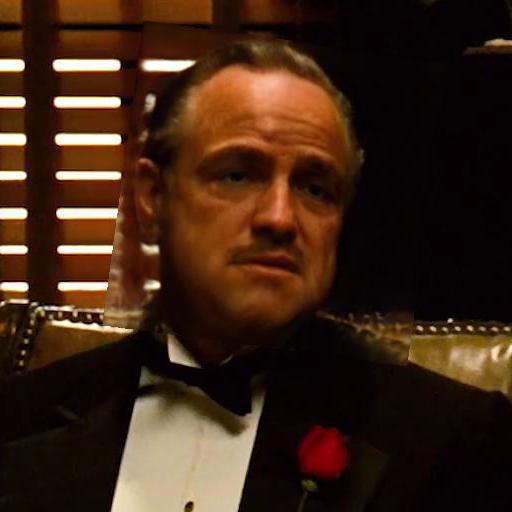}
     \includegraphics[width=\linewidth]{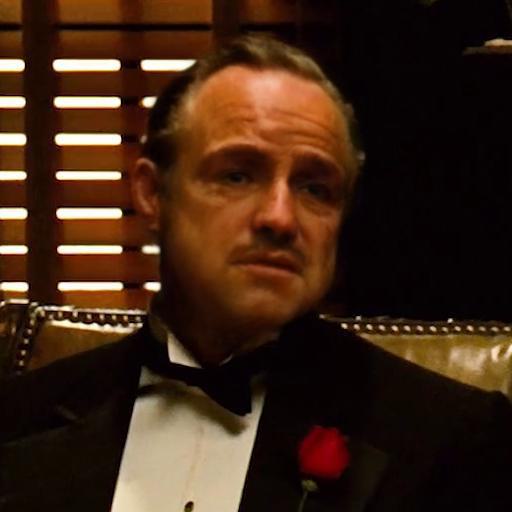}
     \end{minipage}
     }
    \hspace{-1.8mm}
    \subfloat[]{
     \begin{minipage}{0.105\linewidth}
     \includegraphics[width=\linewidth]{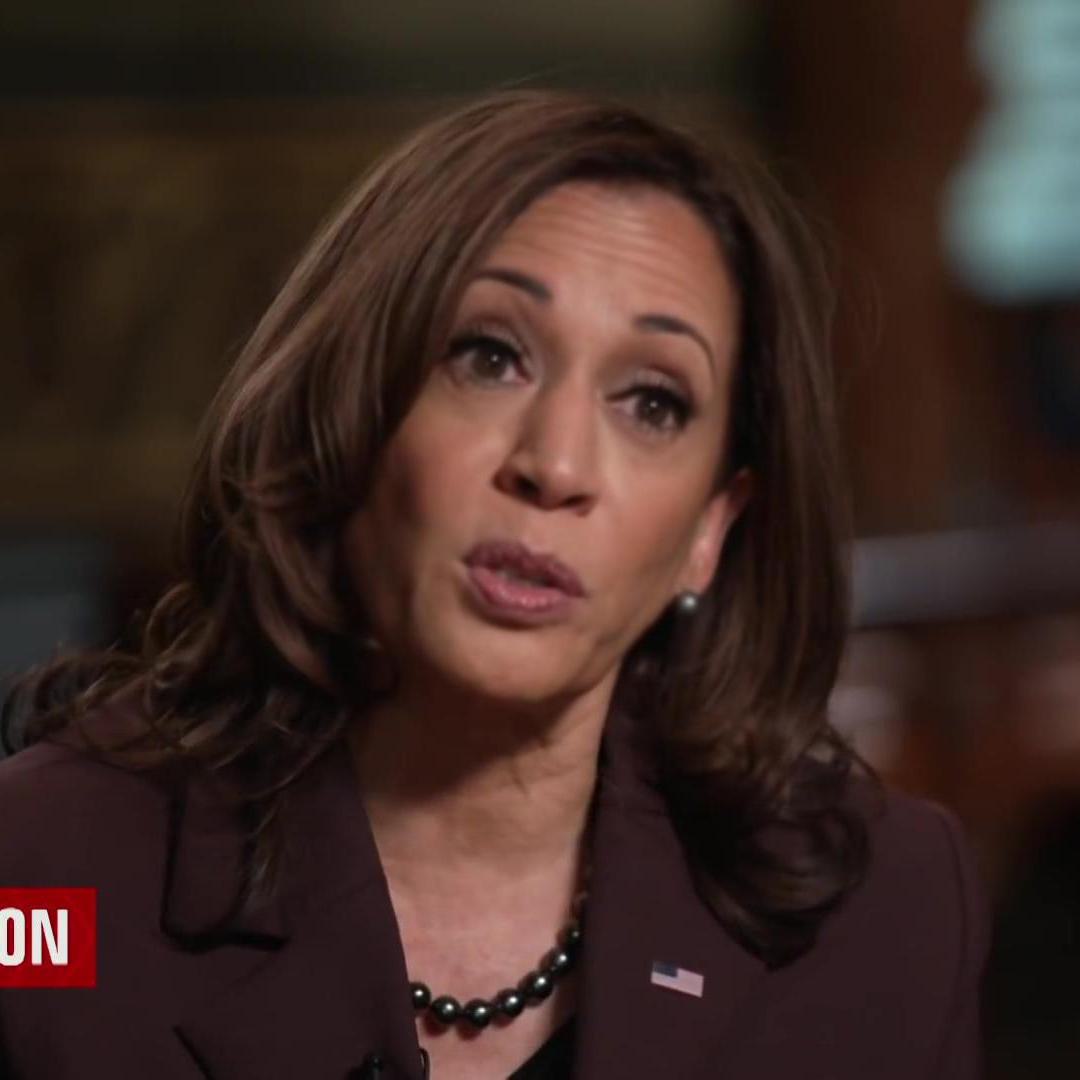}
     \includegraphics[width=\linewidth]{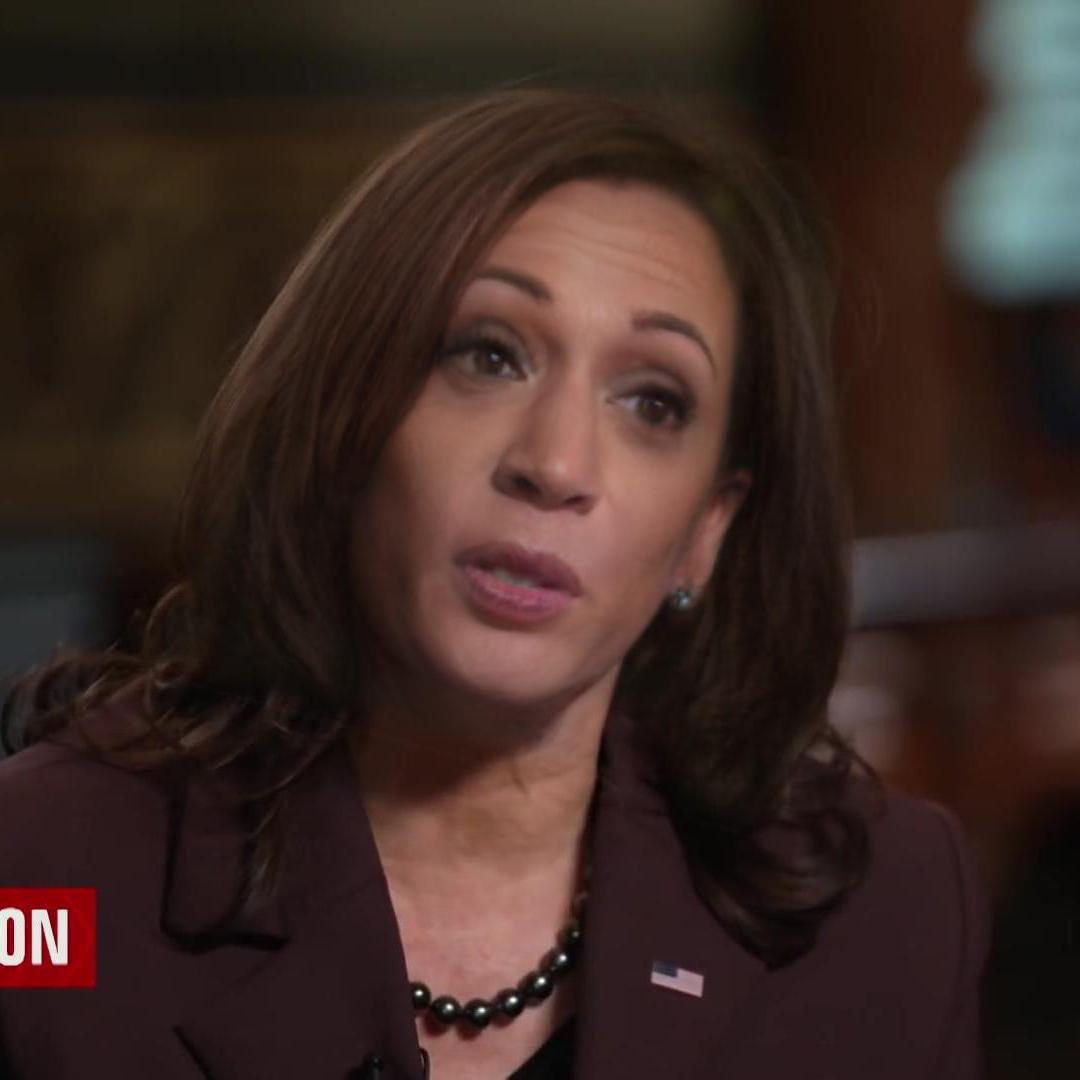}
     \includegraphics[width=\linewidth]{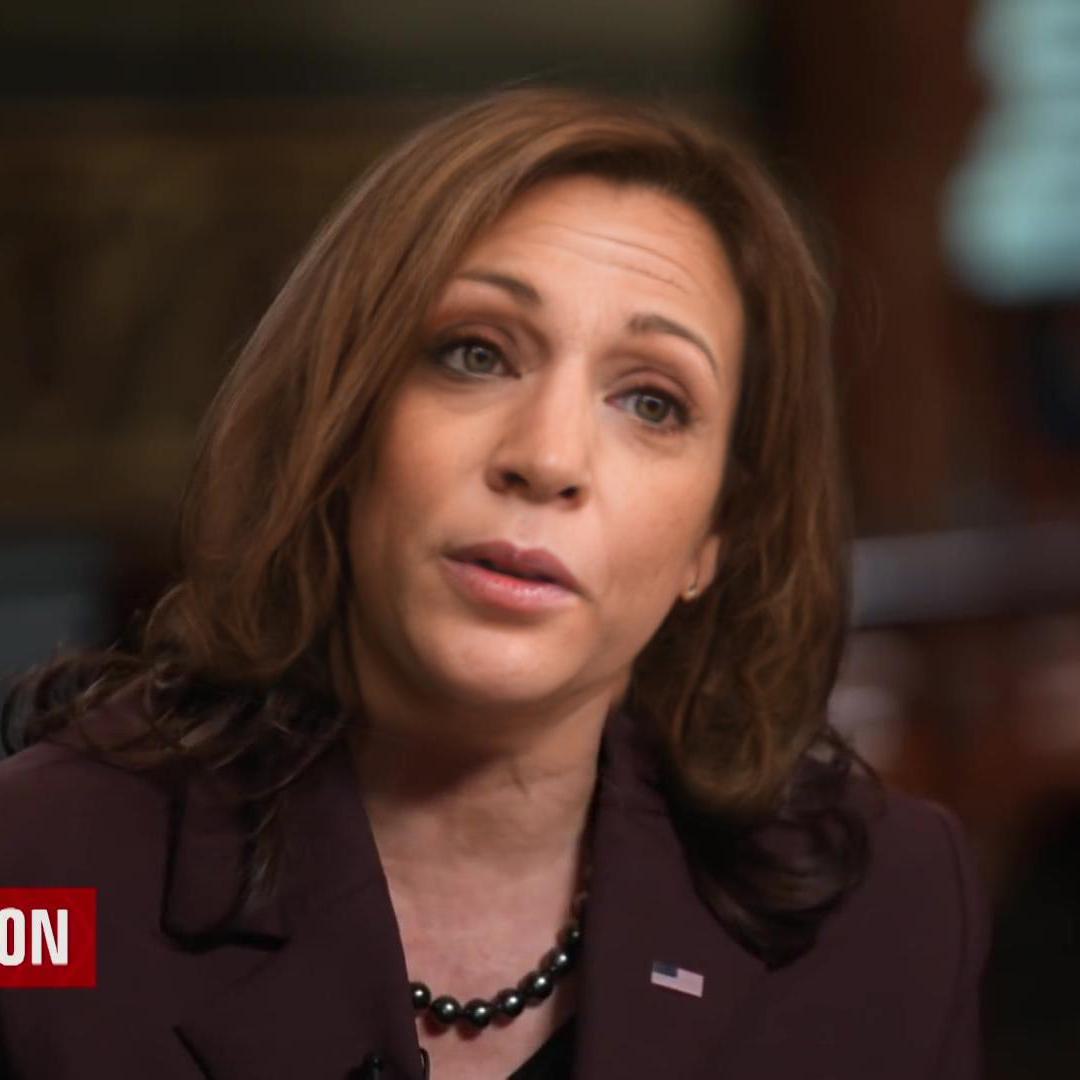}
     \includegraphics[width=\linewidth]{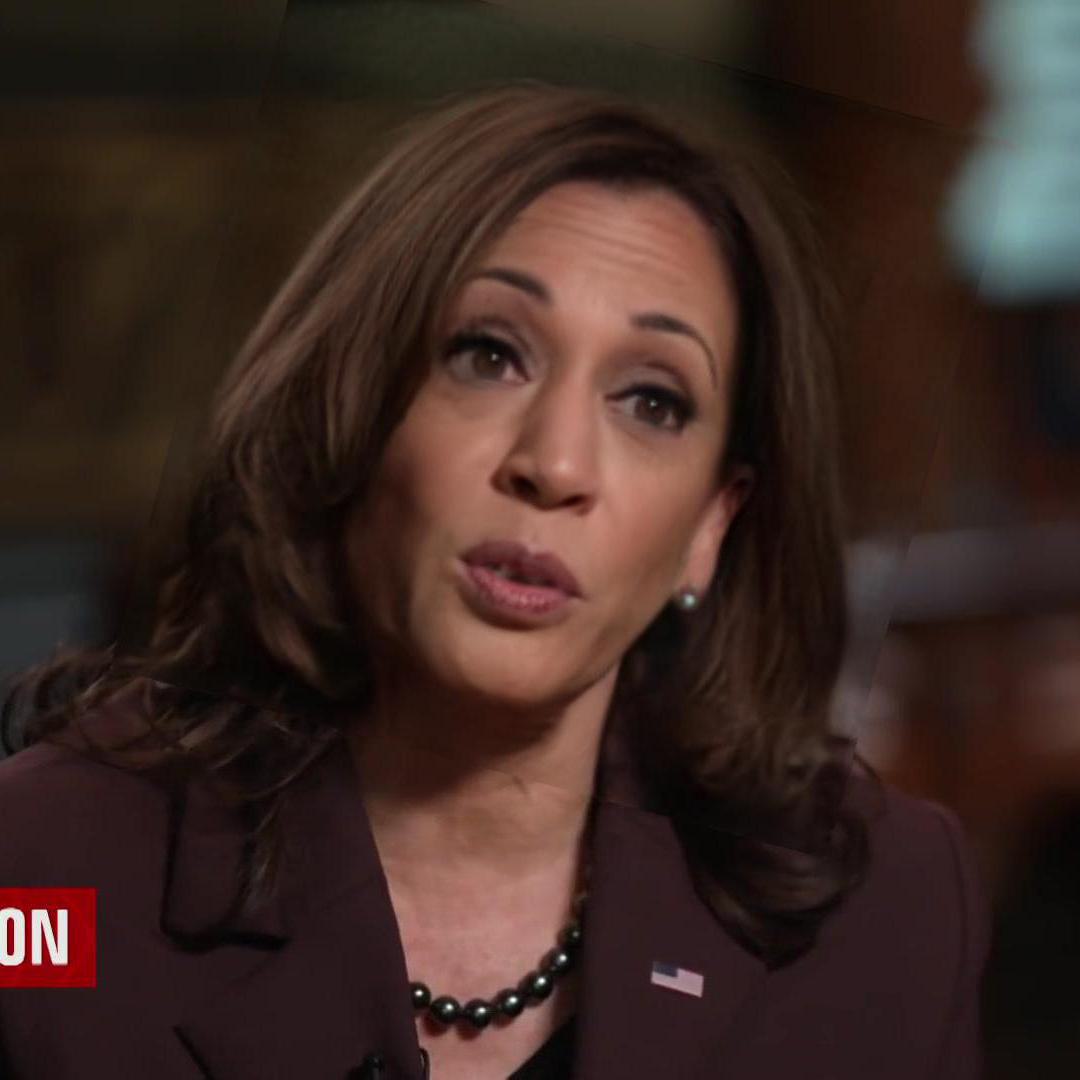}
     \includegraphics[width=\linewidth]{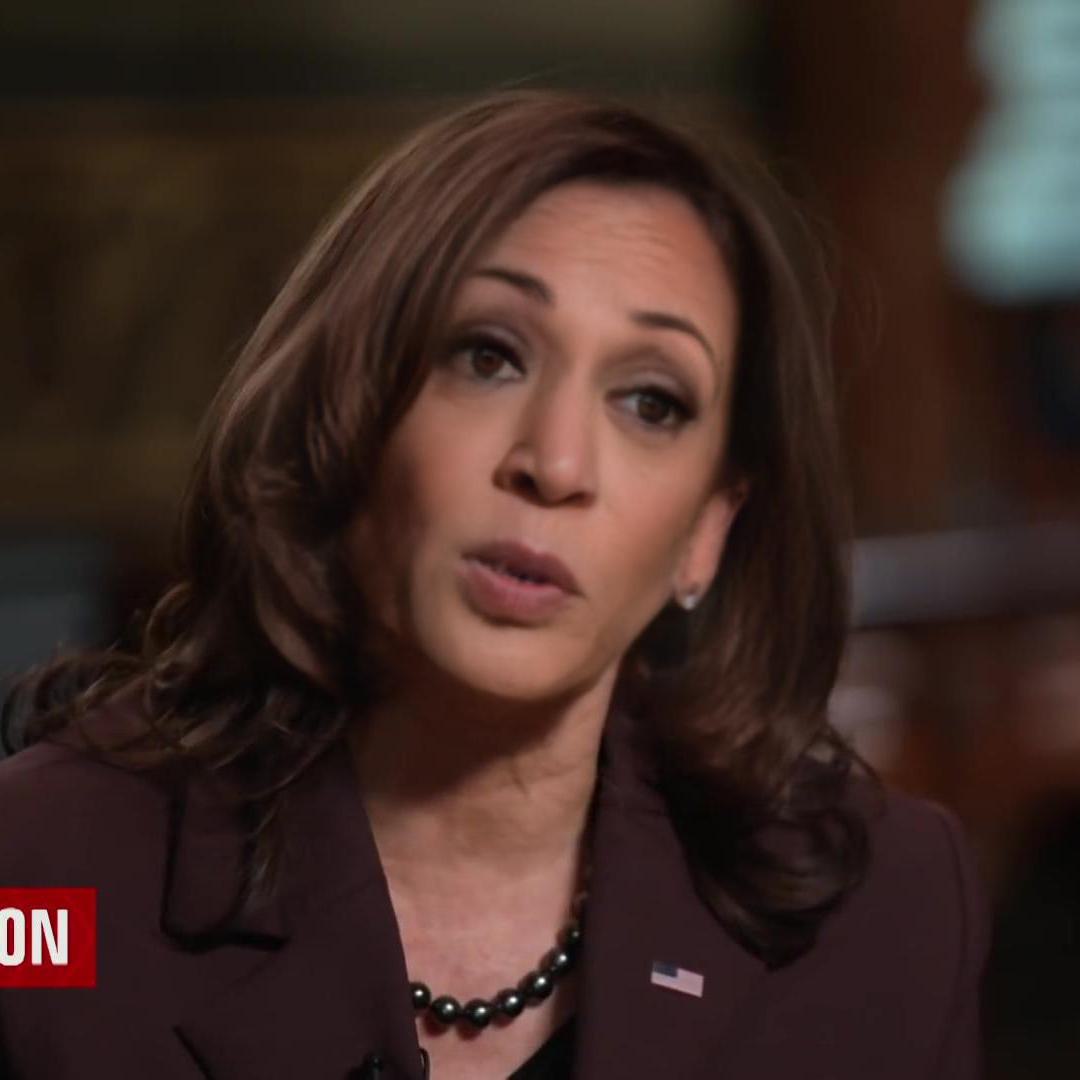}
     \end{minipage}
     }
    \hspace{-2.8mm}
    \subfloat[]{
     \begin{minipage}{0.105\linewidth}
     \includegraphics[width=\linewidth]{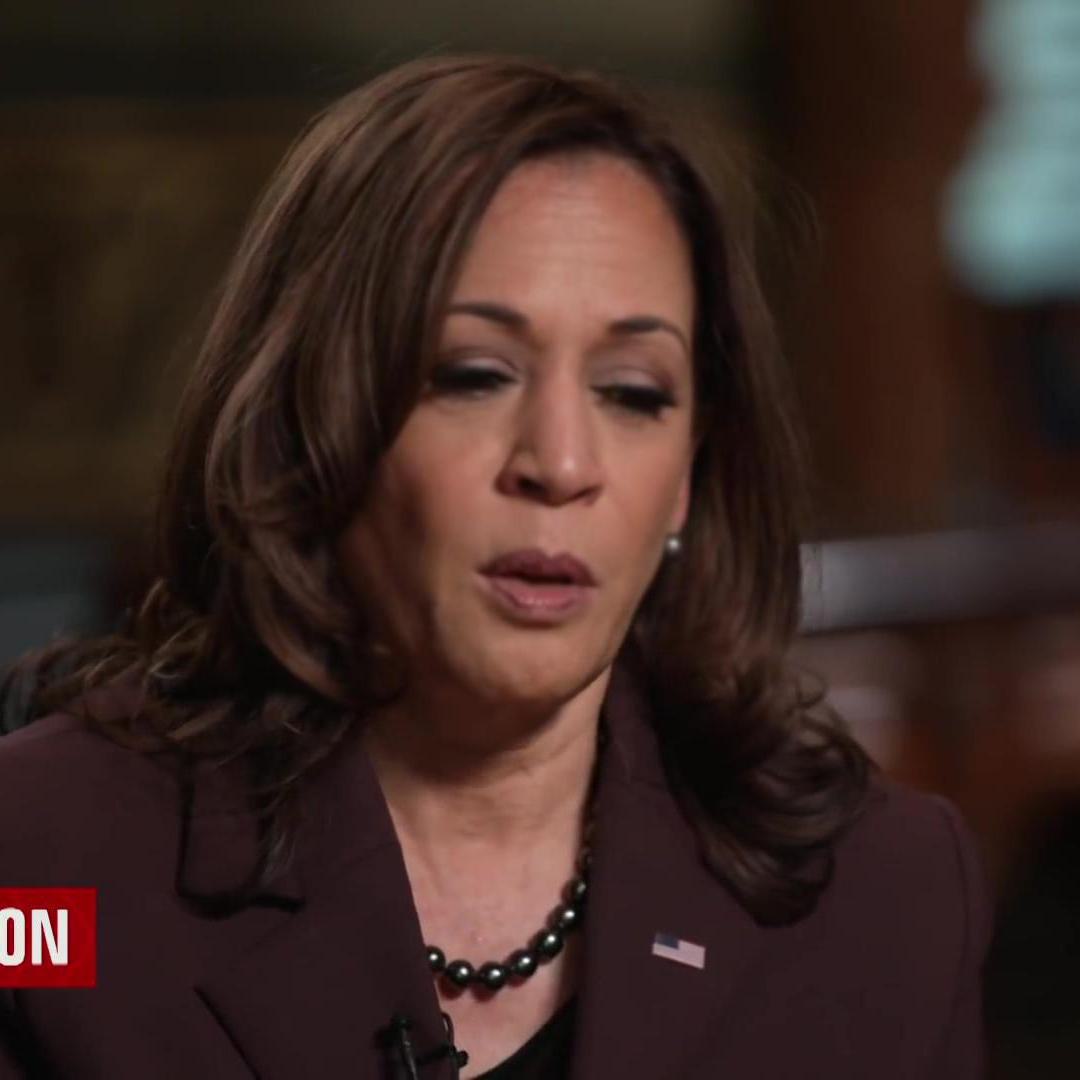}
     \includegraphics[width=\linewidth]{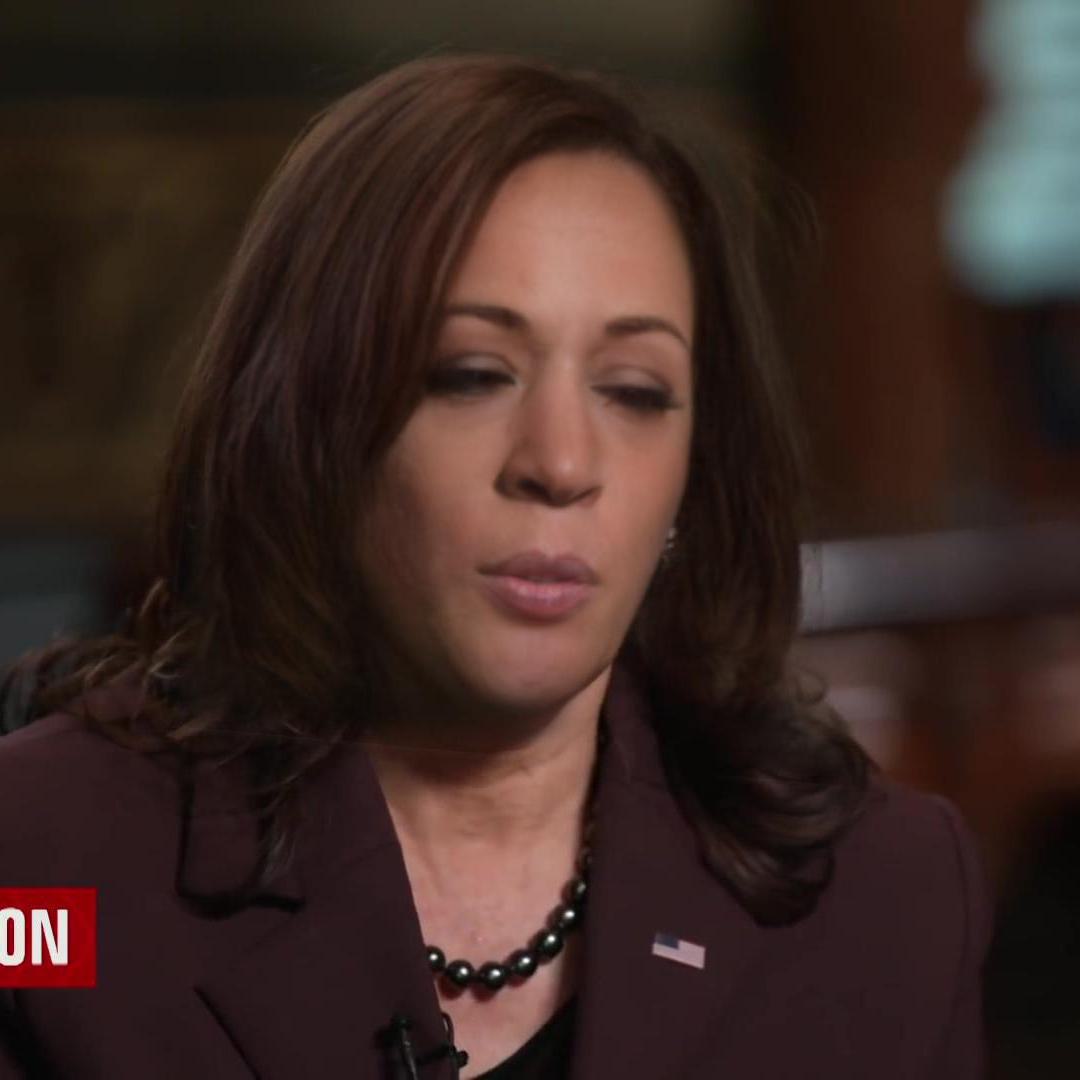}
     \includegraphics[width=\linewidth]{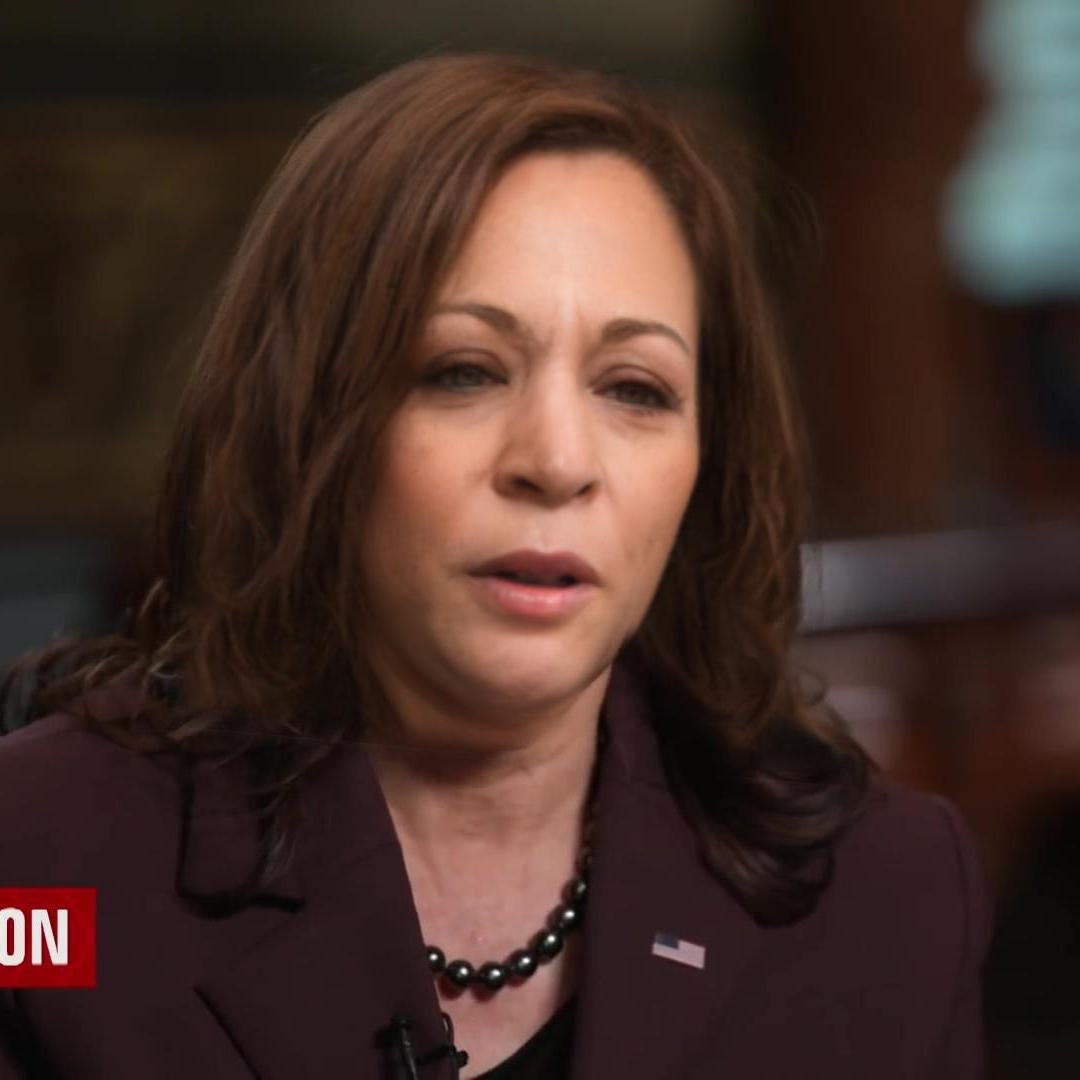}
     \includegraphics[width=\linewidth]{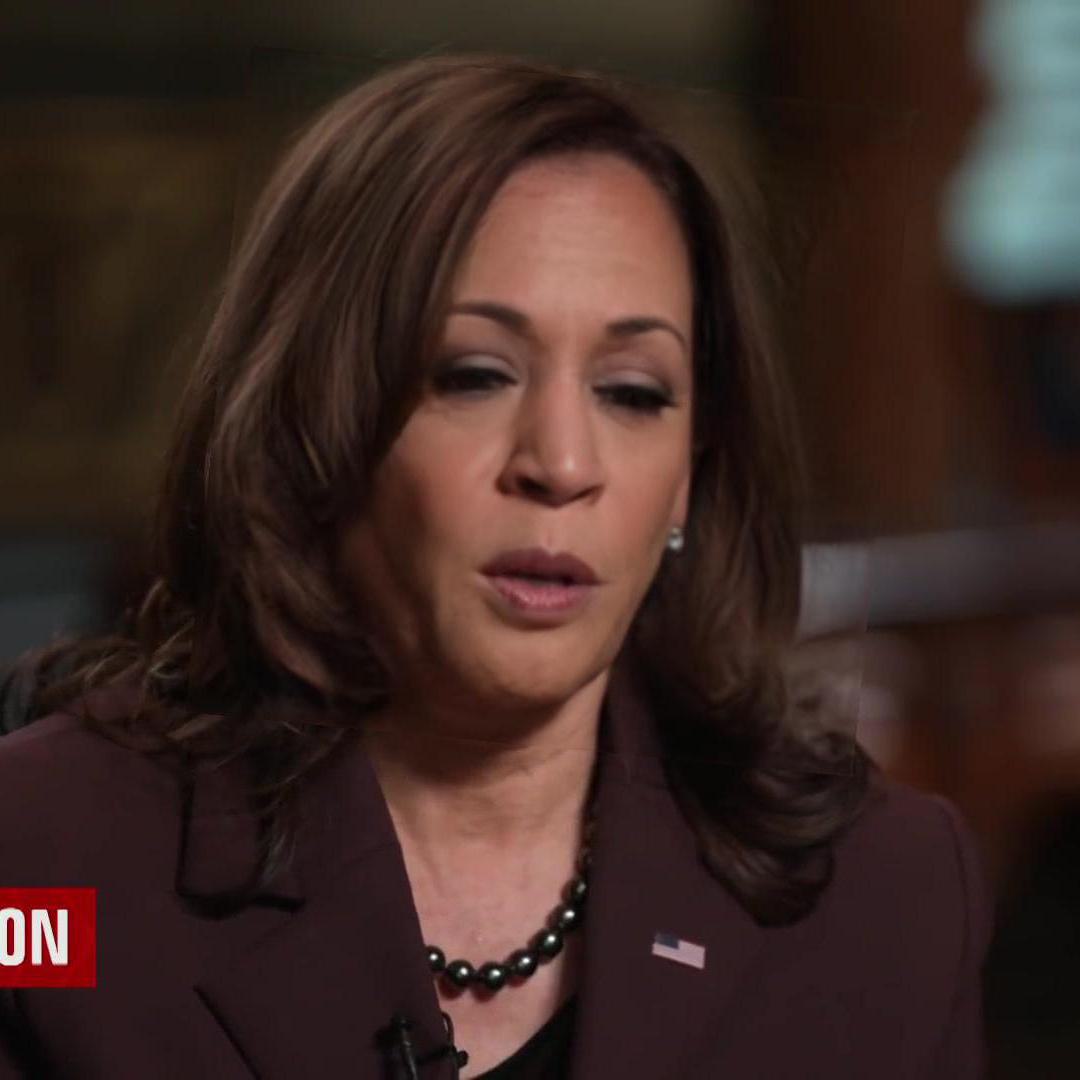}
     \includegraphics[width=\linewidth]{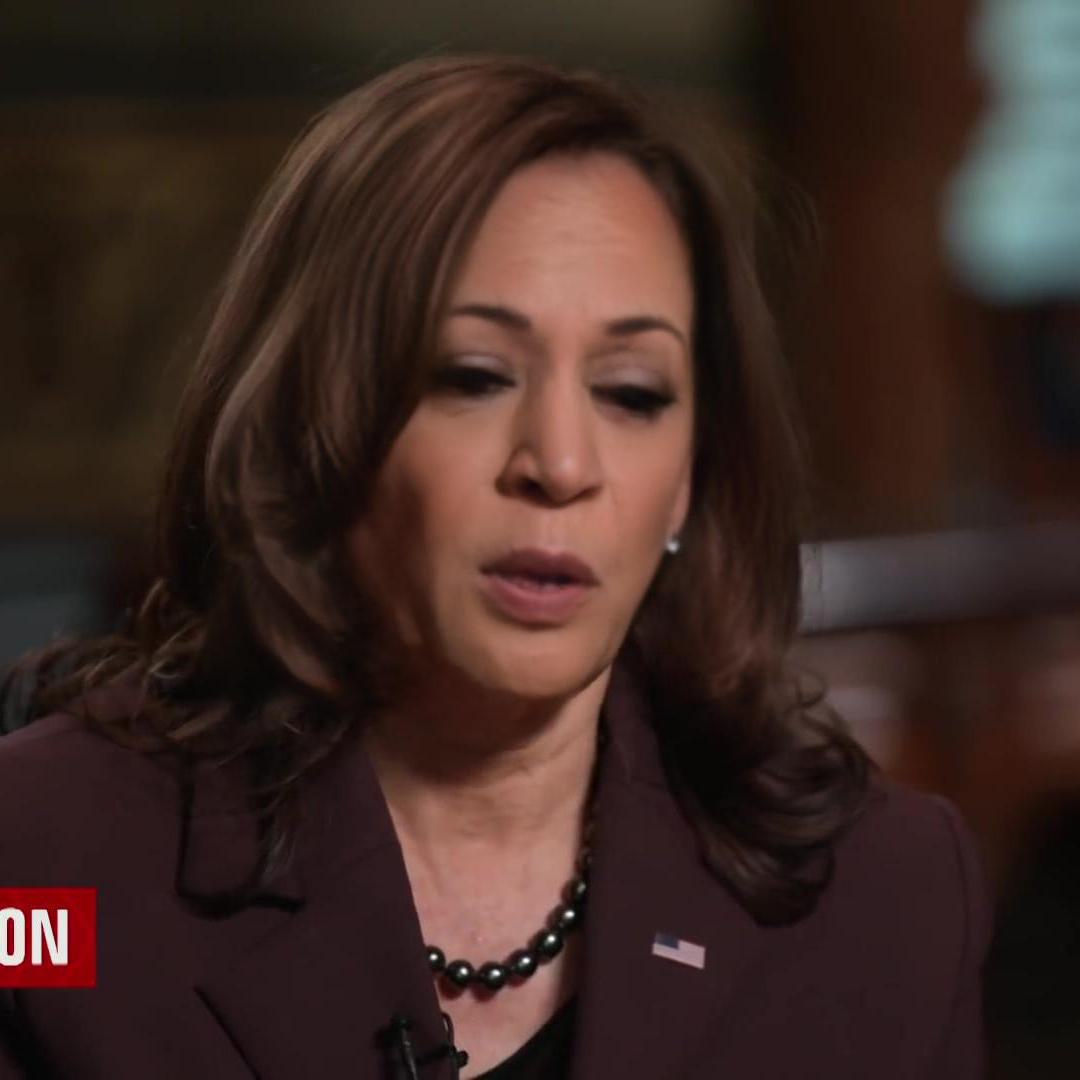}
     \end{minipage}
     }
    \hspace{-2.8mm}
    \subfloat[]{
     \begin{minipage}{0.105\linewidth}
     \includegraphics[width=\linewidth]{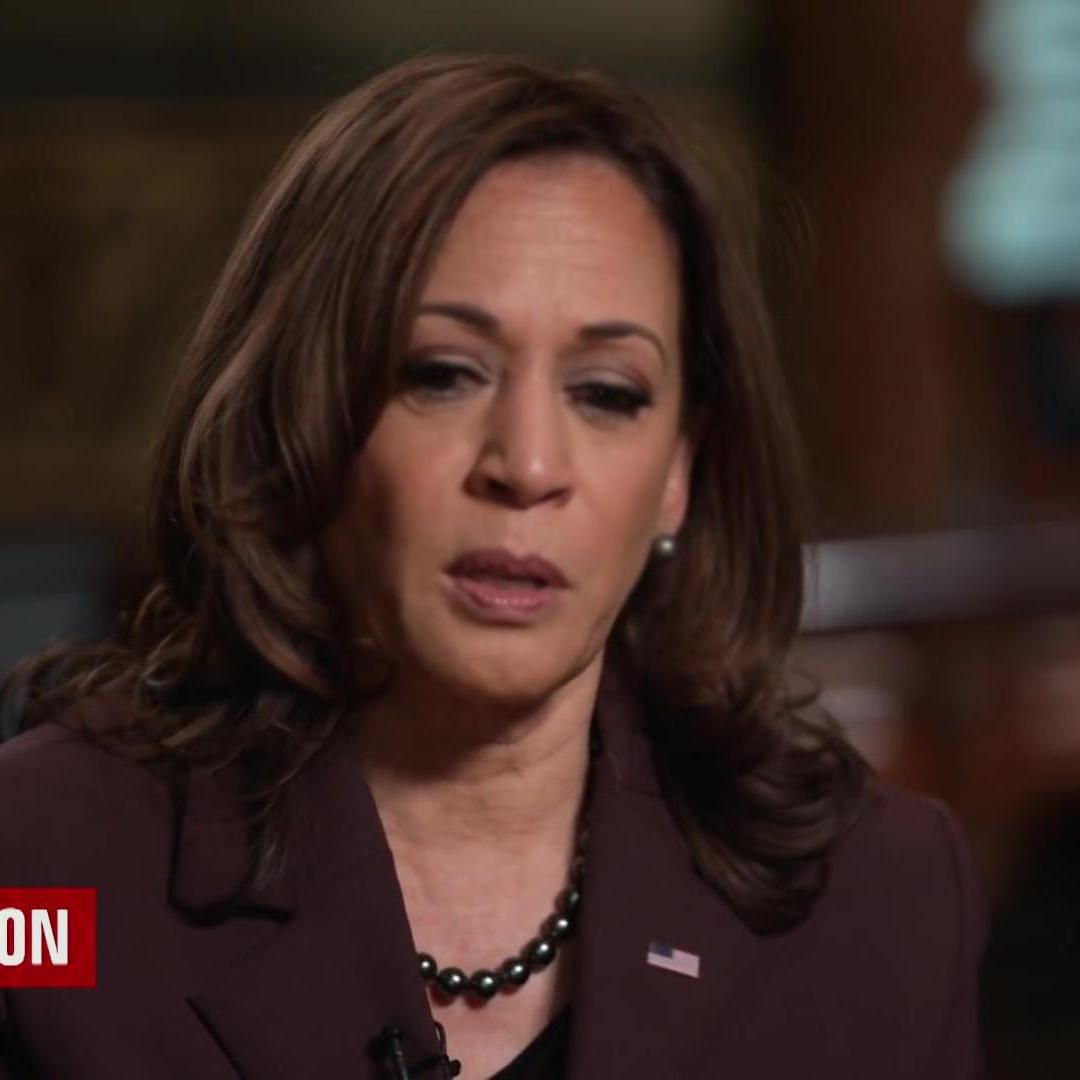}
     \includegraphics[width=\linewidth]{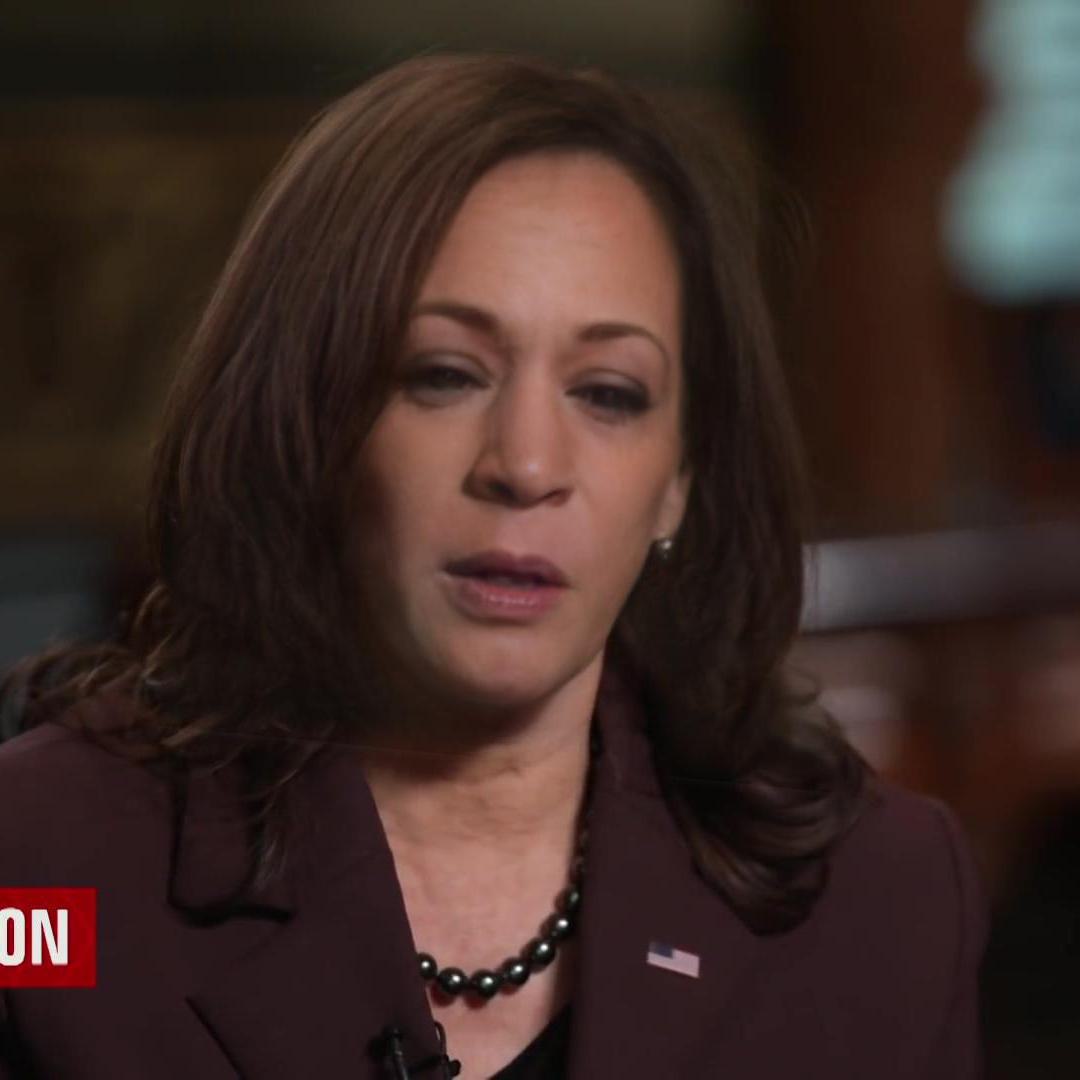}
     \includegraphics[width=\linewidth]{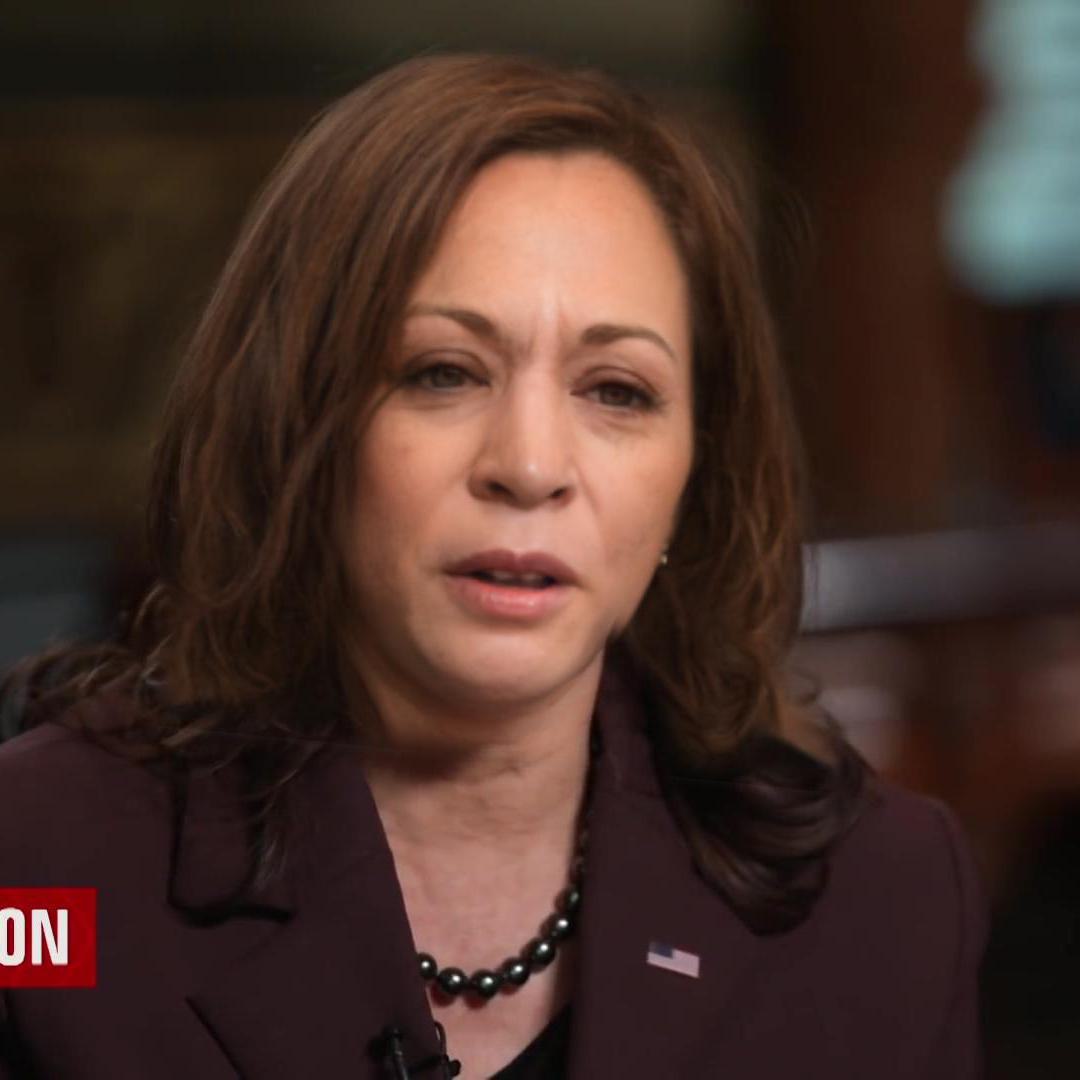}
     \includegraphics[width=\linewidth]{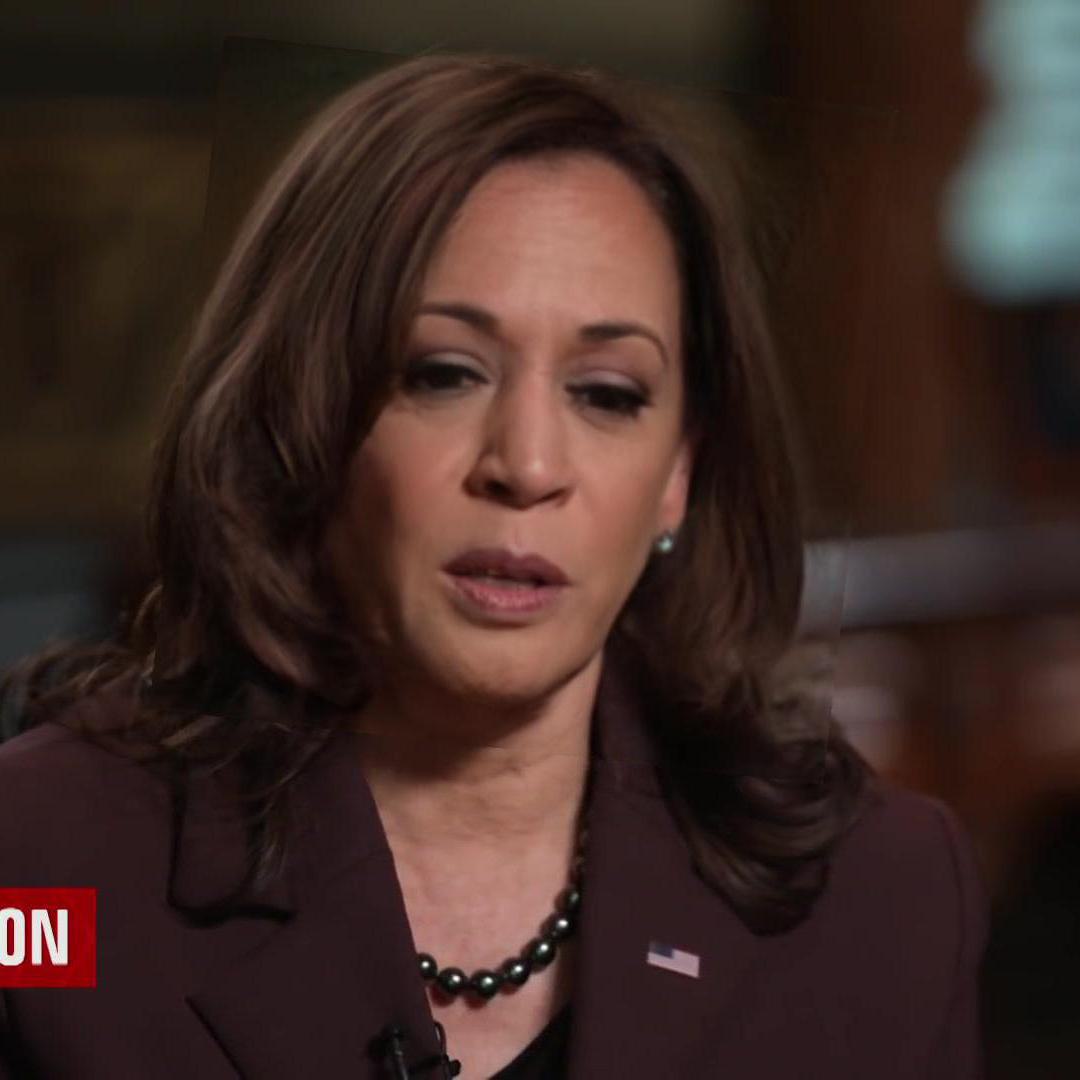}
     \includegraphics[width=\linewidth]{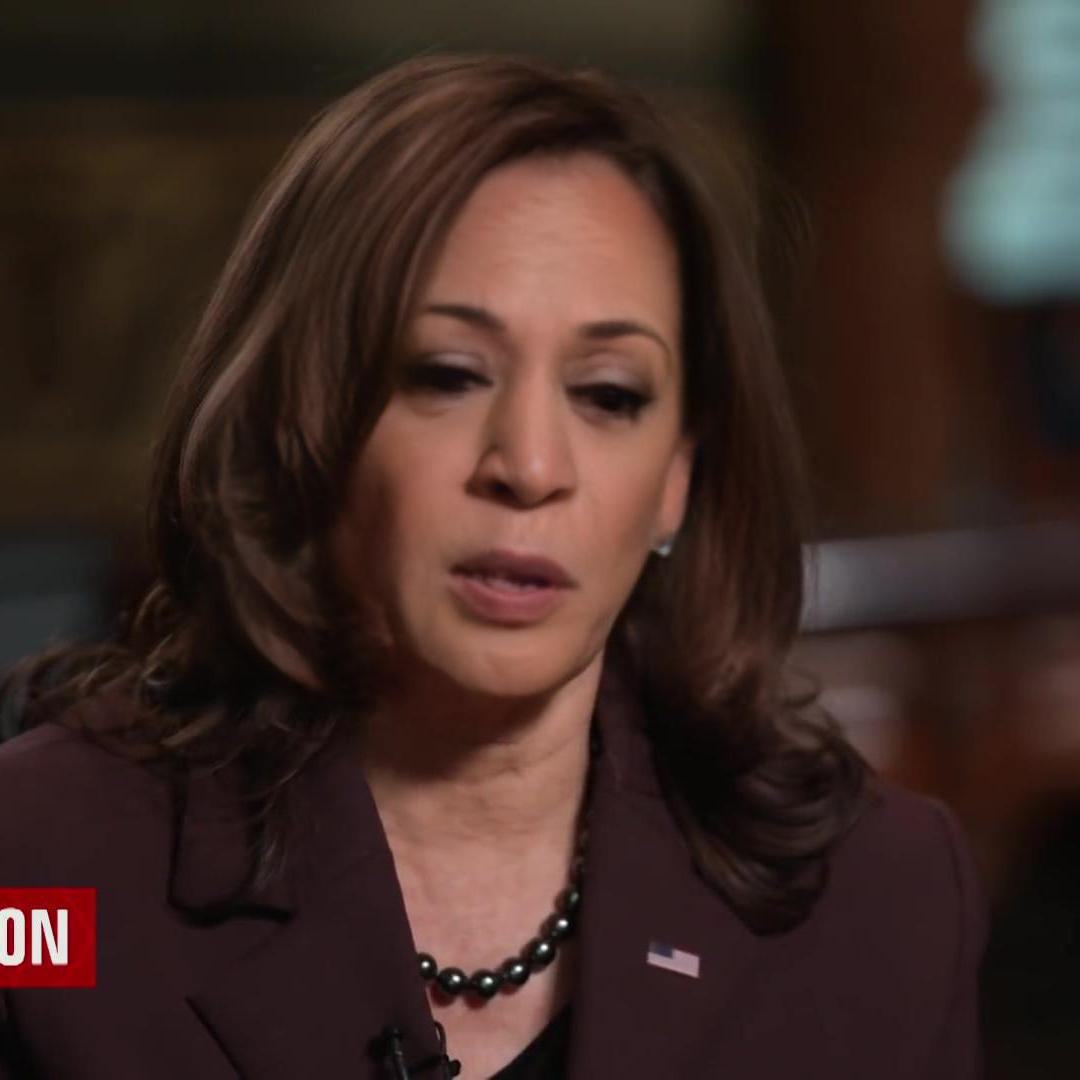}
     \end{minipage}
     }
\rotatebox[origin=c]{270}{\hspace{-0cm} \footnotesize{Original} \hspace{7.5mm} \footnotesize{IGCI} \hspace{7.5mm} \footnotesize{Latent-T.} \hspace{5mm} \footnotesize{STIT / TCSVE} \hspace{5mm} \footnotesize{RIGID} }
\vspace{-4mm}
\caption{Qualitative comparison on the video inversion task. Our learning-based RIGID can faithfully reconstruct the video frames that is comparable to those expensive optimization-based works.}
\label{fig:reconstruction_exp}
\end{figure*}

\section{Experiments}

\subsection{Implementation Details}
\label{sec:imp}

We implement the proposed framework in Pytorch on a PC with an Nvidia GeForce RTX 3090. We chose the StyleGAN2~\cite{karras2020analyzing} pre-trained on the FFHQ dataset~\cite{karras2019style} with the resolution of $256\times 256$ as our generator due to its strong edit ability. The corresponding $w$ latent code has the dimension of $14\times512$, and we set the early $10\times512$ as the former part and the rest of $4\times512$ as the latter part in the latent frequency disentanglement. The framework is optimized by the Adam optimizer~\cite{kingma2014adam} with the learning rate of $1e^{-4}$. We empirically set the balancing weights in Eq.~\ref{eq10} as $\lambda_{1}=1$, $\lambda_{2}=2$, and $\lambda_{3}=5$. Limited by the memory size of GPU, we set $T=6$ in Eq.~\ref{eq8}, which indicates each training episode contains 6 consecutive frames. In addition, we inject the noise map $n_t$ at the resolution of $32\times 32$ to the generator. 

The recurrent encoder consists of 7 convolutional (Conv) layers, a ConvLSTM layer, and a fully connected (FC) layer. We take ``\texttt{Fused Leaky ReLU}'' as activation in-between Conv layers. The ConvLSTM layer is integrated between the Conv and FC layers. The visible net has a U-Net structure~\cite{ronneberger2015u}, it takes 5 Conv layers as encoder, and 5 Transposed convolutional layers as a decoder. BatchNorm layer and leakyReLU activation are integrated into between layers. Besides, a Sigmoid function is applied on output U-Net for normalizing the visibility map values. Visible net takes the concatenation of two warped results with 6 channels as input and outputs a visibility map with 1 channel. It is trained using Adam optimizer~\cite{kingma2014adam} with the learning rate of $1e^-4$ with 100,000 iterations. Besides, we use pre-trained FlowNet2~\cite{ilg2017flownet} for predicting optical flow, and we implement warping operation using bi-linear interpolation.


\begin{table*}[h]
   \caption{Video inversion comparisons on two datasets. $\downarrow$ denotes the lower the better and the best results are marked in \textbf{bold}, and the values of MSE are magnified 100 times.}
       \vspace{-6mm}
       \begin{center}
       \setlength{\tabcolsep}{0.15cm}{
       \begin{tabular}{c|c|c|c|c|c|c|c|c|c}
           \hline
           \multirow{1}{*}{\diagbox{\small{Methods}}{\small{Metrics}}}       & \multicolumn{4}{c|}{RAVDESS-72 Dataset} & \multicolumn{4}{c|}{In-the-Wild-36 Dataset} &\multirow{2}{*}{IT$\downarrow$(s)}\\
           \cline{2-9}
          &{MSE$\downarrow$}   &{LPIPS$\downarrow$}  &{WE$\downarrow$}   &{FVD$\downarrow$}  &{MSE$\downarrow$($\times$e-2)}  &{LPIPS$\downarrow$}  &{WE$\downarrow$}  &{FVD$\downarrow$}   \\
           \hline
           IGCI~\cite{Xu2021ICCV}          &\textbf{0.99}  &\textbf{0.05} &154.21 &412.33            &\textbf{2.01}&0.13&432.98&276.92 &1.2$\times$e5      \\
           Latent-T.~\cite{yao2021latent}  &5.68                   &0.12 &86.356  &220.46                             &5.31&0.22&367.32&165.35 &\textbf{49.2}     \\
           STIT~\cite{tzaban2022stitch}/TCSVE~\cite{xu2022temporally}          &\textbf{0.99}  &\textbf{0.05} &\textbf{83.21}  &\textbf{171.23}   &2.32&\textbf{0.11}&293.83&81.03 &851.5           \\

           \hline
           RIGID                           &1.04     &\textbf{0.05}   &84.62  &174.55                &2.31&0.12&\textbf{287.32}&\textbf{74.11} &  54.5\\
           \hline
       \end{tabular}
       }
      \end{center}
      \label{table:reconstruction}
      \vspace{-5mm}
\end{table*} 

\subsection{Experimental Settings}
\label{sec:exps}
\textbf{Datasets.} We collect datasets both under control and in the wild environment. We select 72 videos from the controlled RAVDESS dataset~\cite{livingstone2018ryerson}, which we called RAVDESS-72 Dataset. It contains 9,045 frames and each video contains about 120 frames. We also collect 36 videos from the Internet with the various poses, expressions, and backgrounds. We name them as In-the-Wild-36 Dataset, it contains 7,532 frames in total. We combine two datasets together, use 85 videos for training our RIGID and the rest 23 videos for testing. 

\begin{figure*}[t]
    \centering
    \captionsetup[subfloat]{labelformat=empty,justification=centering}
    \subfloat[]{
     \begin{minipage}{0.105\linewidth}
     \includegraphics[width=\linewidth]{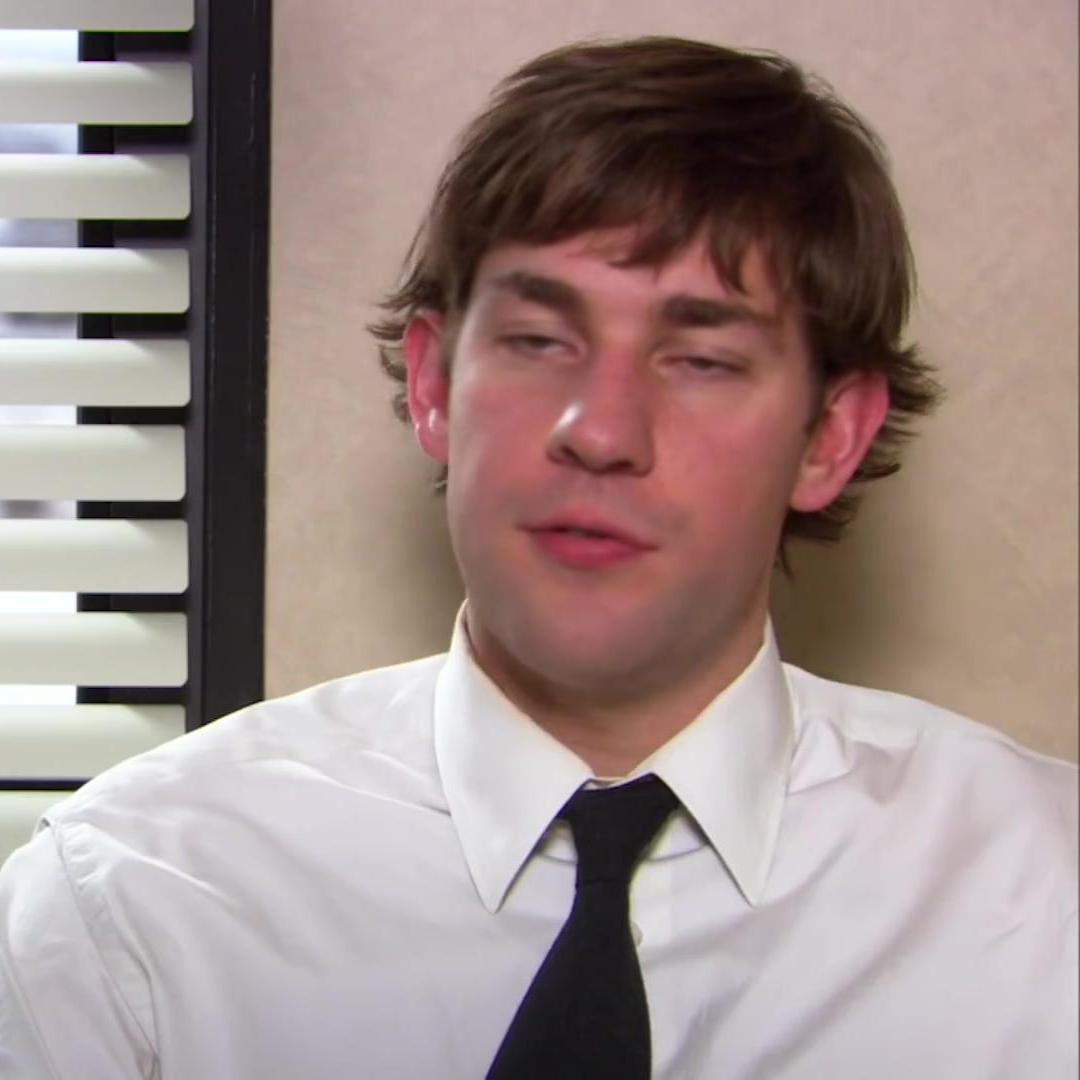}
     \includegraphics[width=\linewidth]{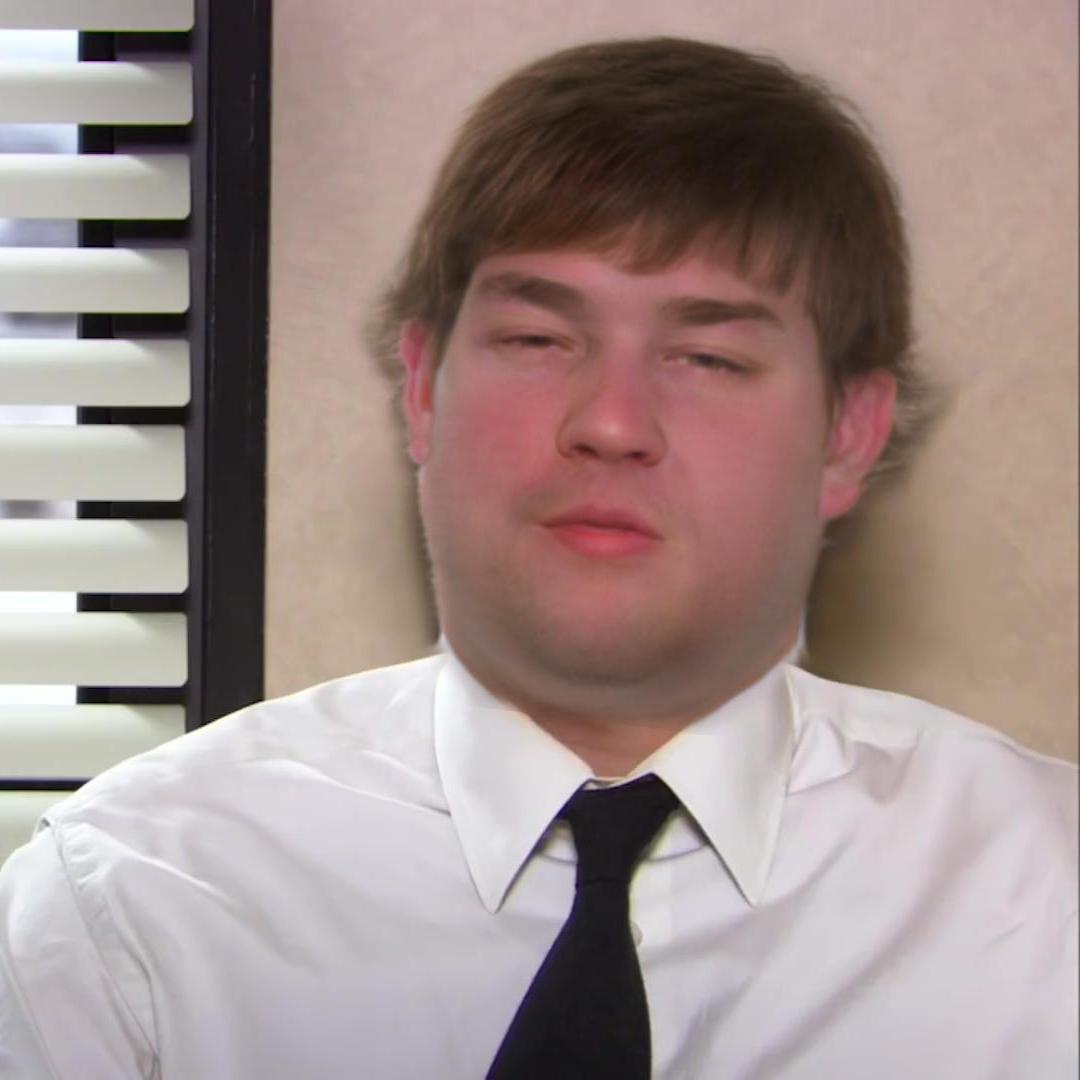}
     \includegraphics[width=\linewidth]{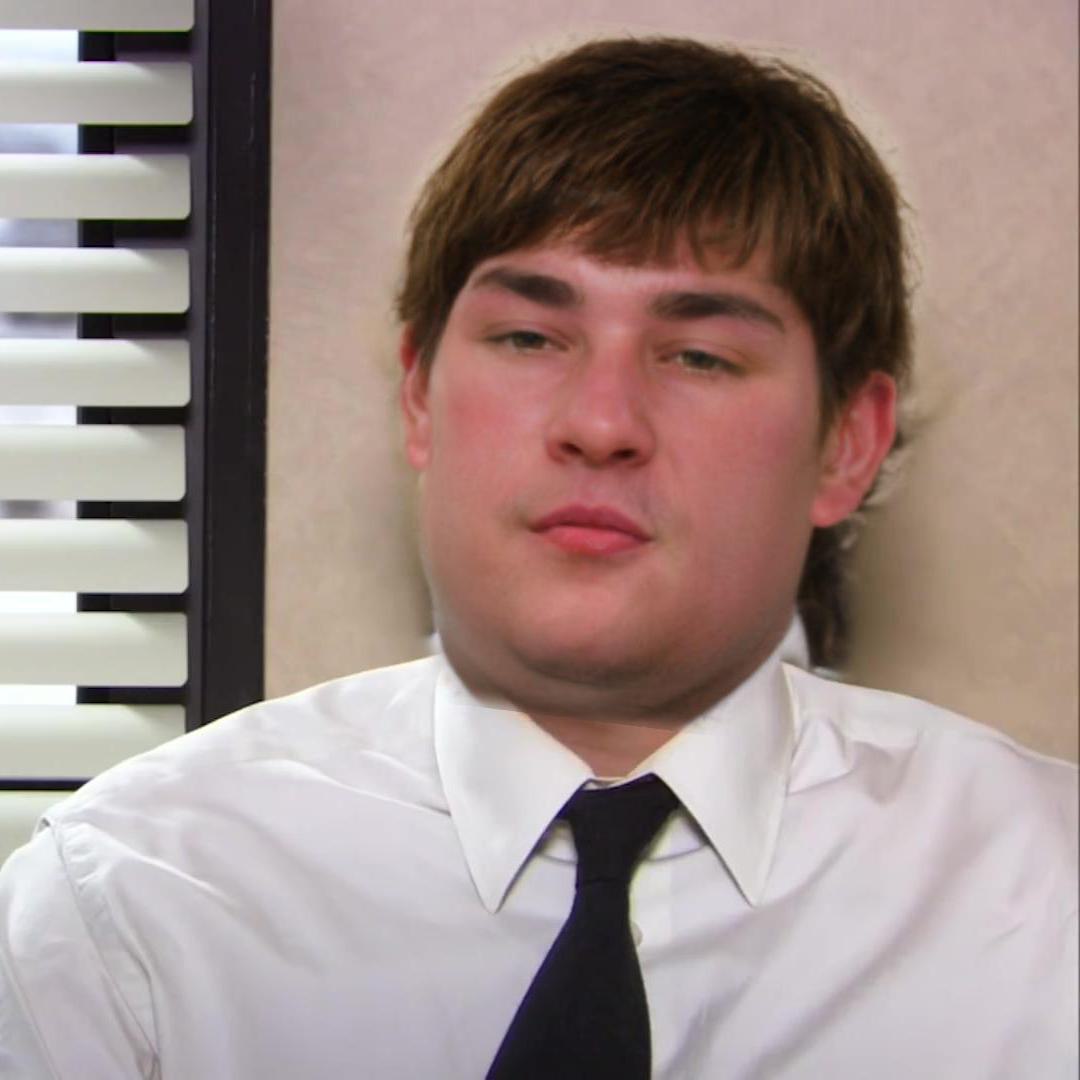}
     \includegraphics[width=\linewidth]{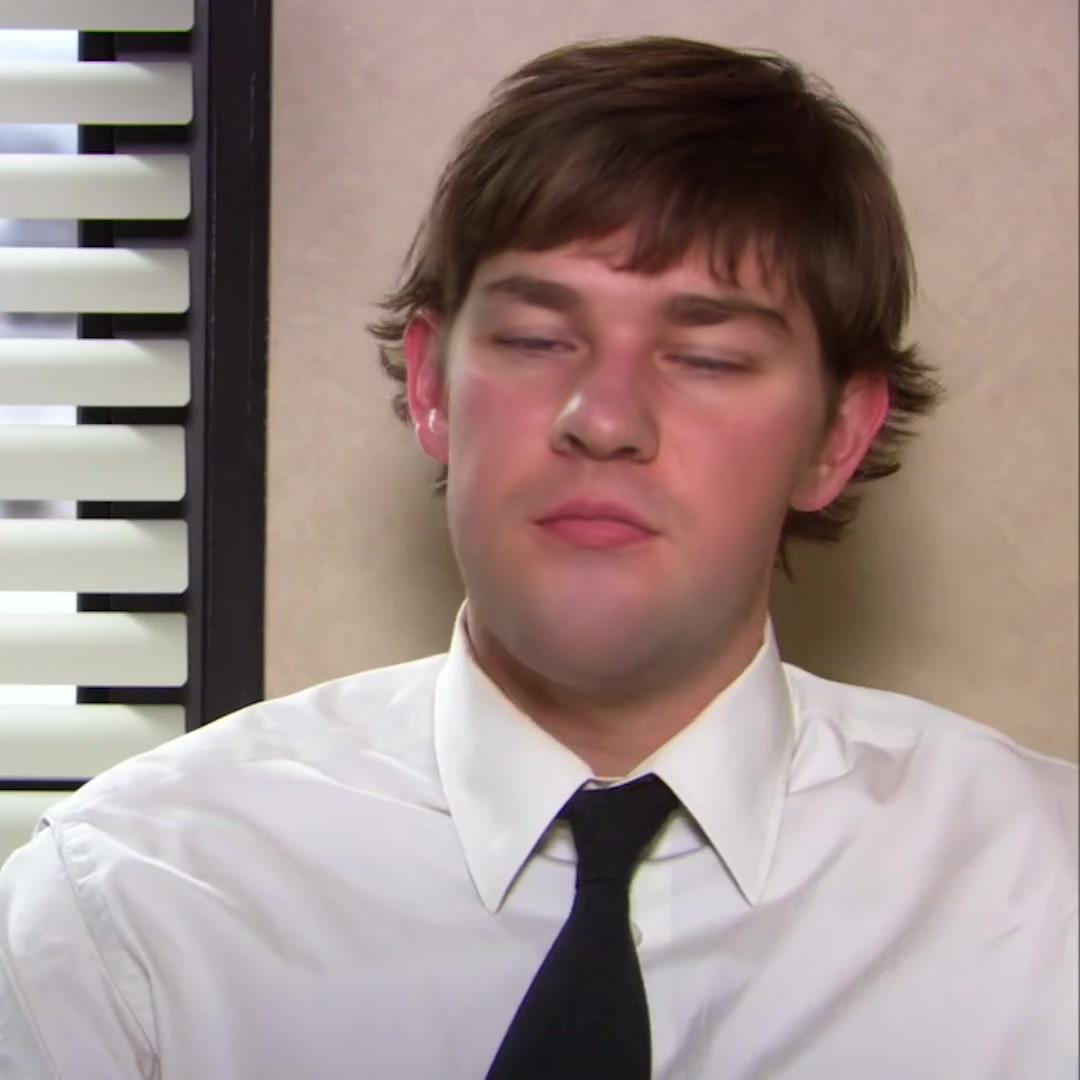}
     \includegraphics[width=\linewidth]{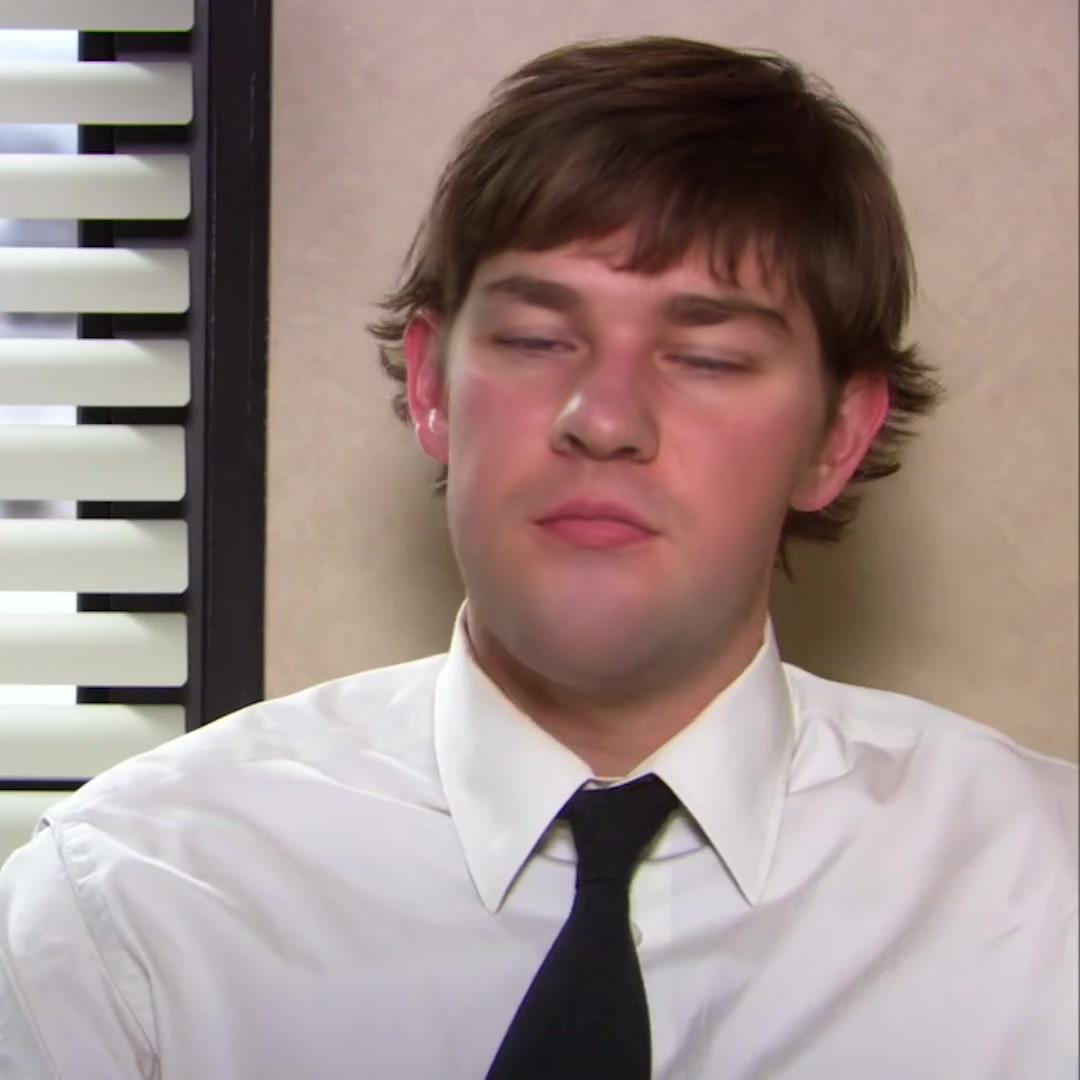}
     \includegraphics[width=\linewidth]{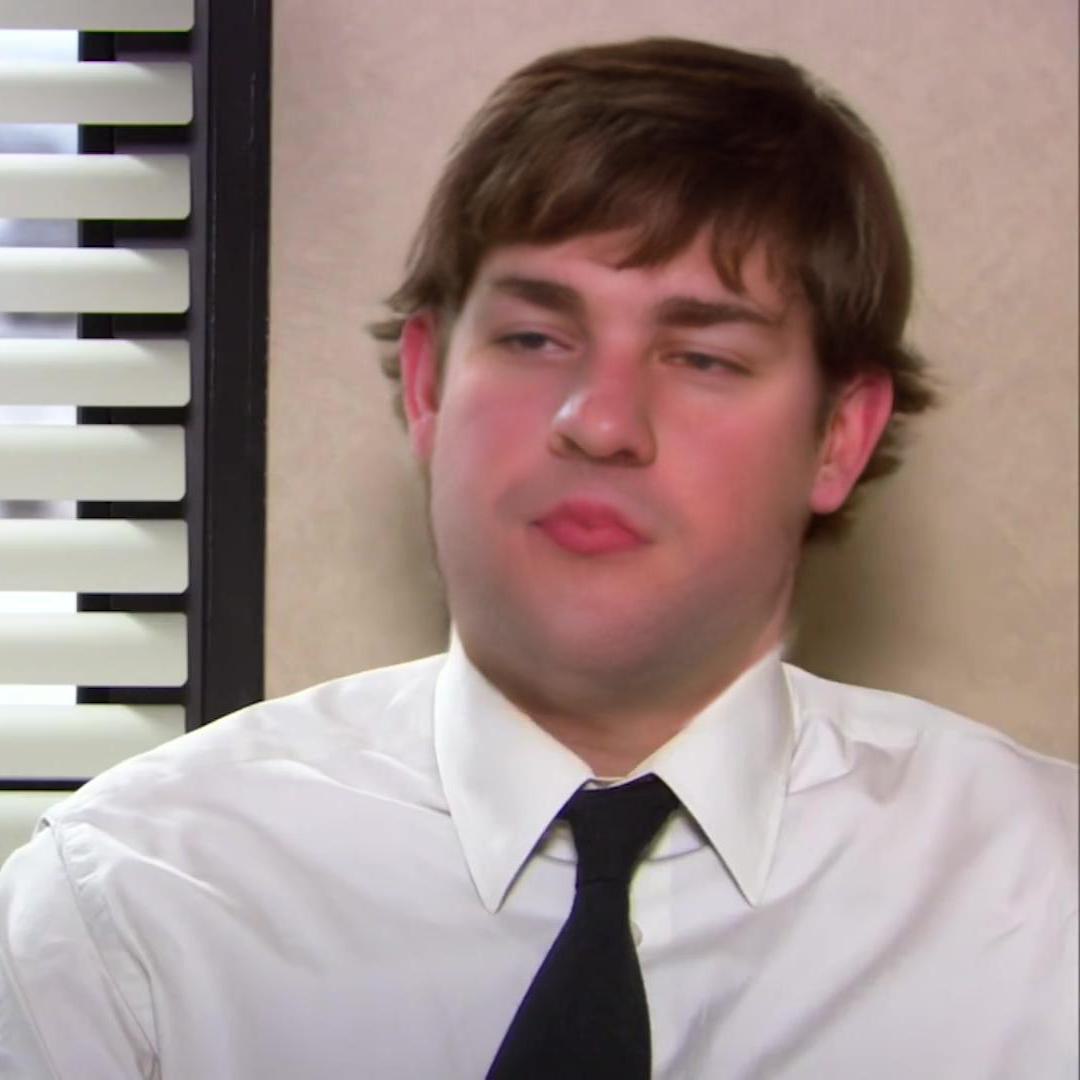}
     \end{minipage}
     }
    \hspace{-2.8mm}
    \subfloat[\texttt{+Chubby}]{
     \begin{minipage}{0.105\linewidth}
     \includegraphics[width=\linewidth]{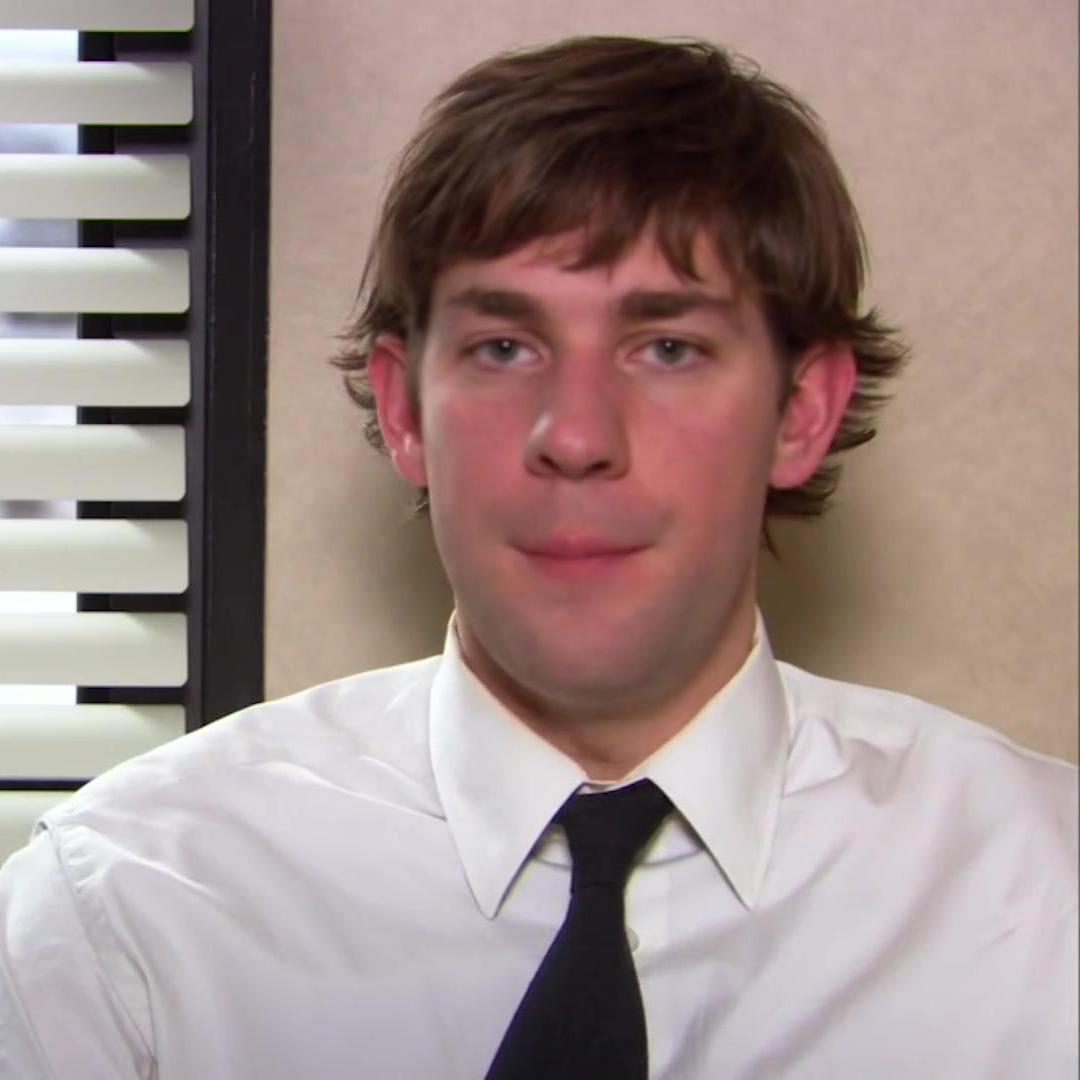}
     \includegraphics[width=\linewidth]{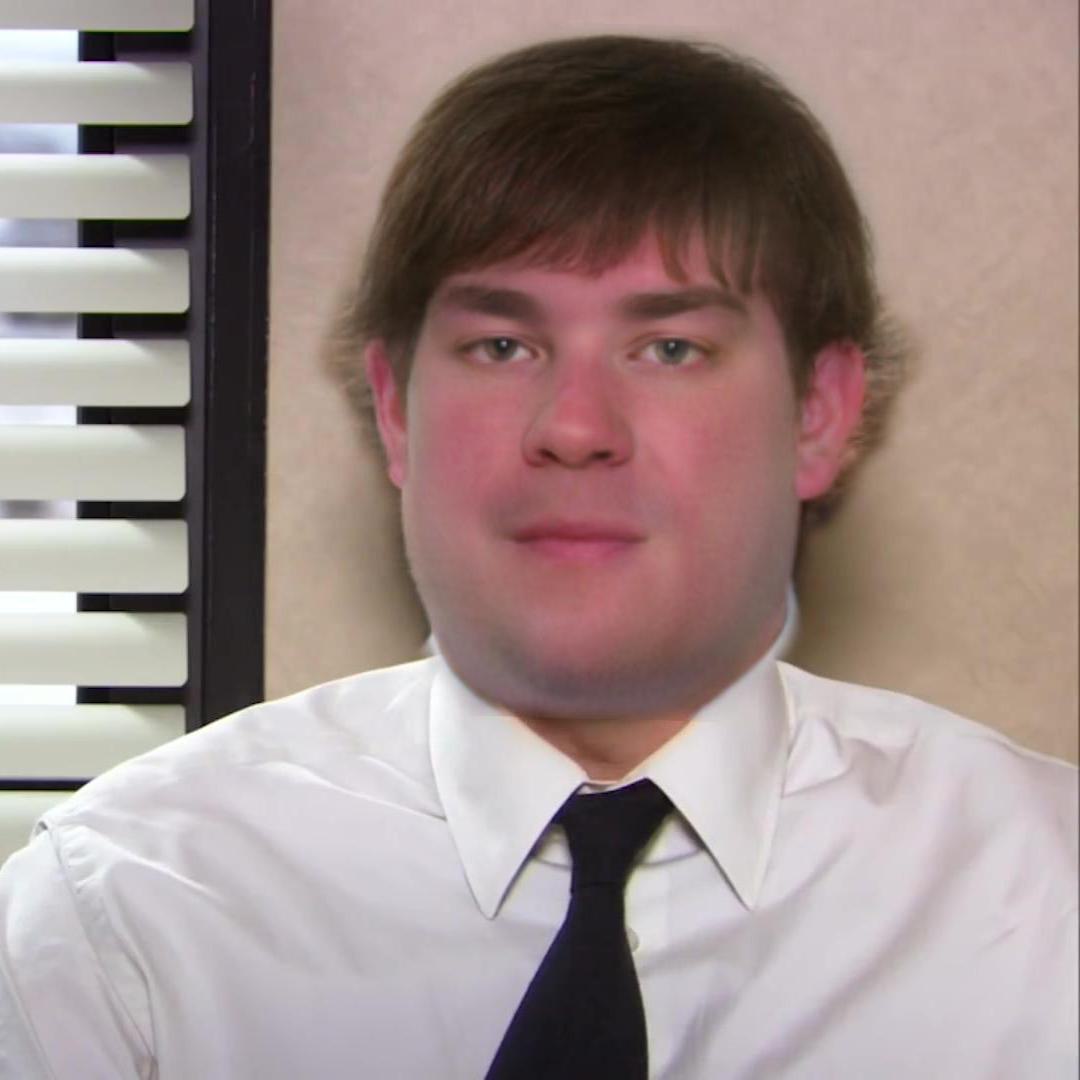}
     \includegraphics[width=\linewidth]{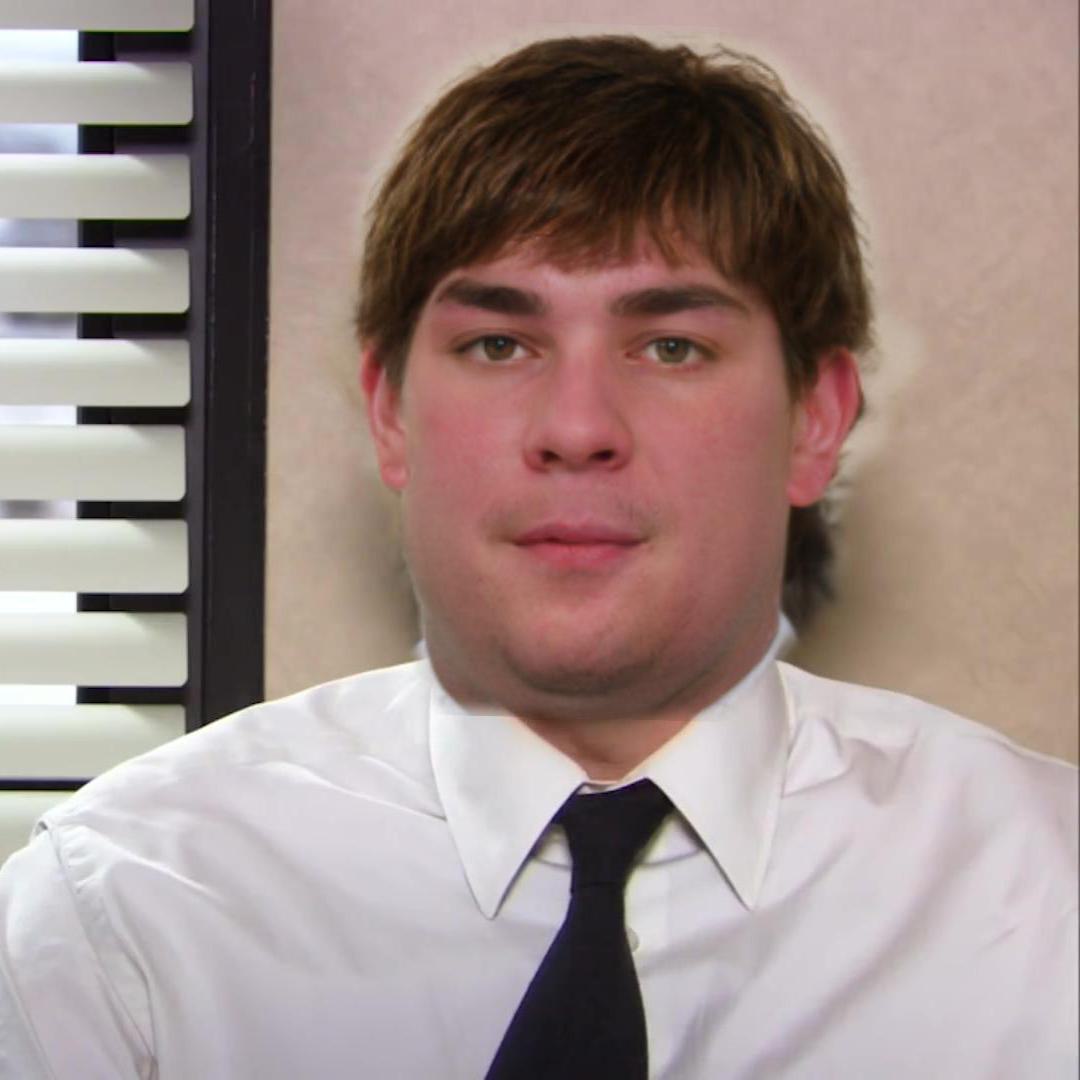}
     \includegraphics[width=\linewidth]{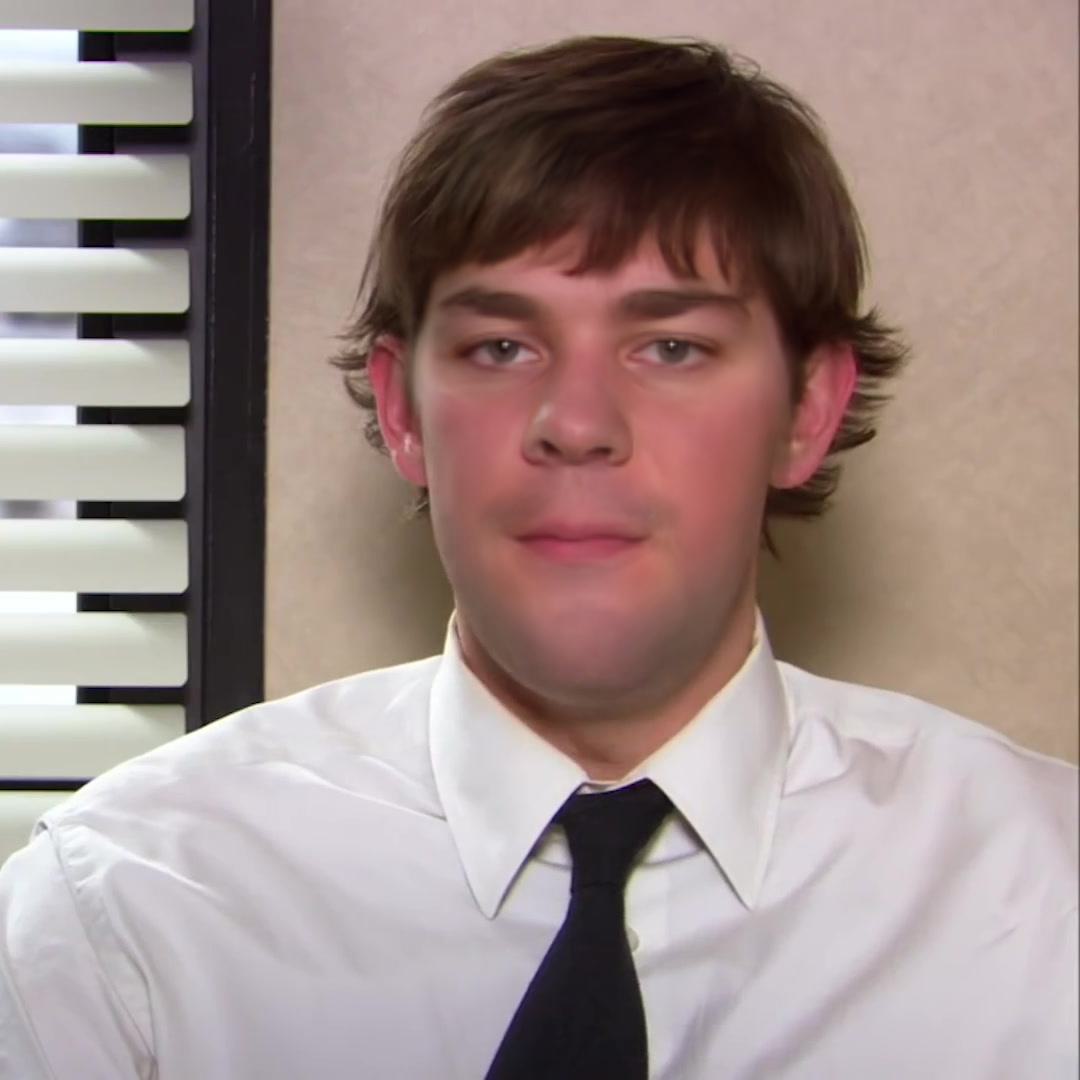}
     \includegraphics[width=\linewidth]{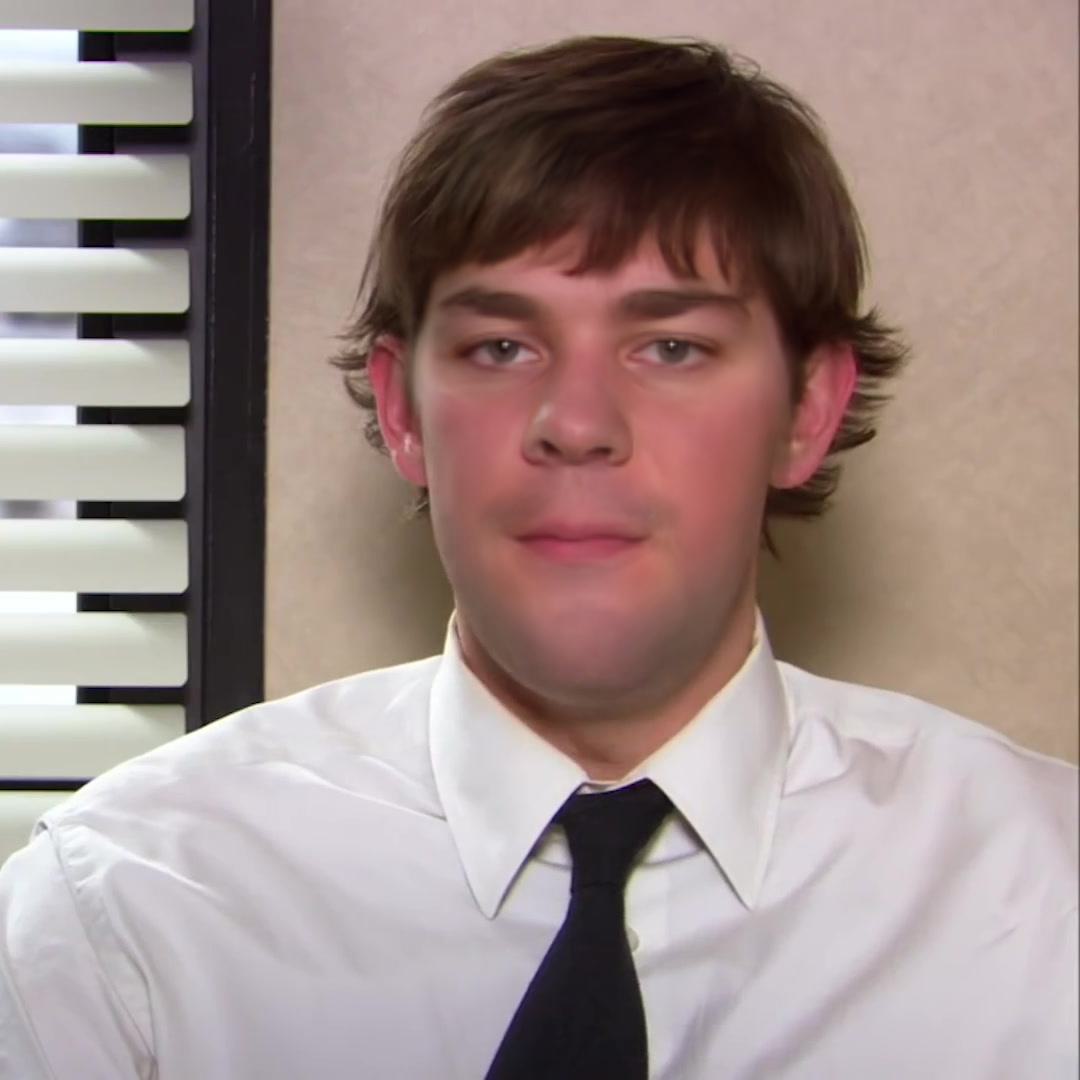}
     \includegraphics[width=\linewidth]{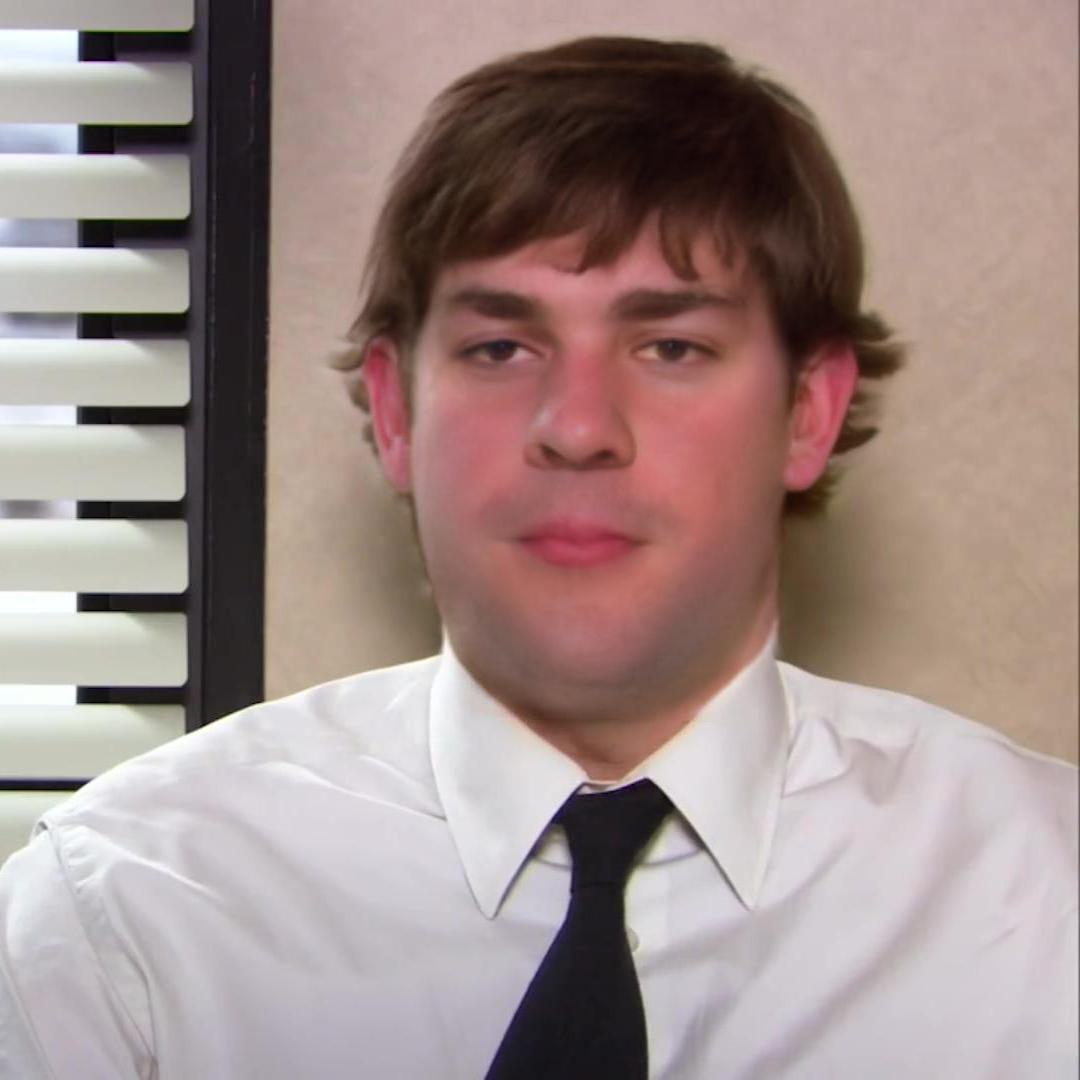}
     \end{minipage}
     }
    \hspace{-2.8mm}
    \subfloat[]{
     \begin{minipage}{0.105\linewidth}
     \includegraphics[width=\linewidth]{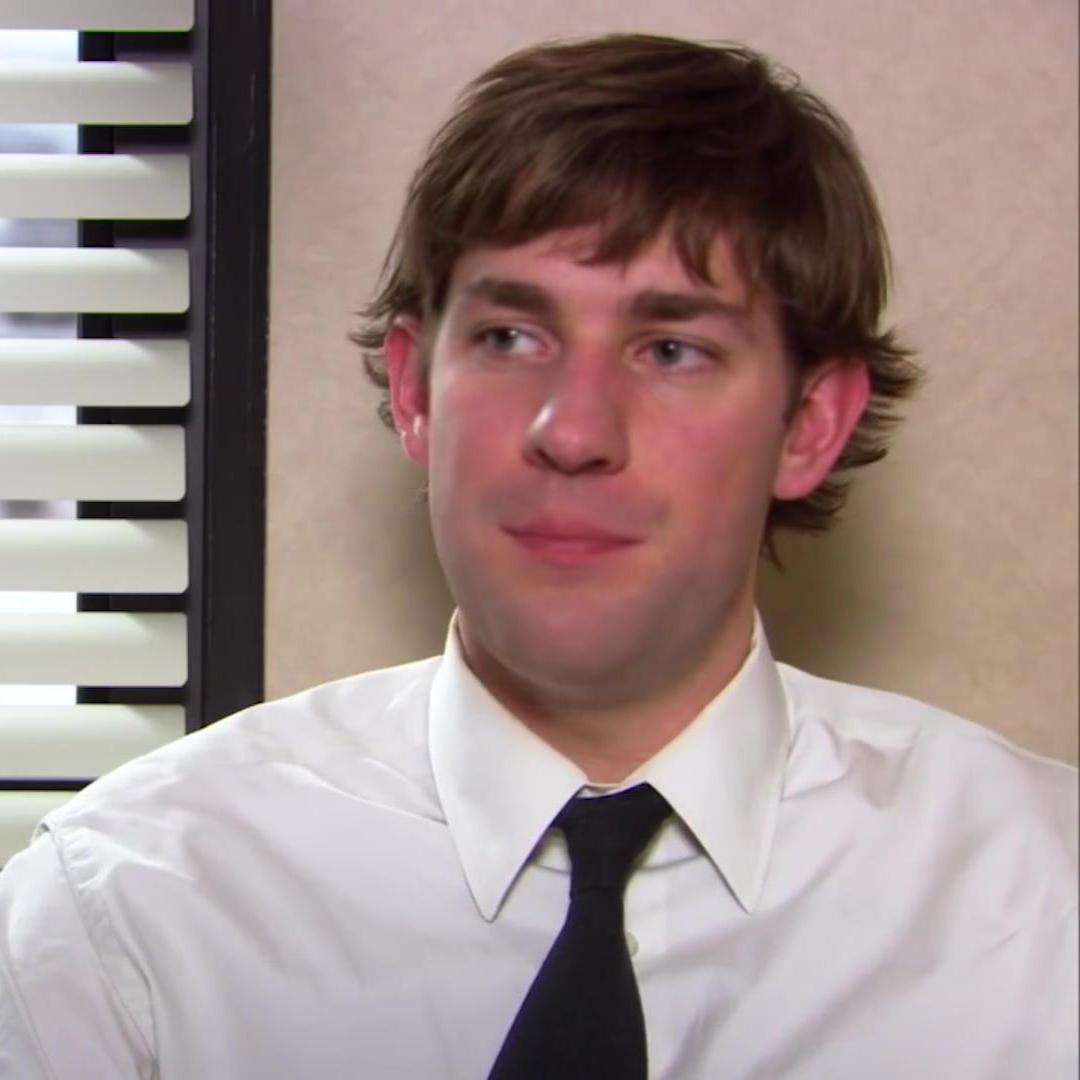}
     \includegraphics[width=\linewidth]{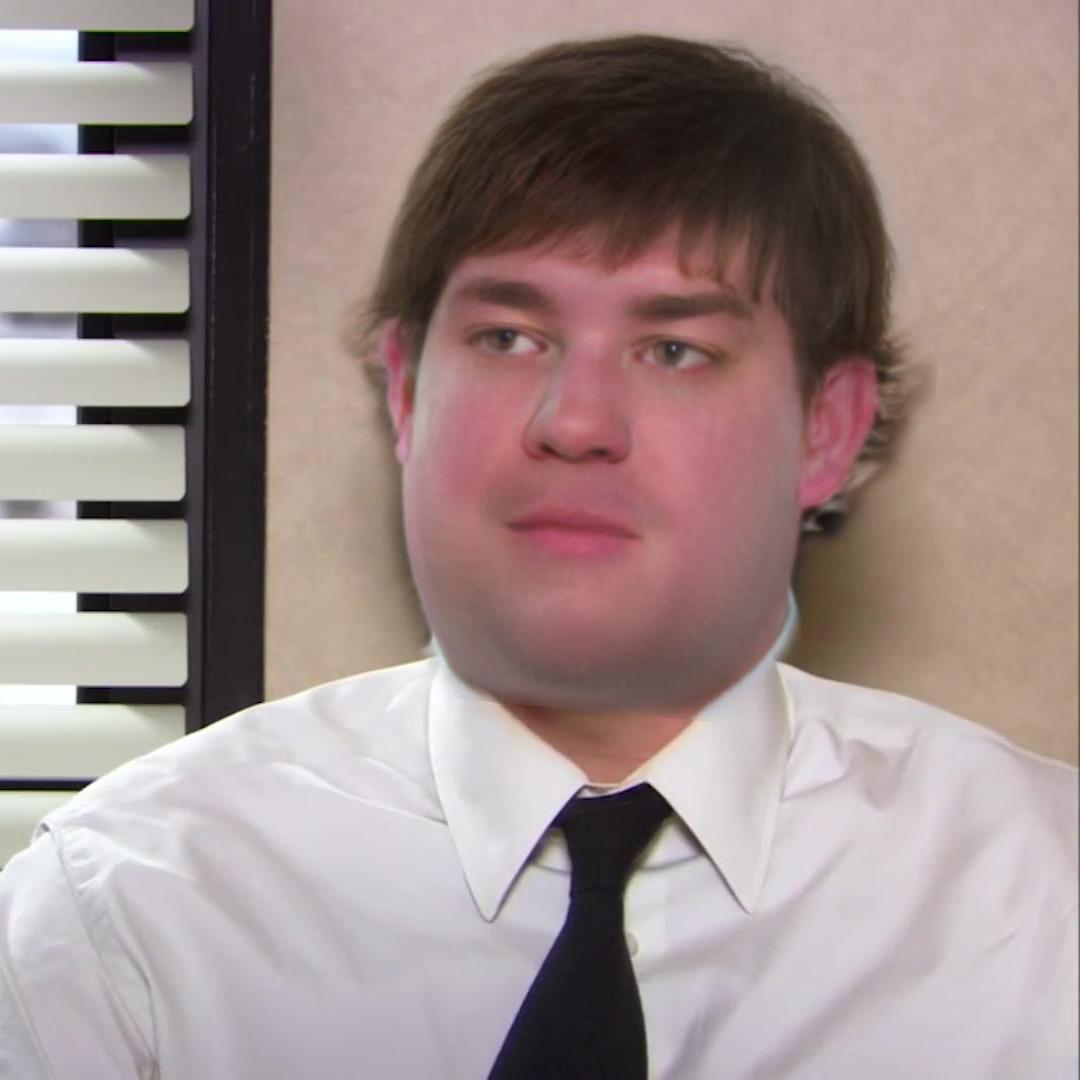}
     \includegraphics[width=\linewidth]{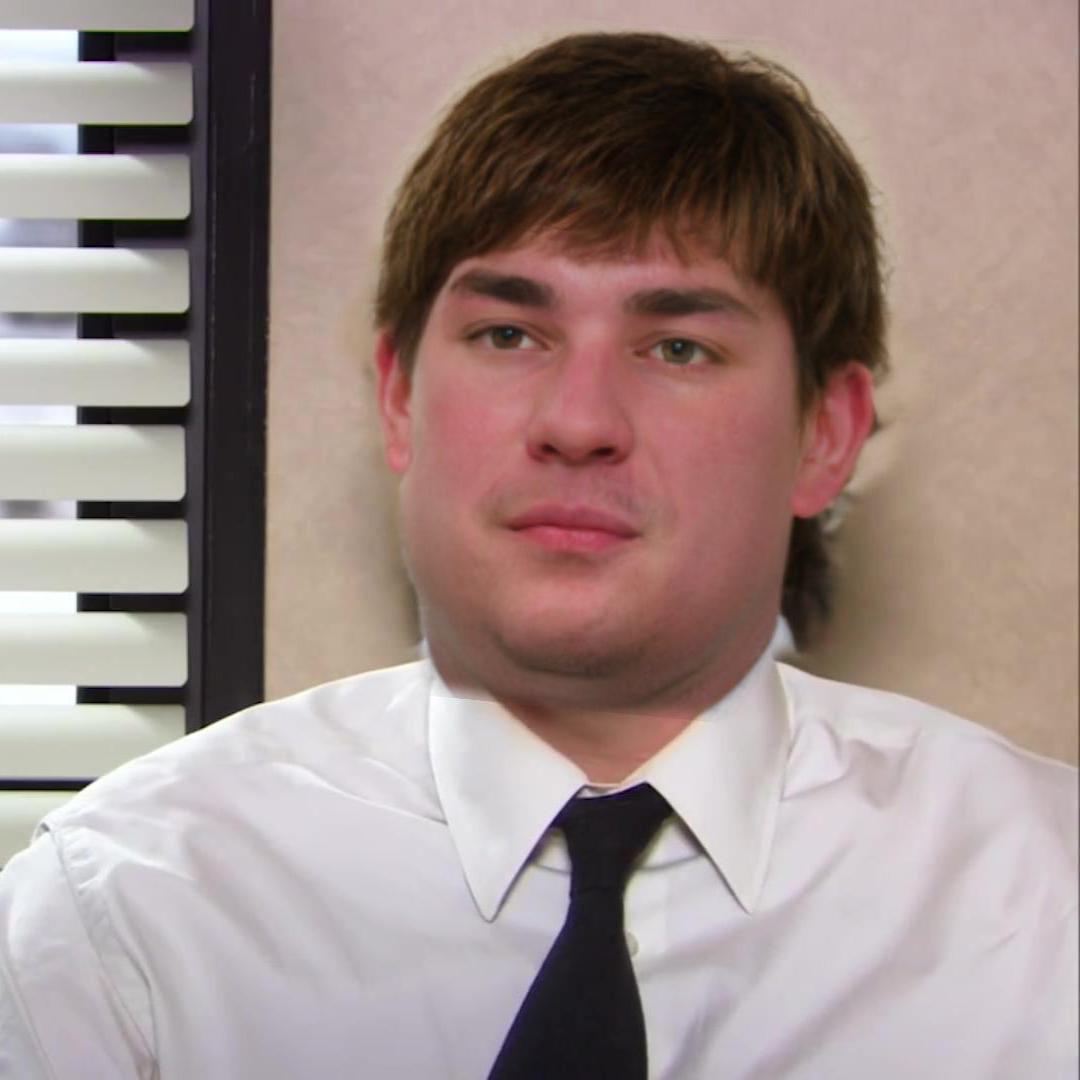}
     \includegraphics[width=\linewidth]{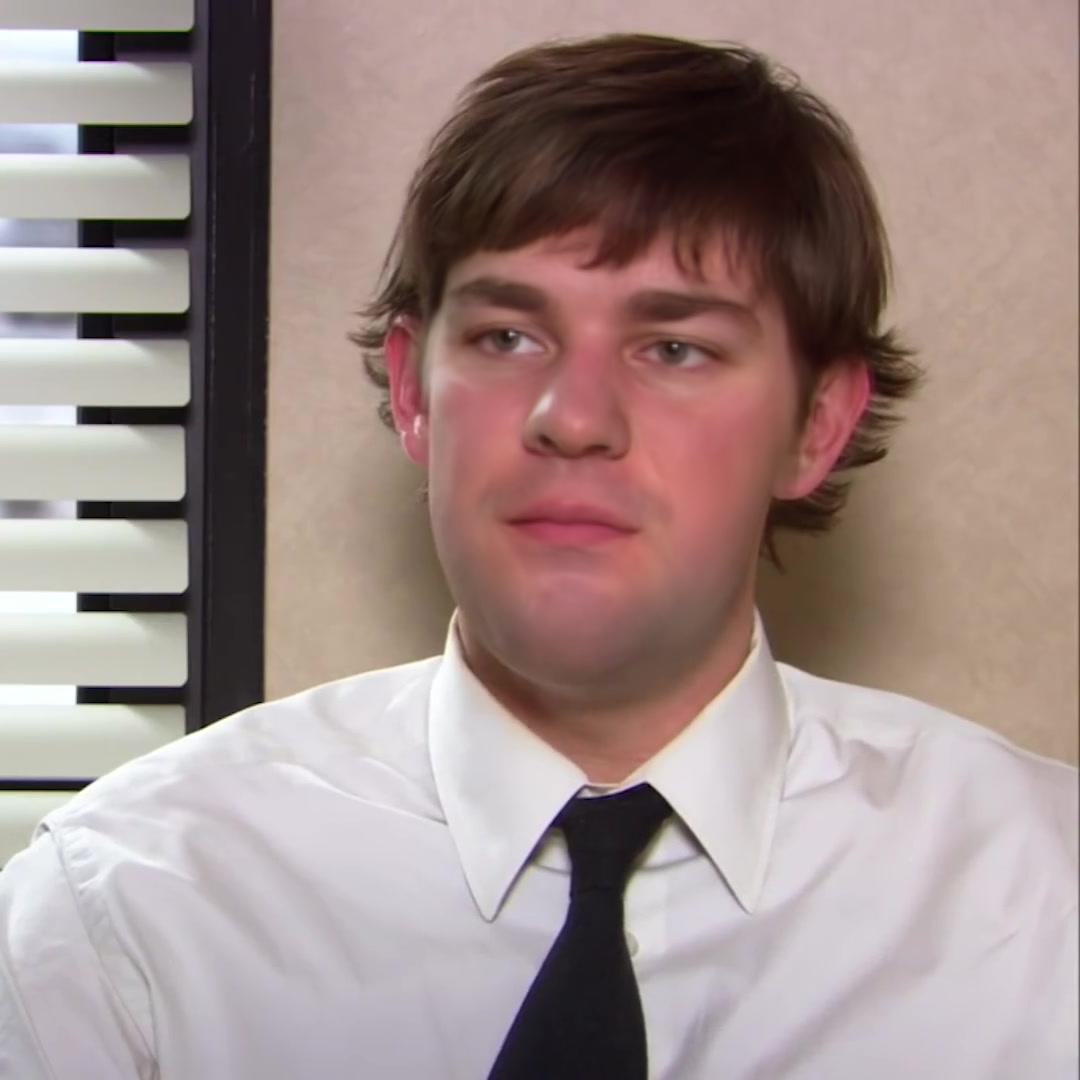}
     \includegraphics[width=\linewidth]{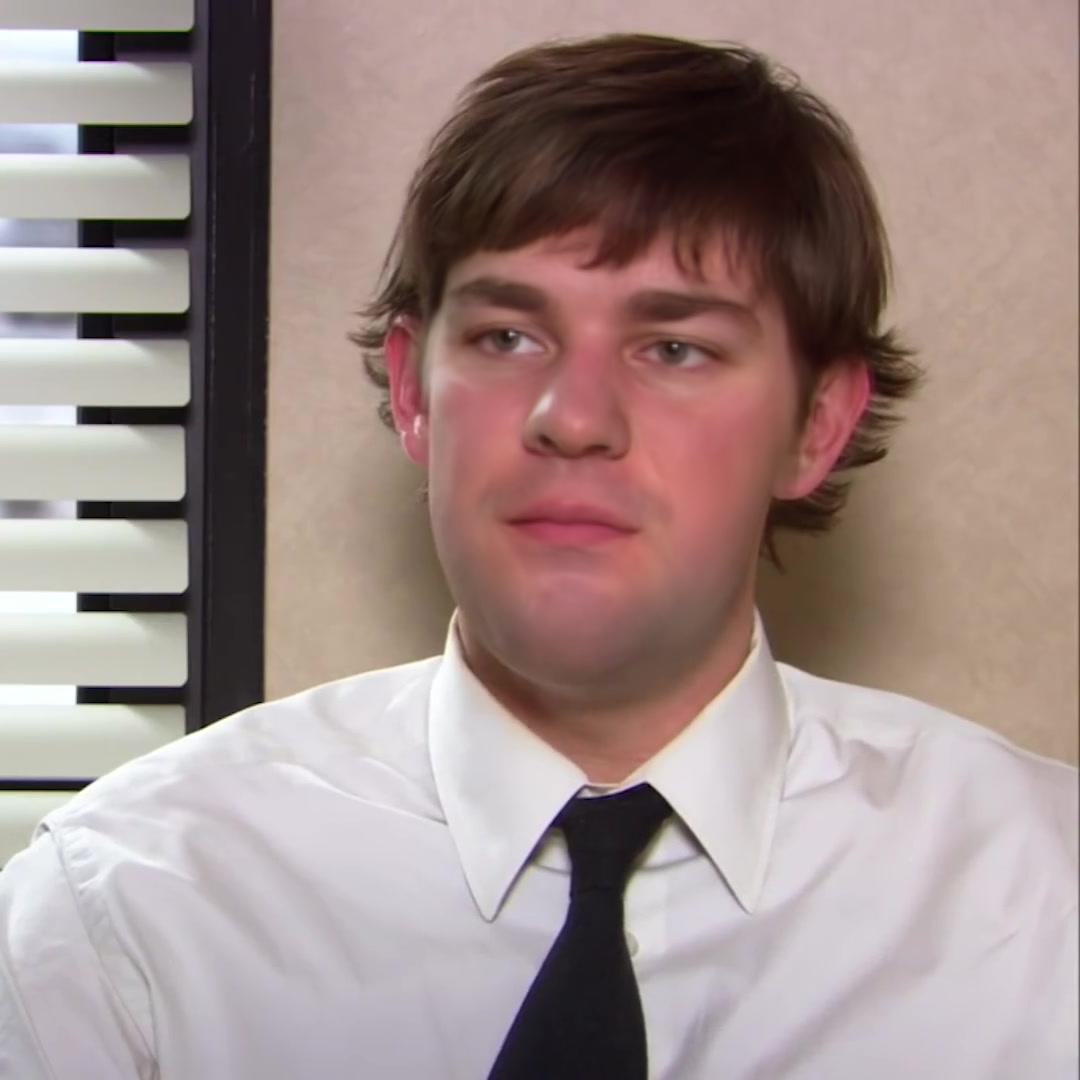}
     \includegraphics[width=\linewidth]{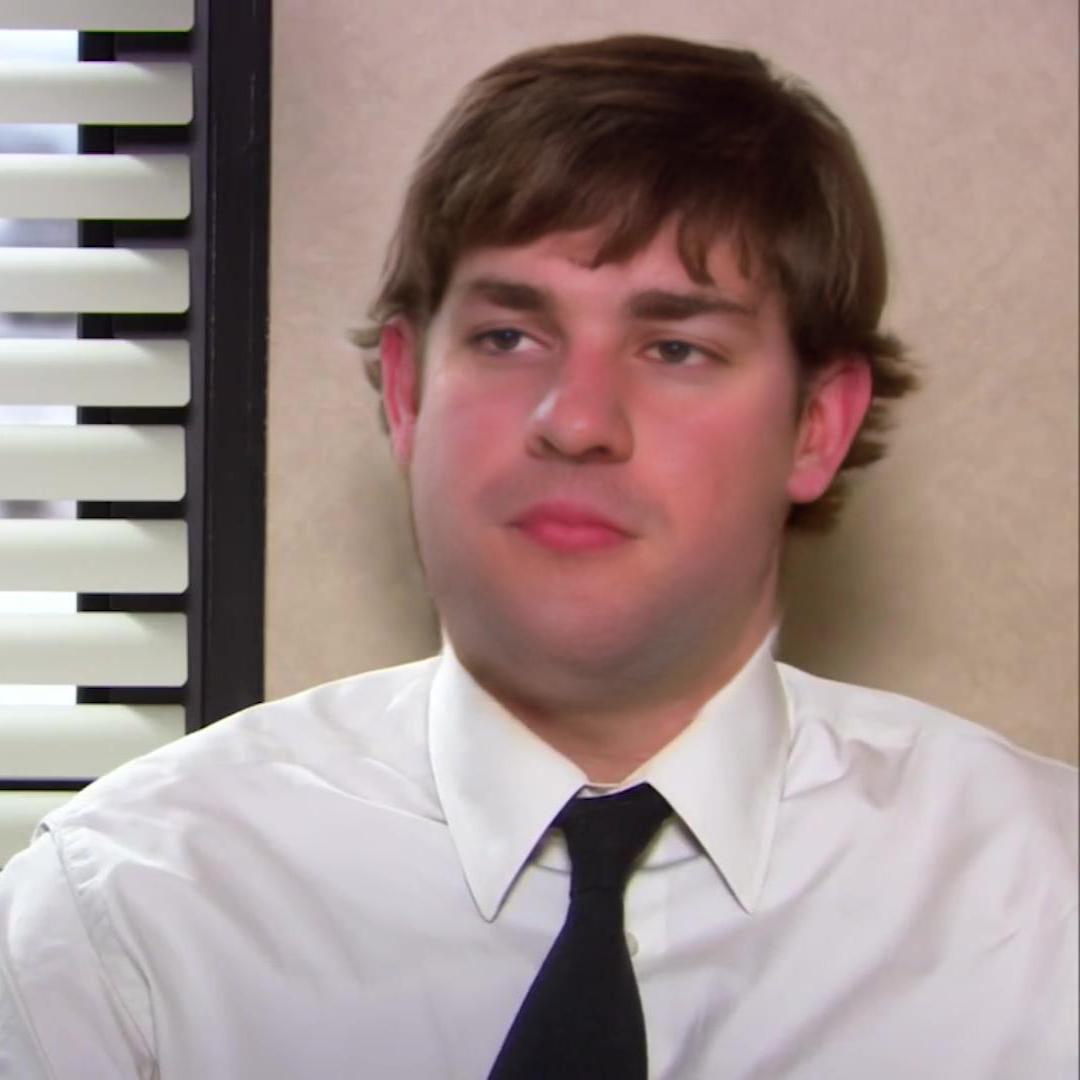}
     \end{minipage}
     }
    \hspace{-1.8mm}
    \subfloat[]{
     \begin{minipage}{0.105\linewidth}
     \includegraphics[width=\linewidth]{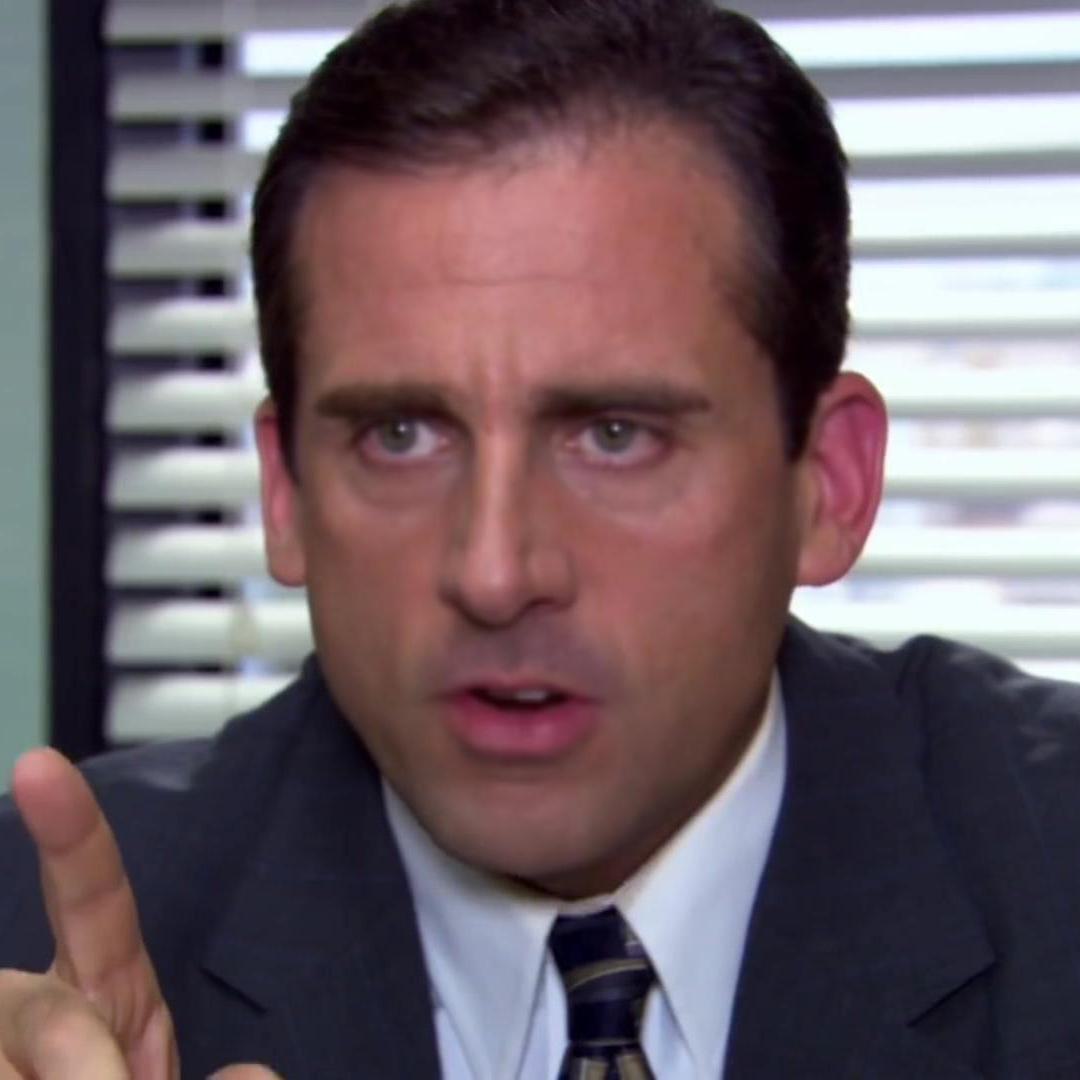}
     \includegraphics[width=\linewidth]{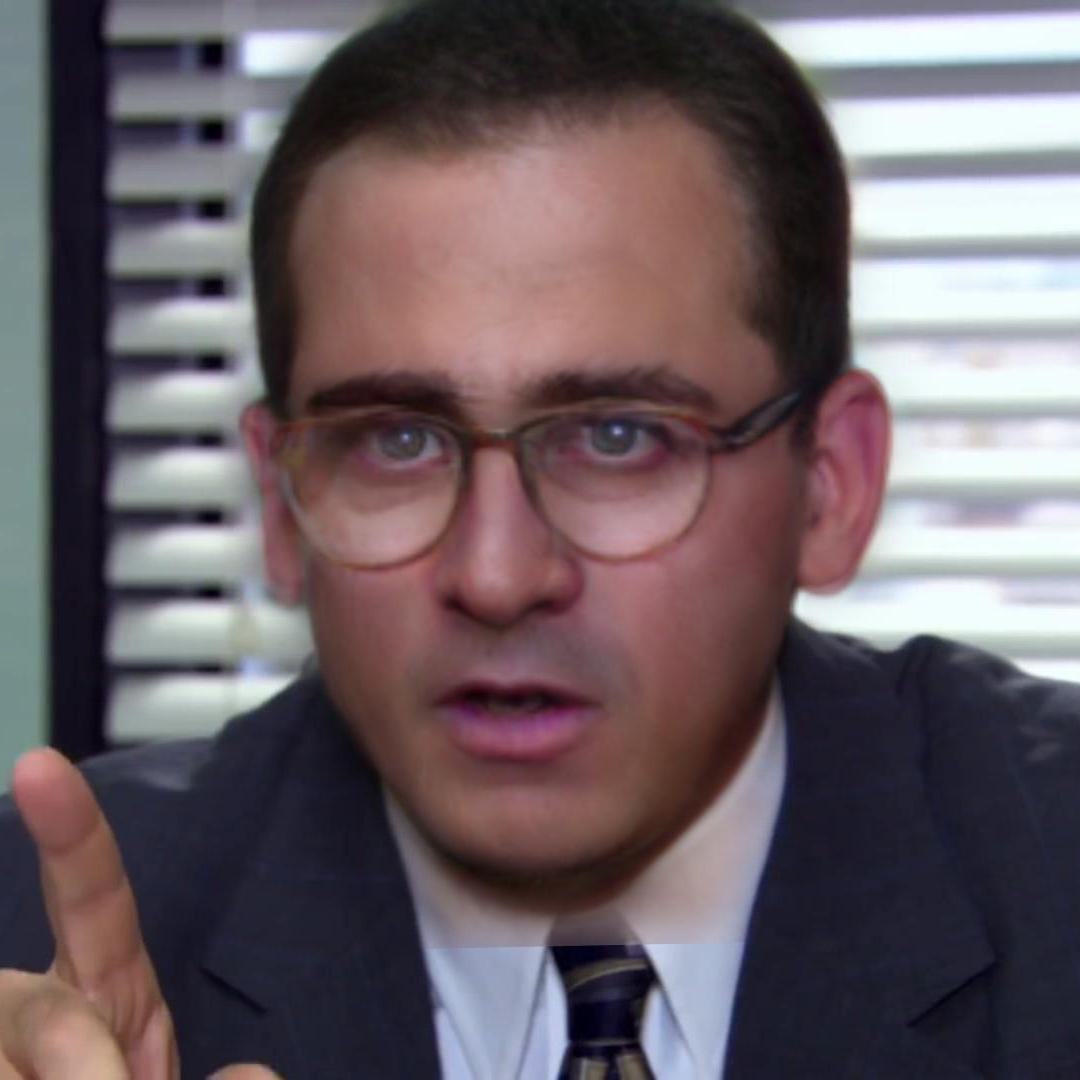}
     \includegraphics[width=\linewidth]{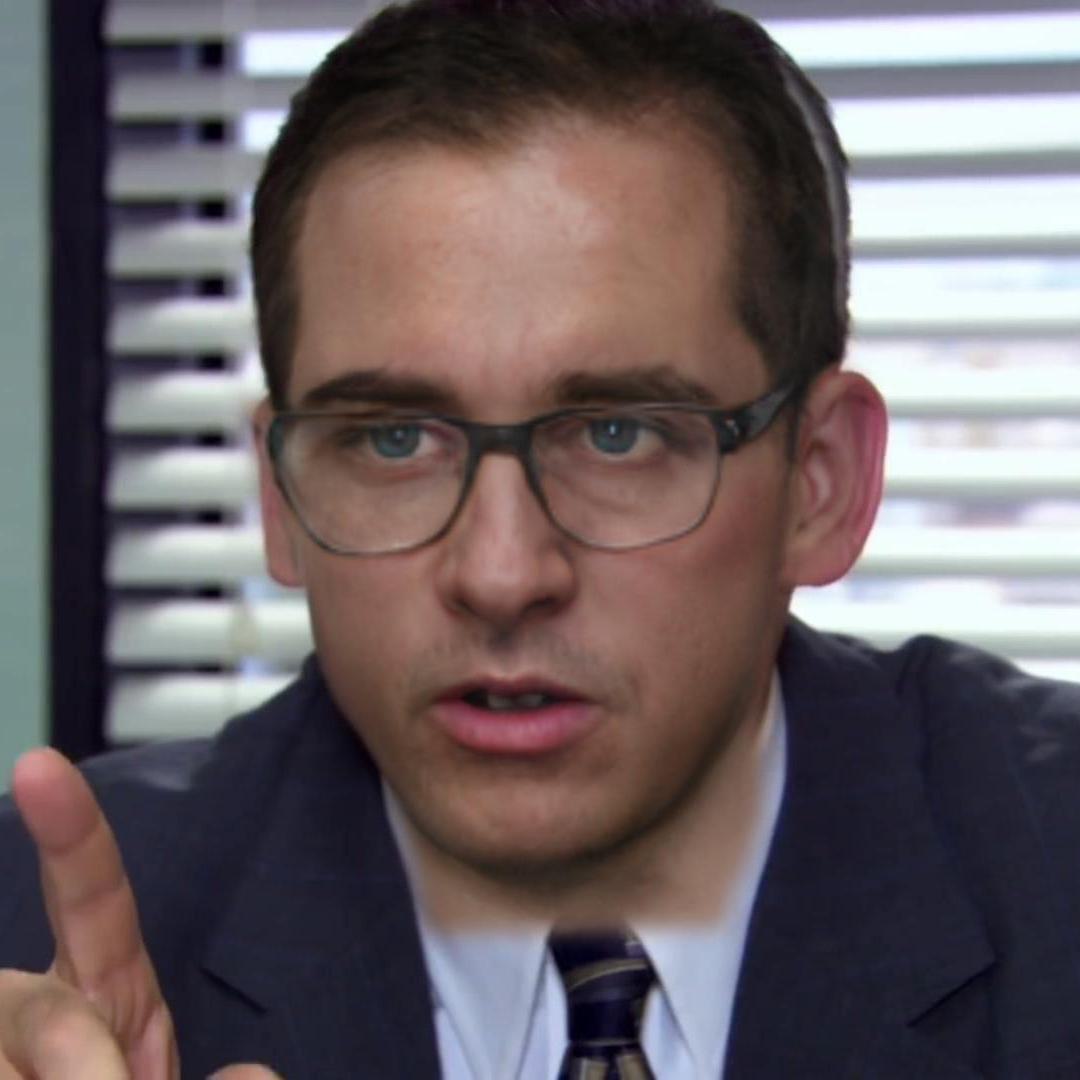}
     \includegraphics[width=\linewidth]{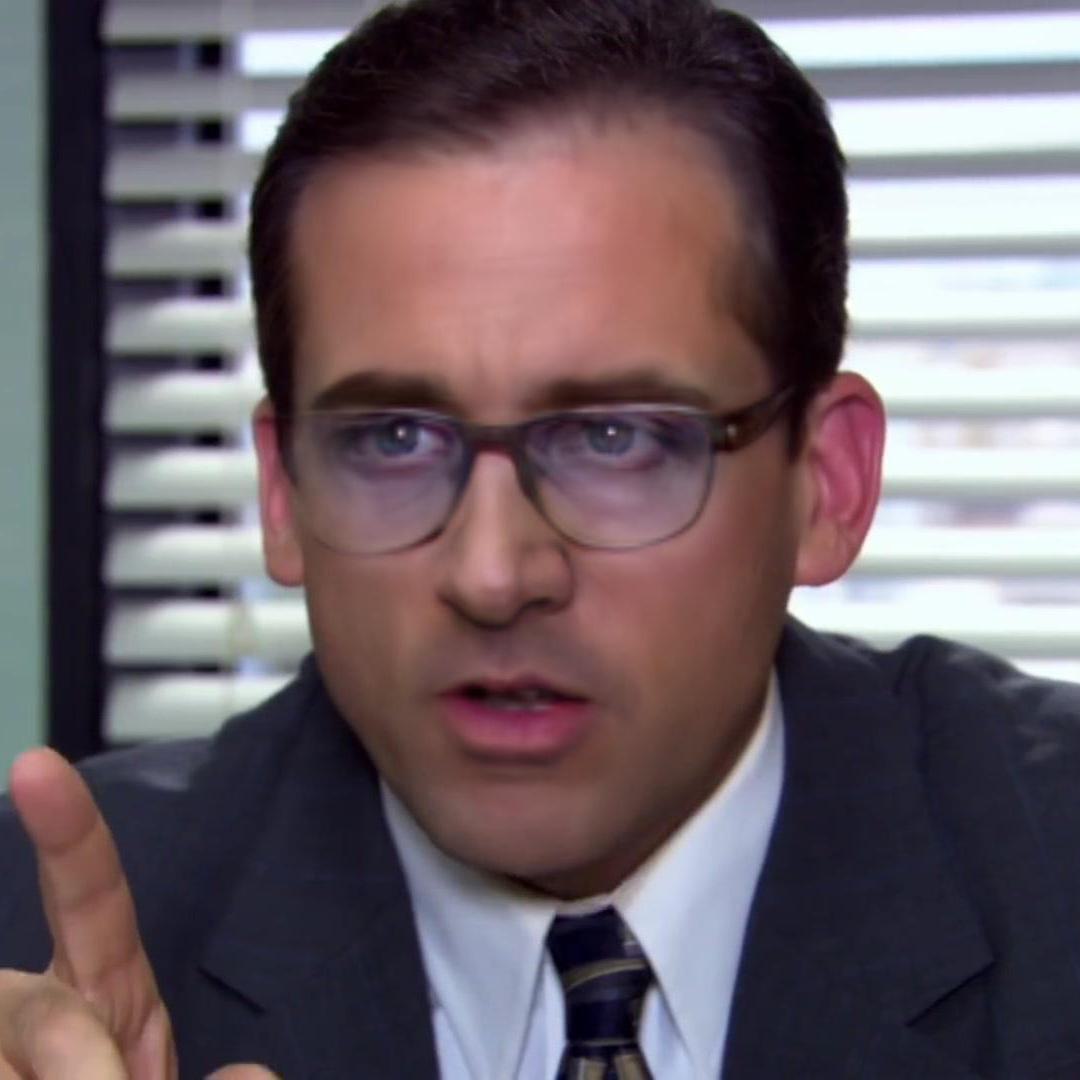}
     \includegraphics[width=\linewidth]{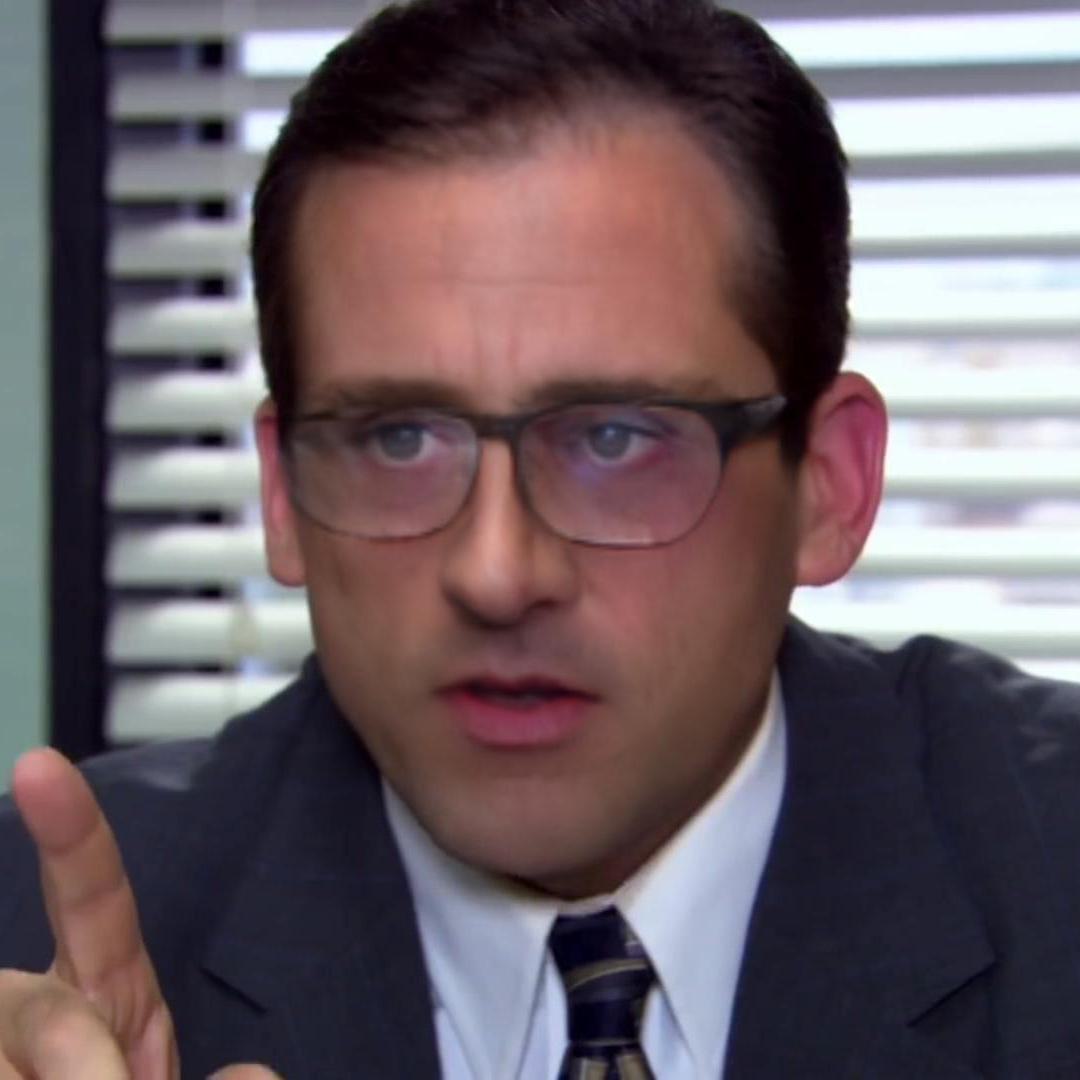}
     \includegraphics[width=\linewidth]{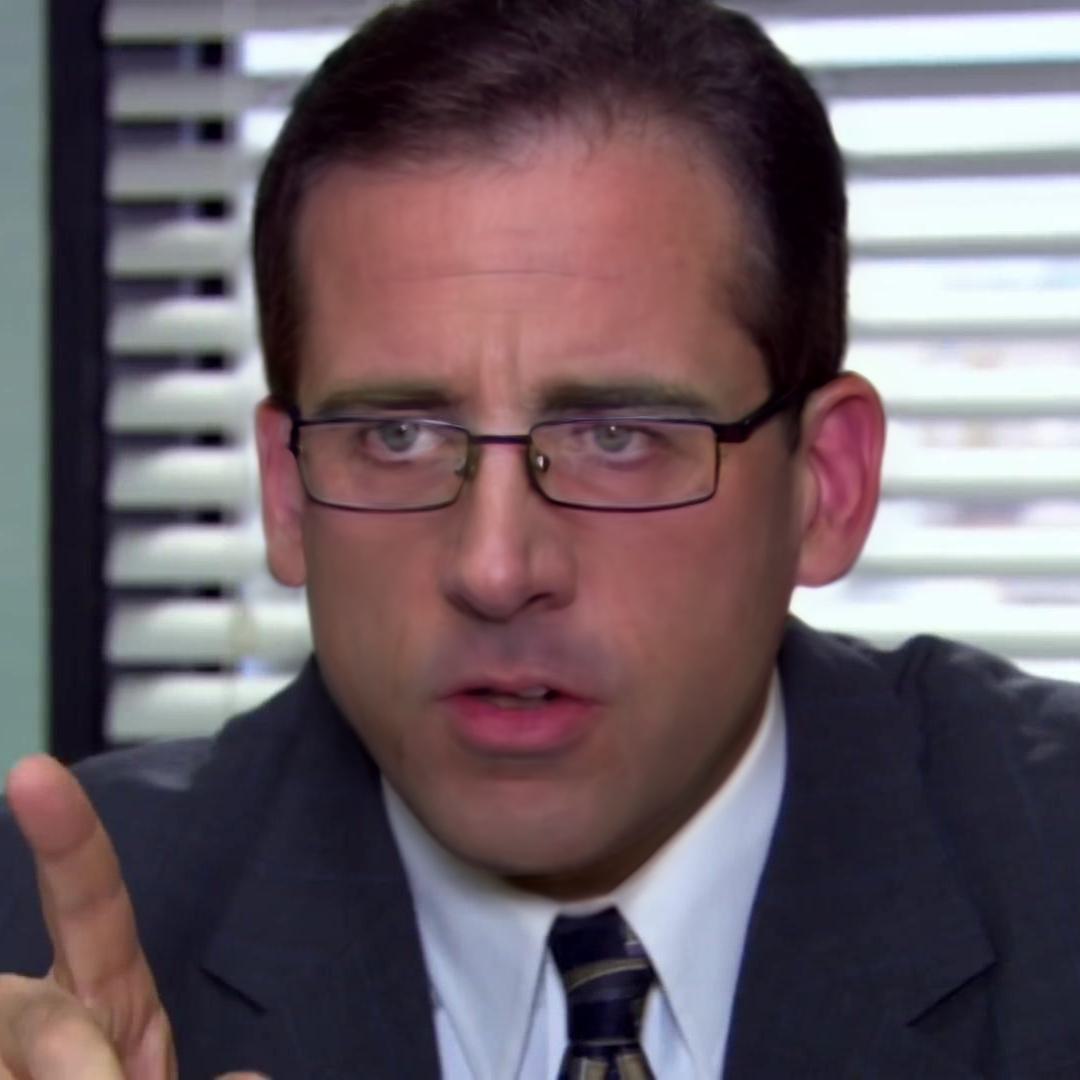}
     \end{minipage}
     }
    \hspace{-2.8mm}
    \subfloat[\texttt{+EyeGlasses}]{
     \begin{minipage}{0.105\linewidth}
     \includegraphics[width=\linewidth]{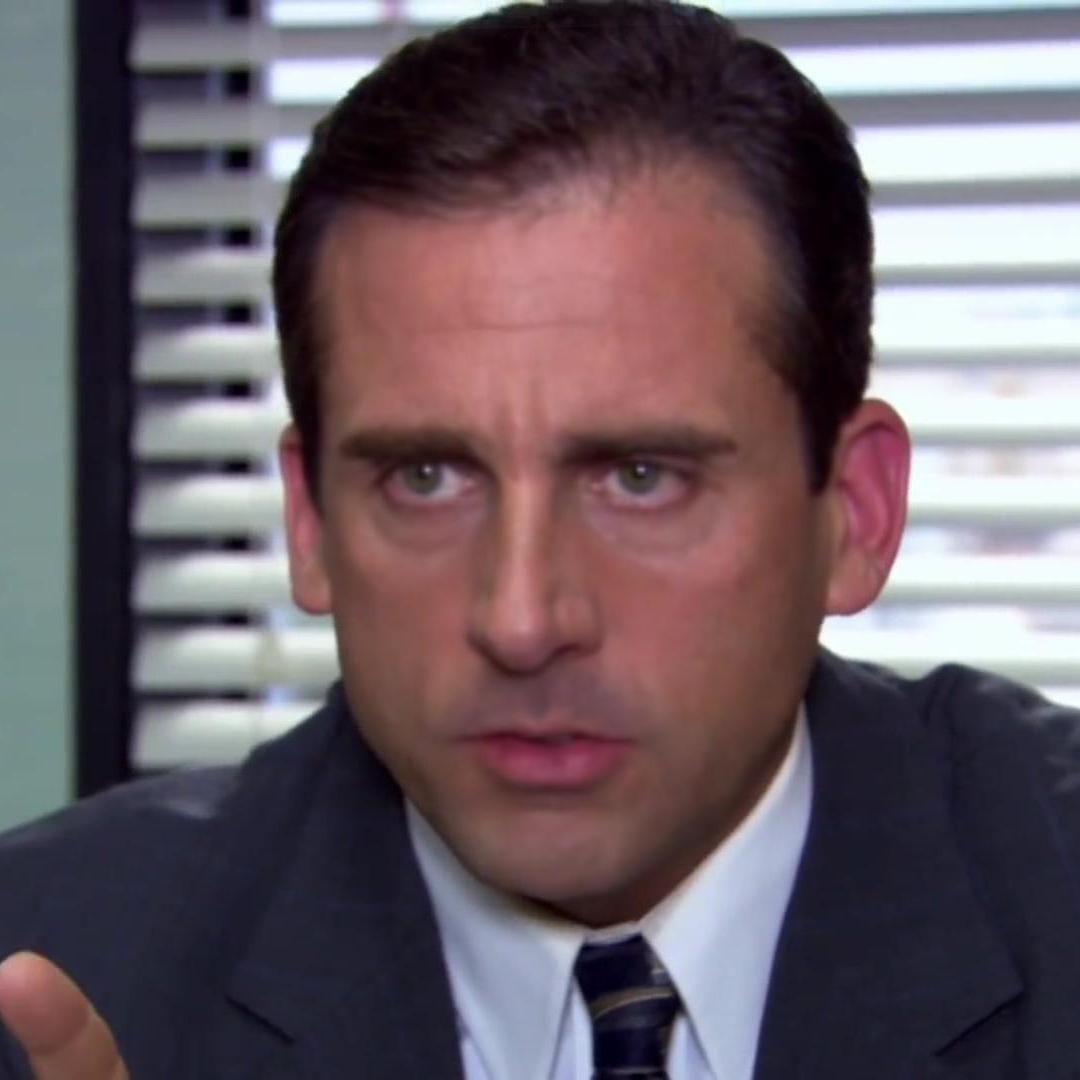}
     \includegraphics[width=\linewidth]{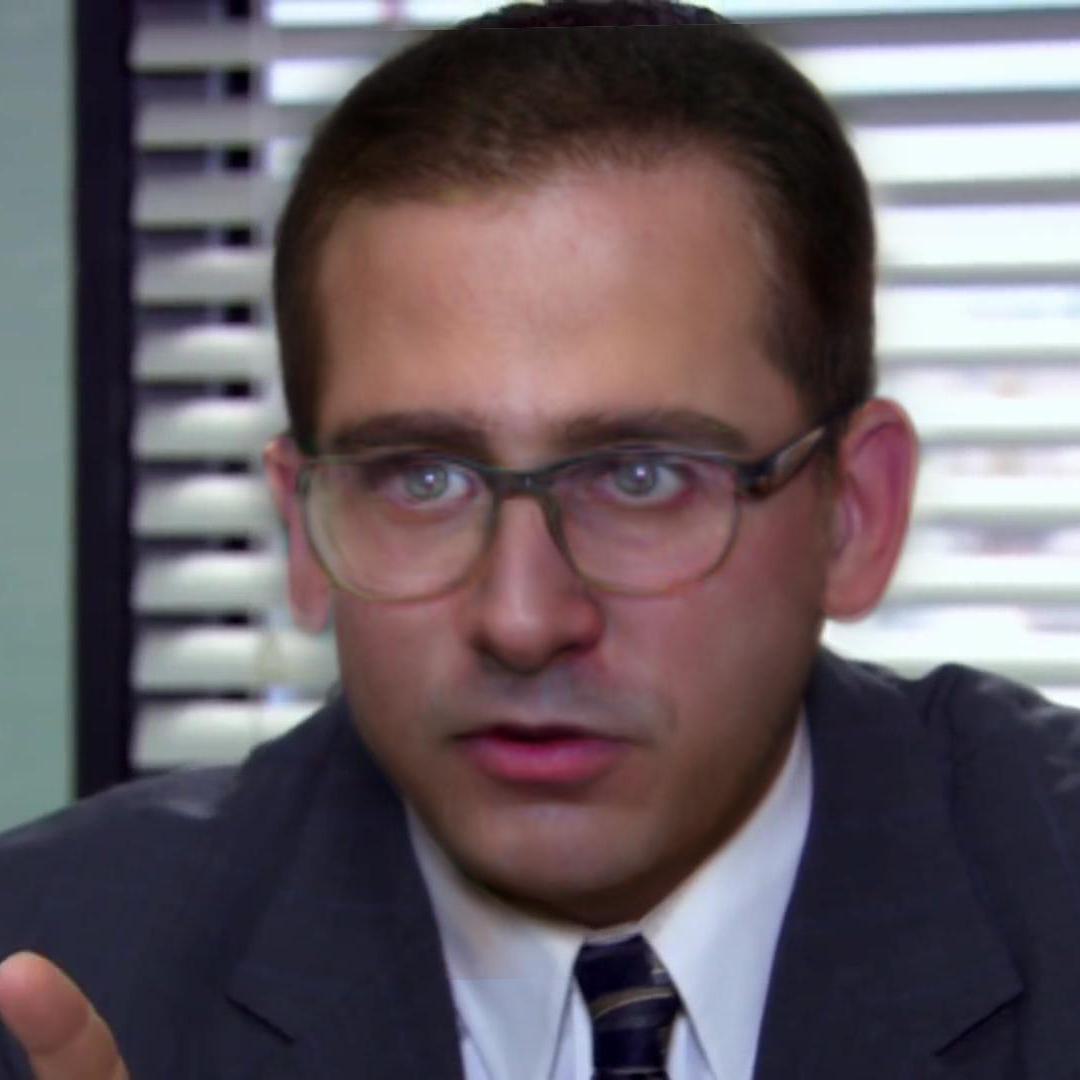}
     \includegraphics[width=\linewidth]{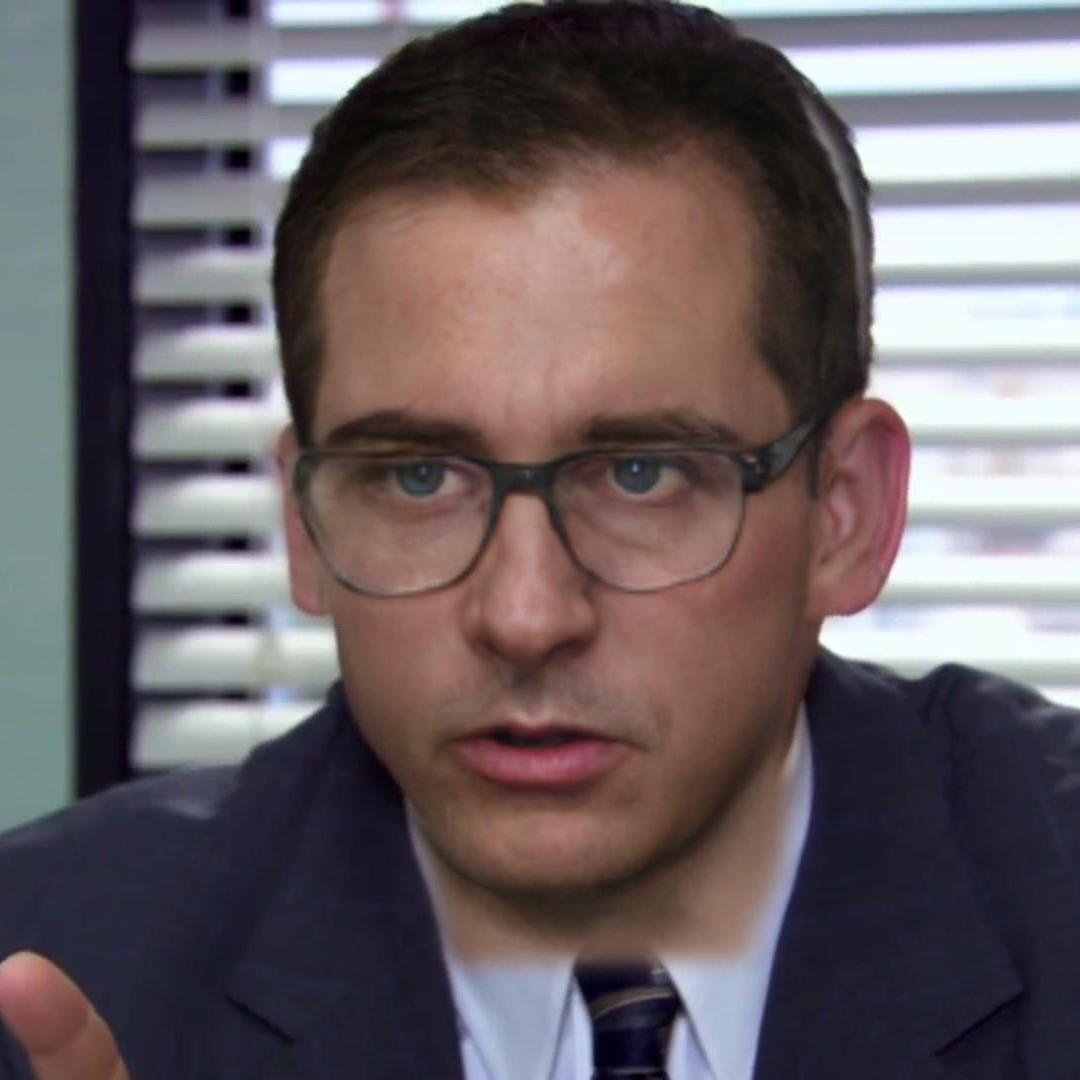}
     \includegraphics[width=\linewidth]{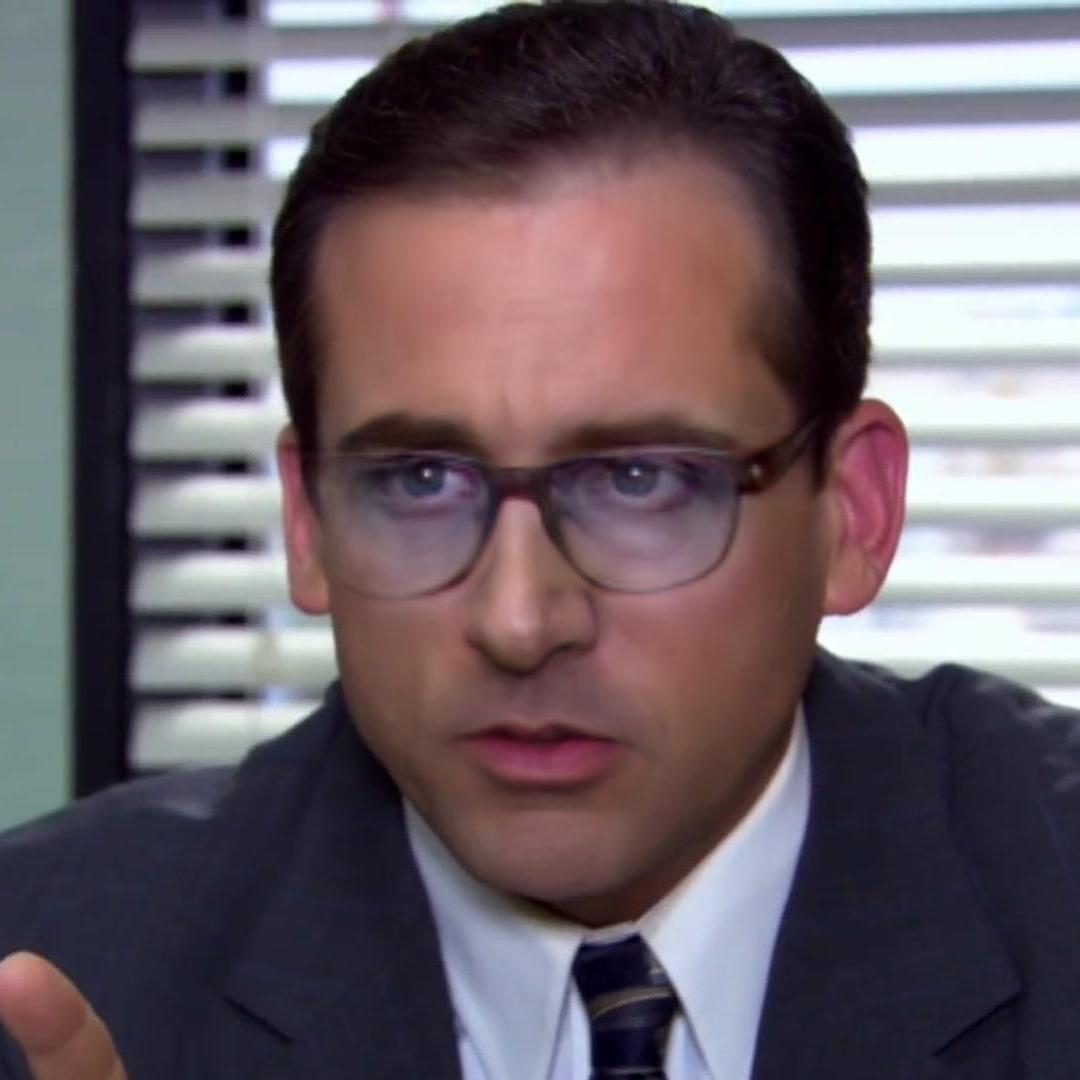}
     \includegraphics[width=\linewidth]{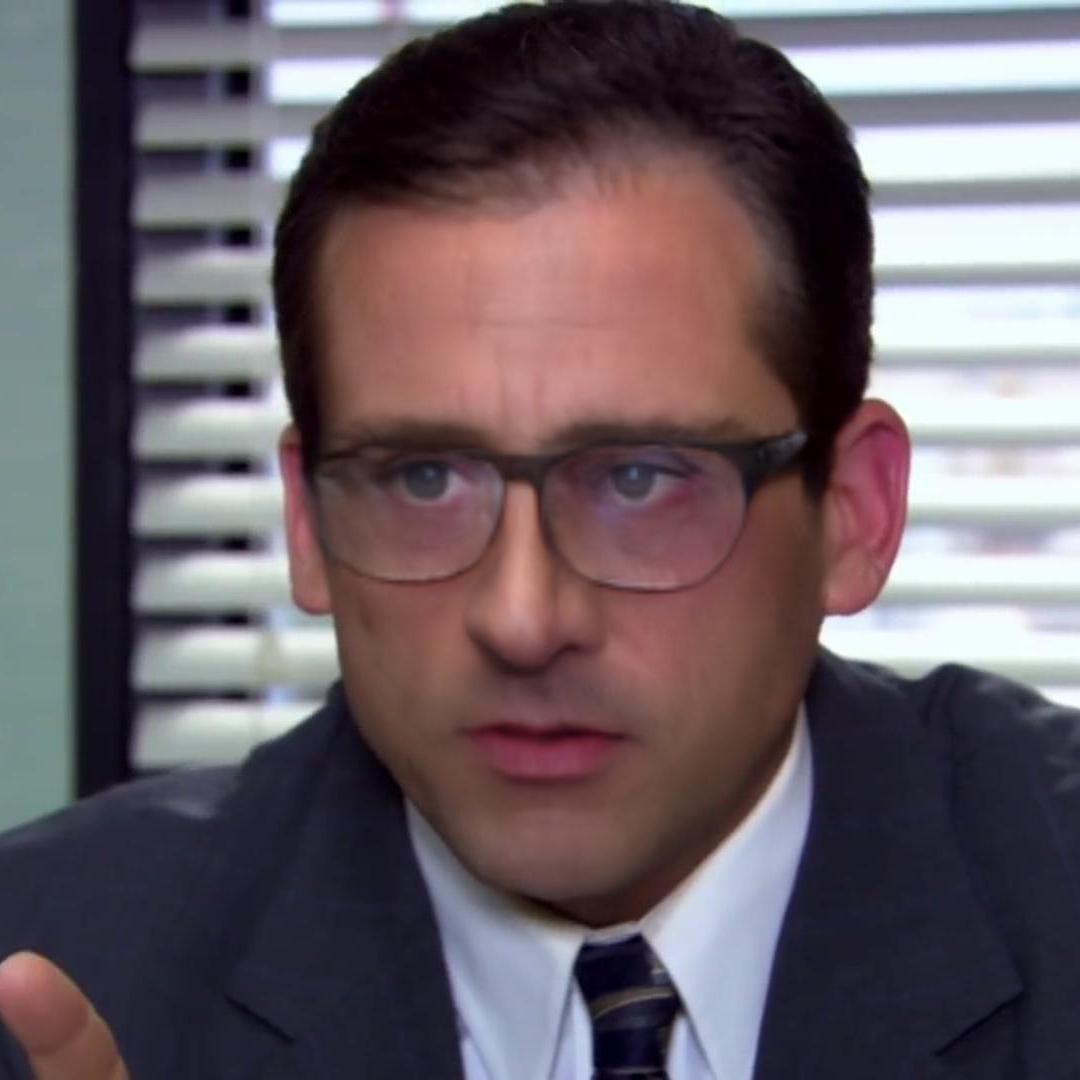}
     \includegraphics[width=\linewidth]{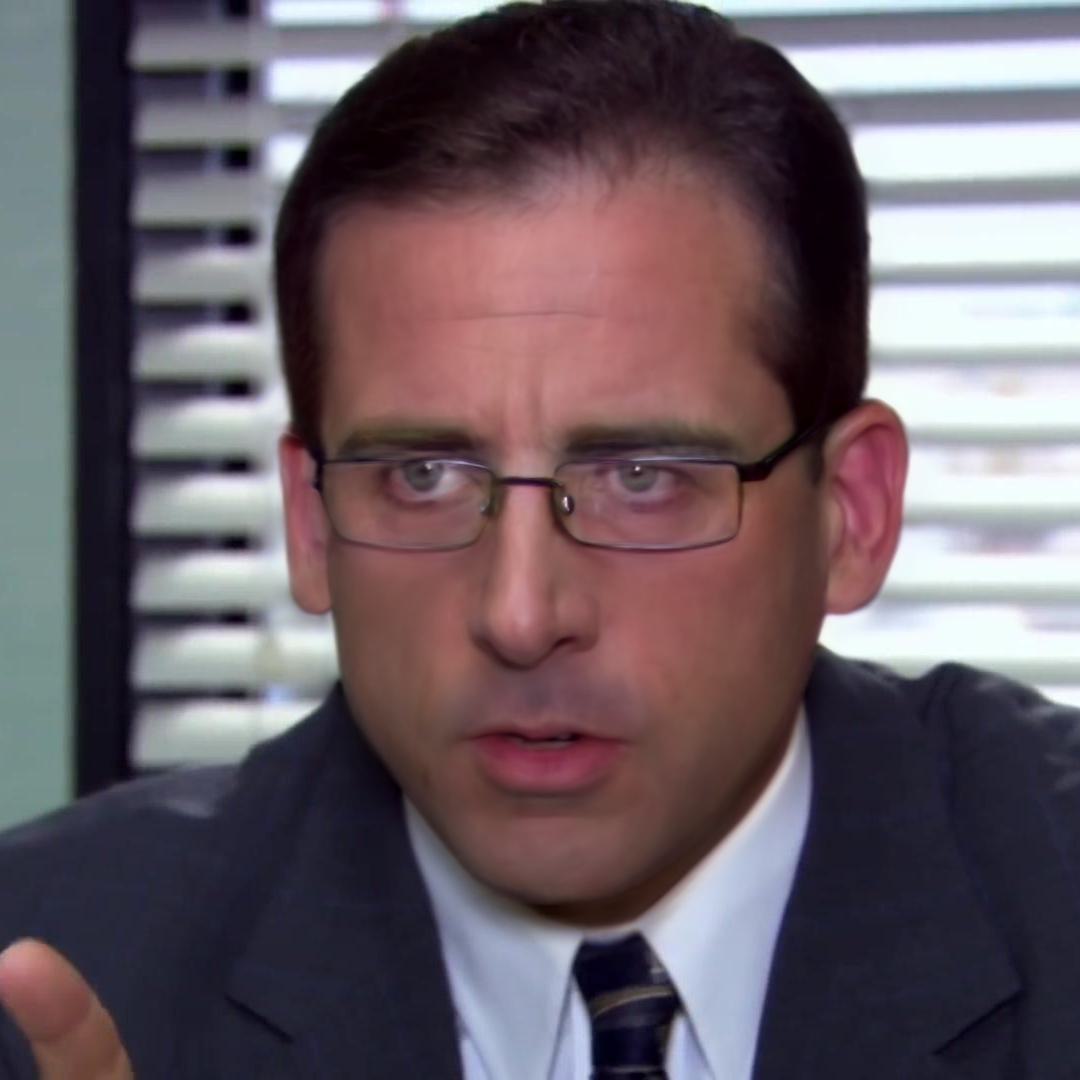}
     \end{minipage}
     }
    \hspace{-2.8mm}
    \subfloat[]{
     \begin{minipage}{0.105\linewidth}
     \includegraphics[width=\linewidth]{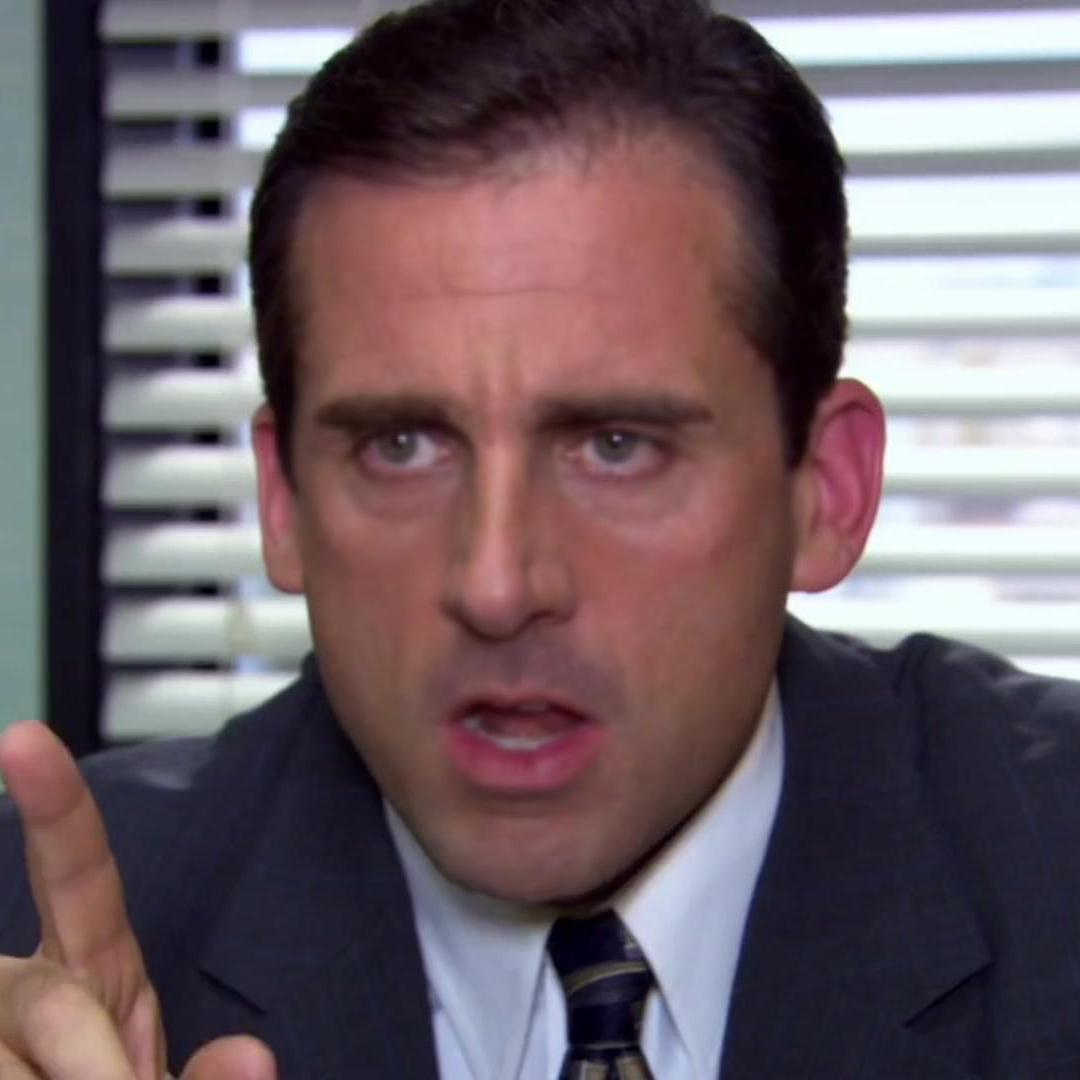}
     \includegraphics[width=\linewidth]{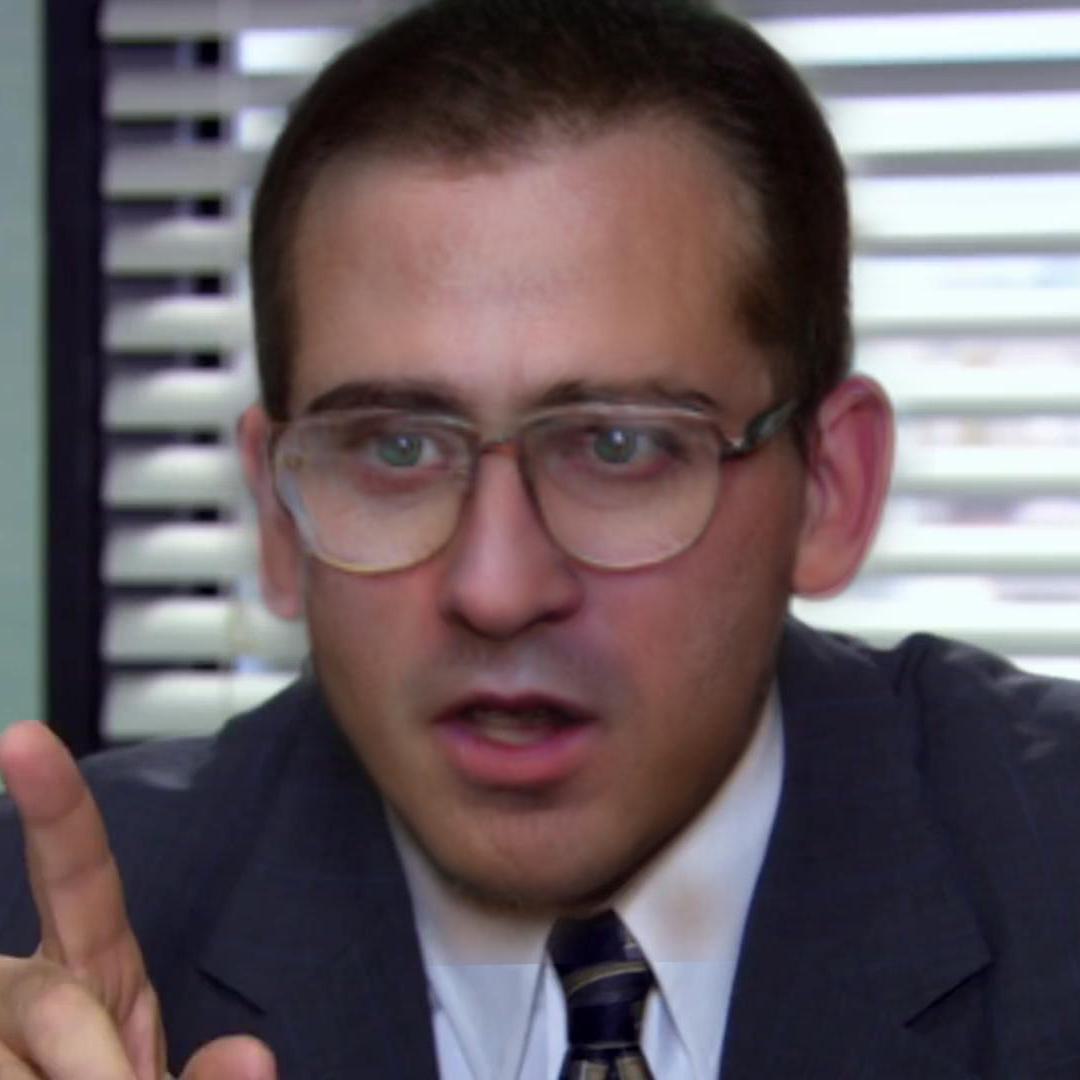}
     \includegraphics[width=\linewidth]{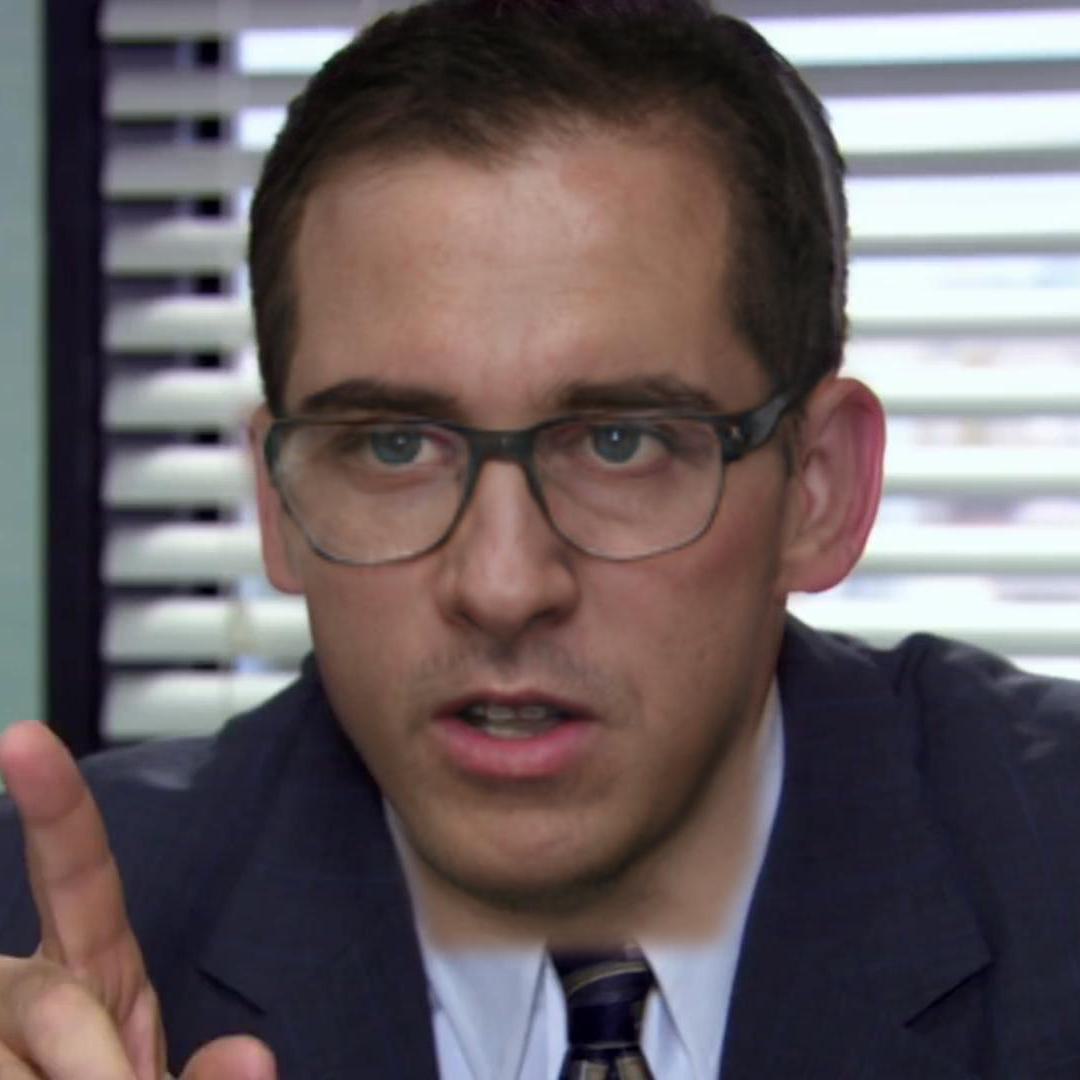}
     \includegraphics[width=\linewidth]{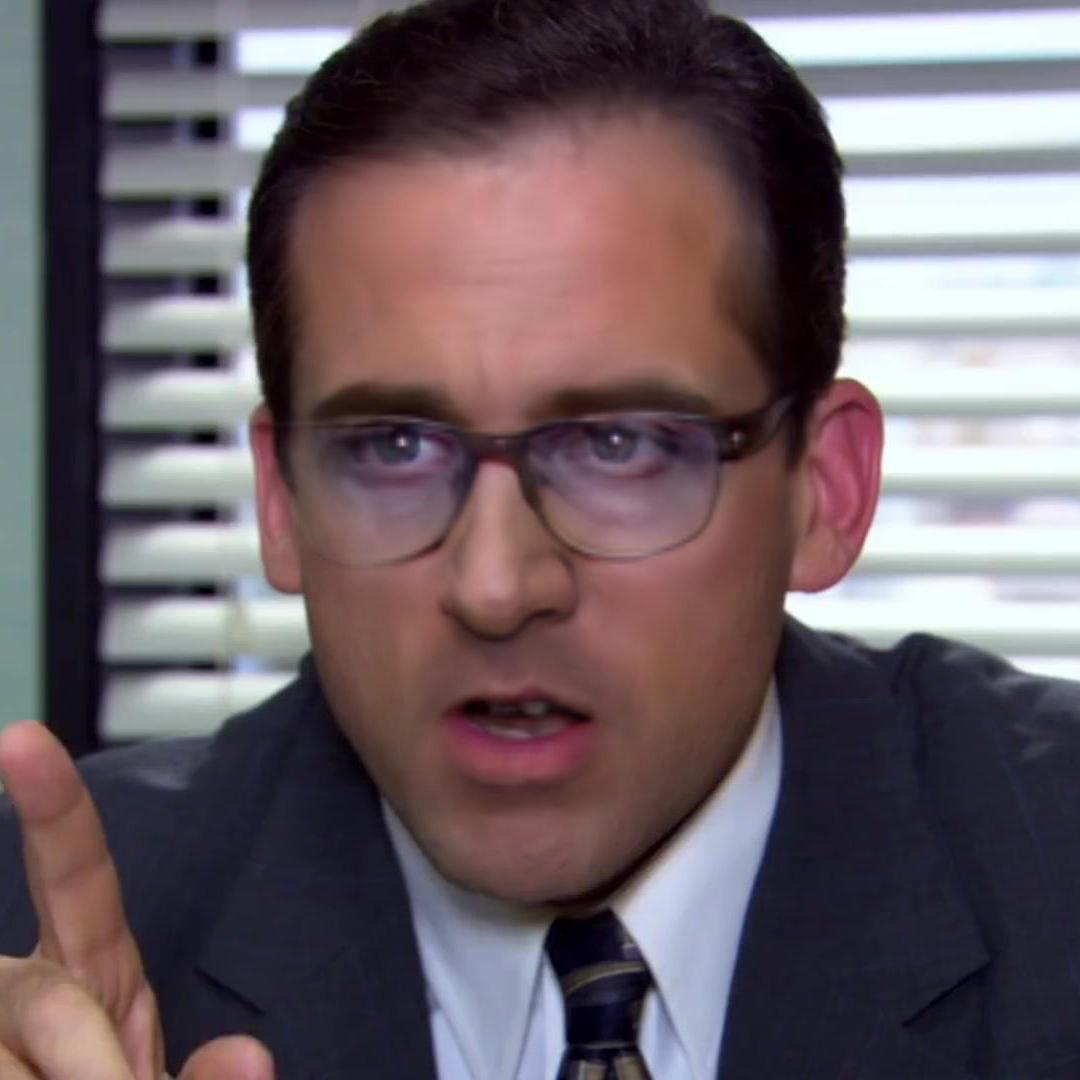}
     \includegraphics[width=\linewidth]{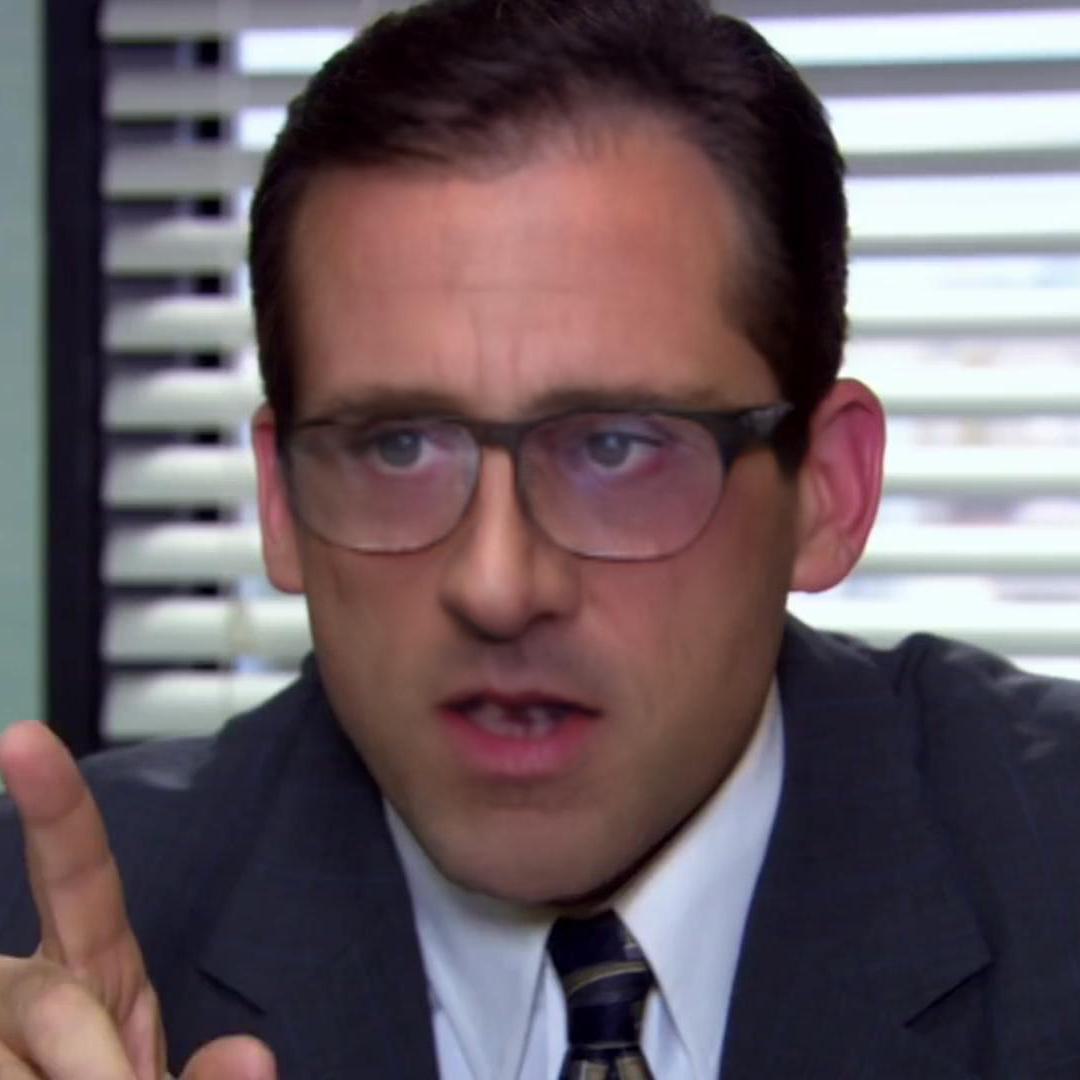}
     \includegraphics[width=\linewidth]{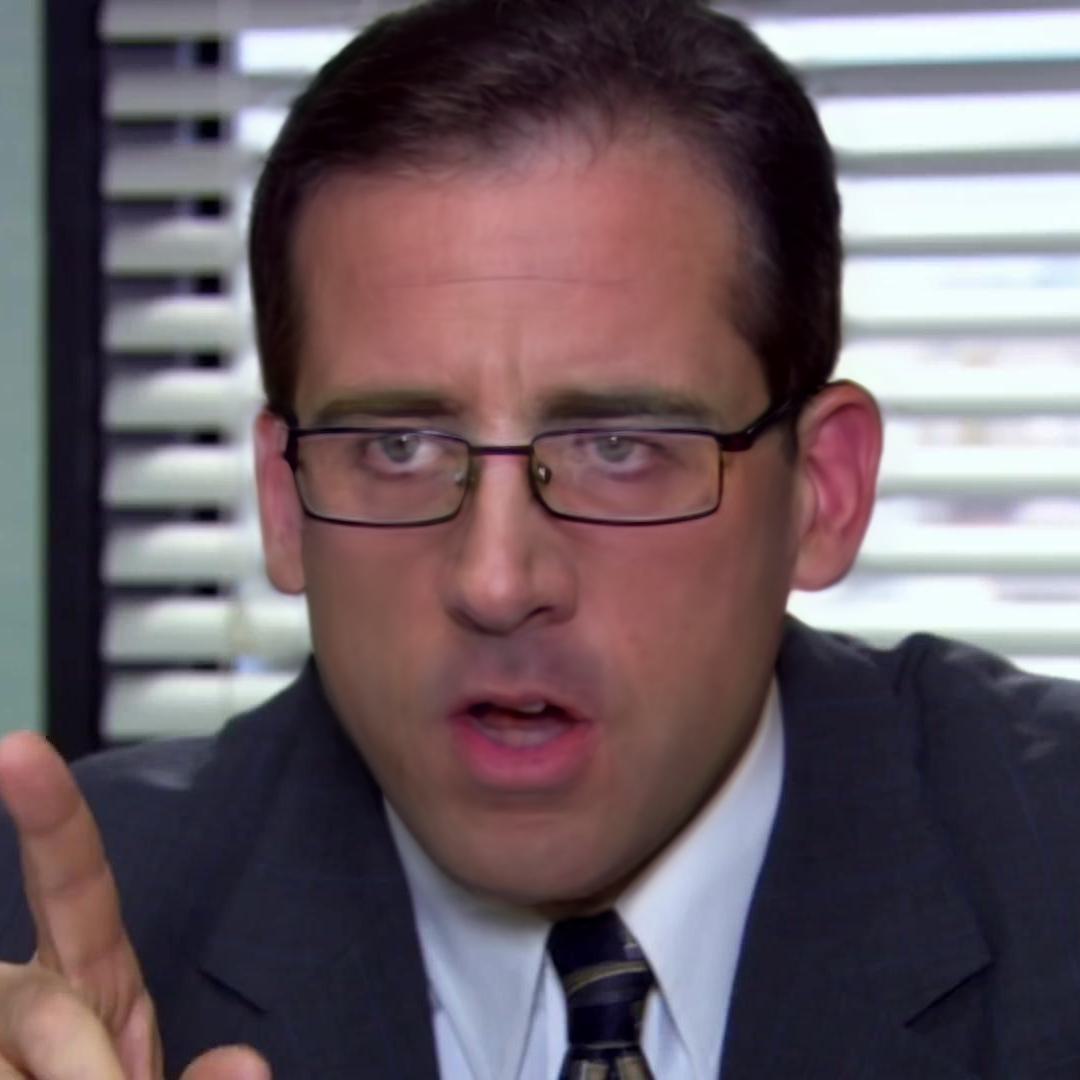}
     \end{minipage}
     }
     \hspace{-1.8mm}
     \subfloat[]{
     \begin{minipage}{0.105\linewidth}
     \includegraphics[width=\linewidth]{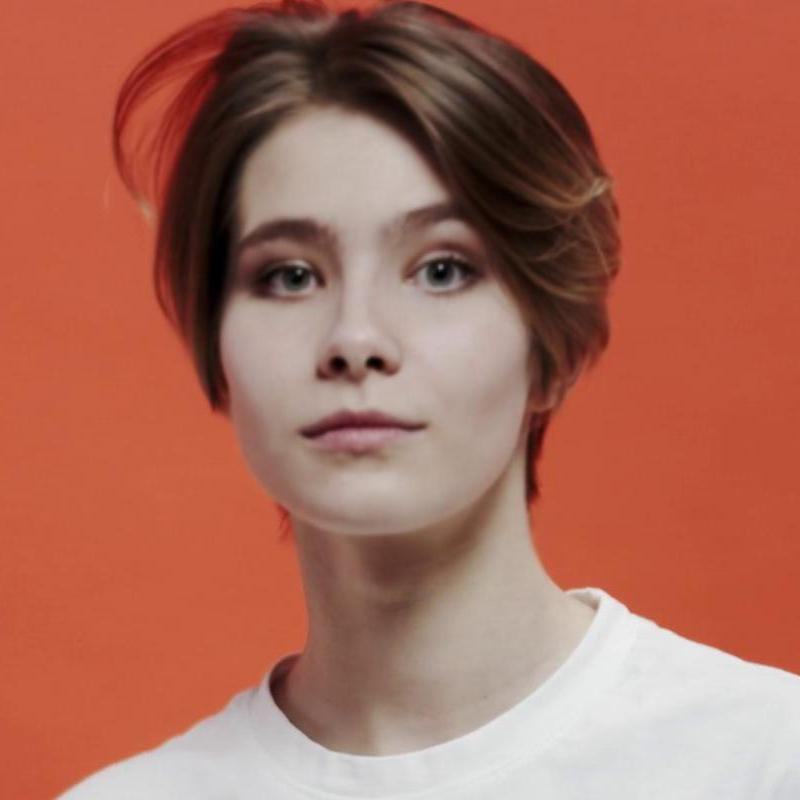}
     \includegraphics[width=\linewidth]{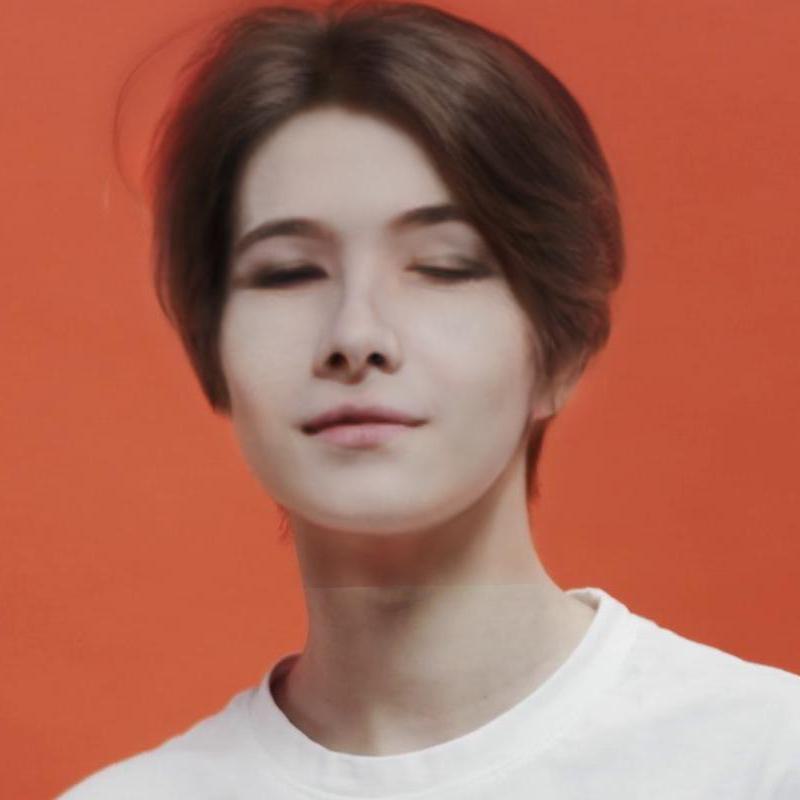}
     \includegraphics[width=\linewidth]{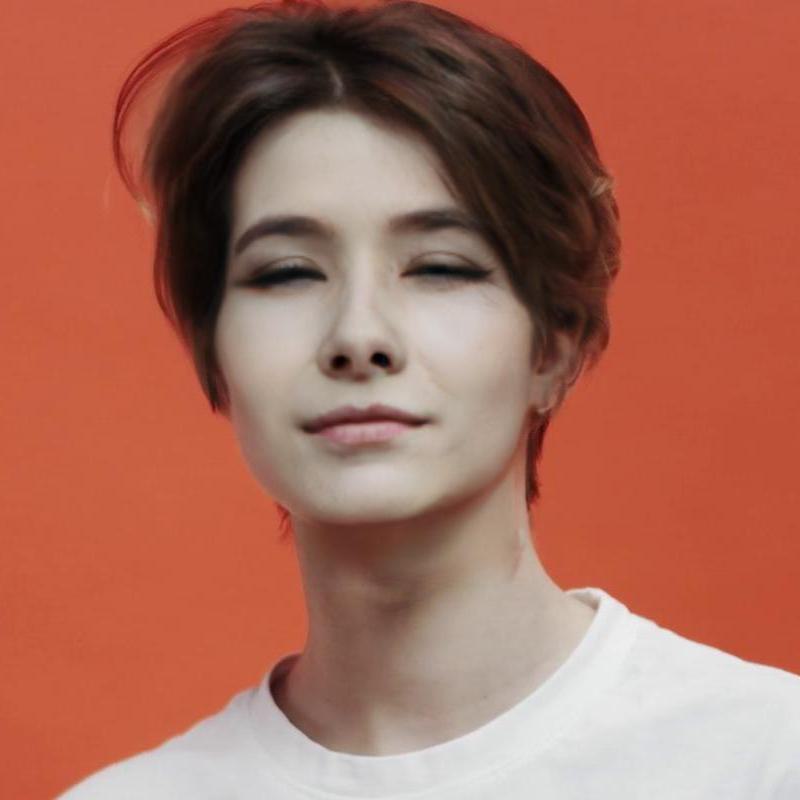}
     \includegraphics[width=\linewidth]{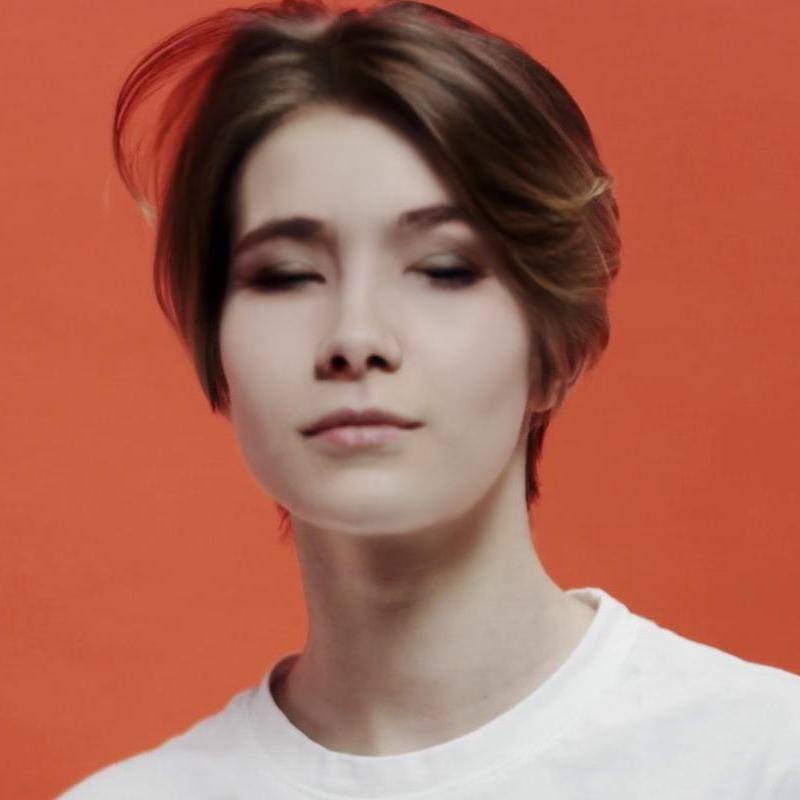}
     \includegraphics[width=\linewidth]{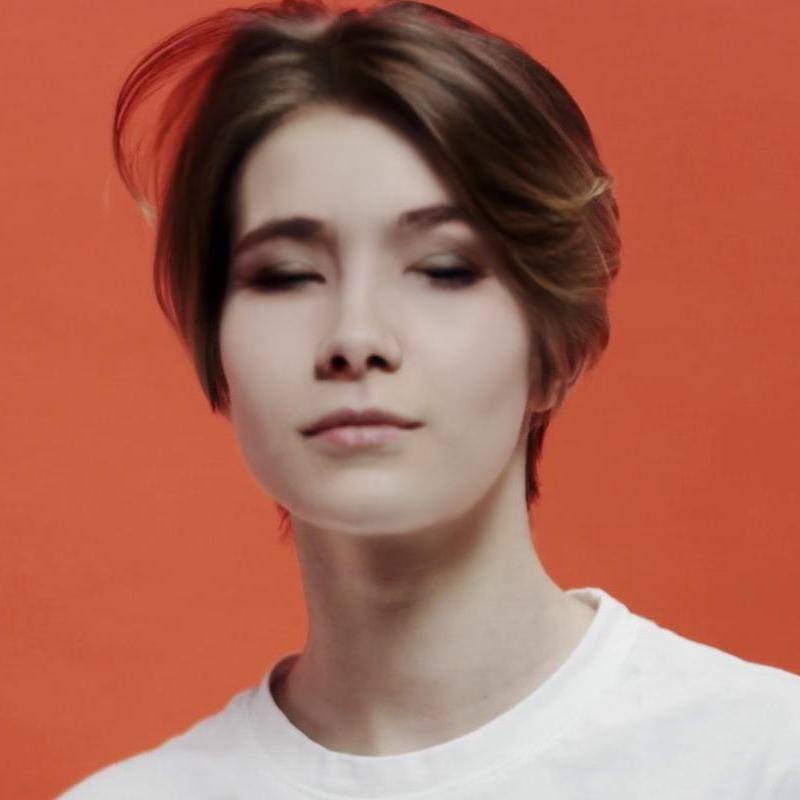}
     \includegraphics[width=\linewidth]{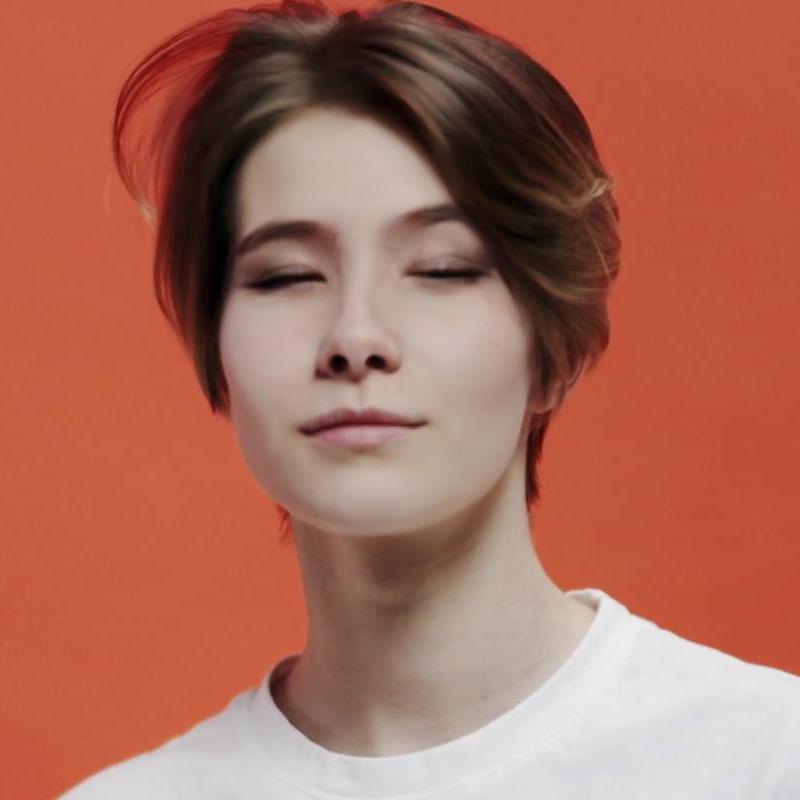}
     \end{minipage}
     }
    \hspace{-2.8mm}
    \subfloat[\texttt{+NarrowEyes}]{
     \begin{minipage}{0.105\linewidth}
     \includegraphics[width=\linewidth]{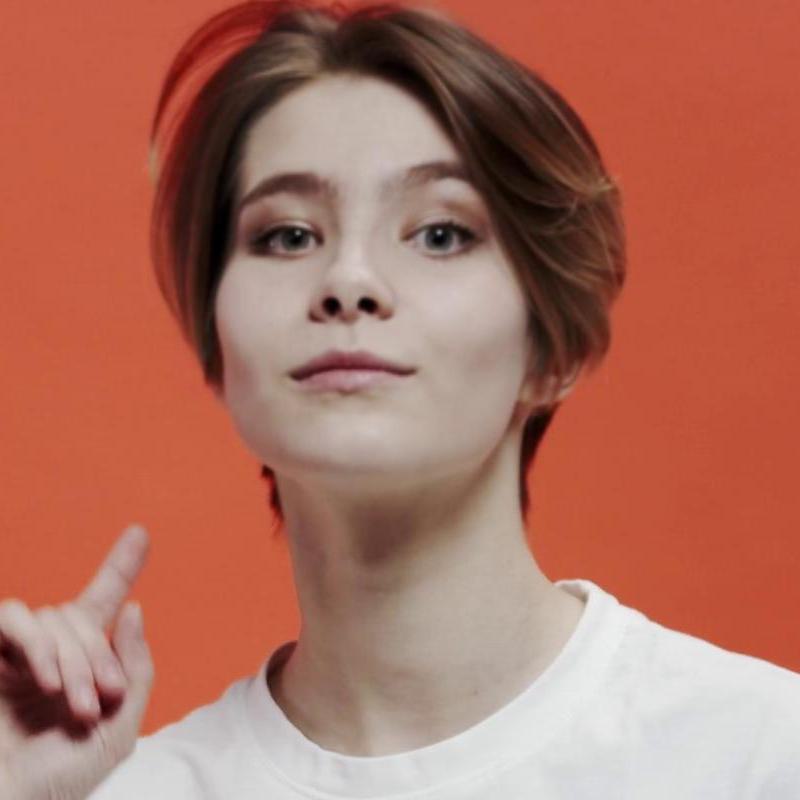}
     \includegraphics[width=\linewidth]{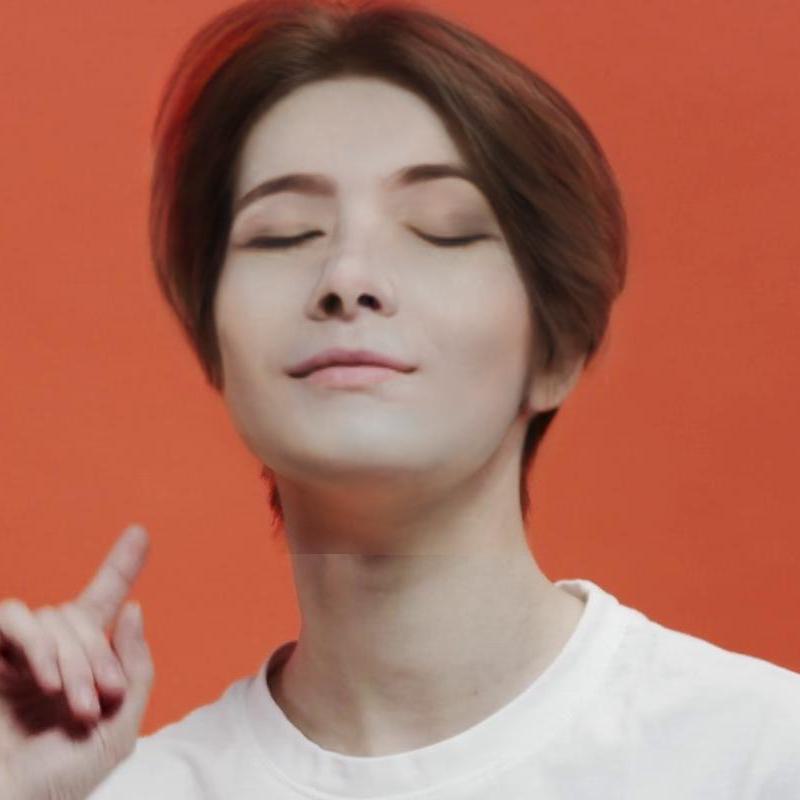}
     \includegraphics[width=\linewidth]{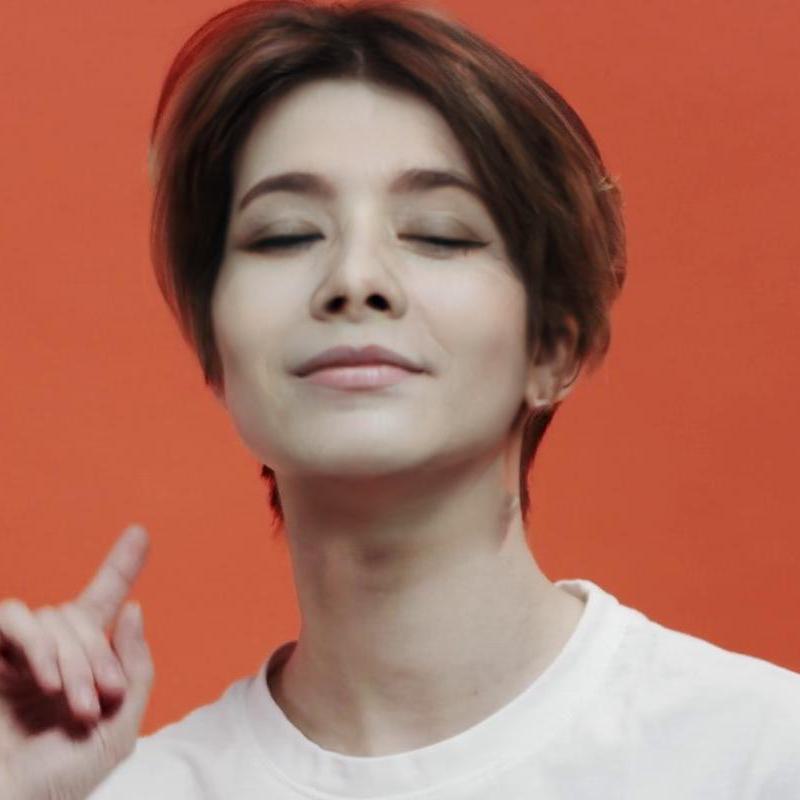}
     \includegraphics[width=\linewidth]{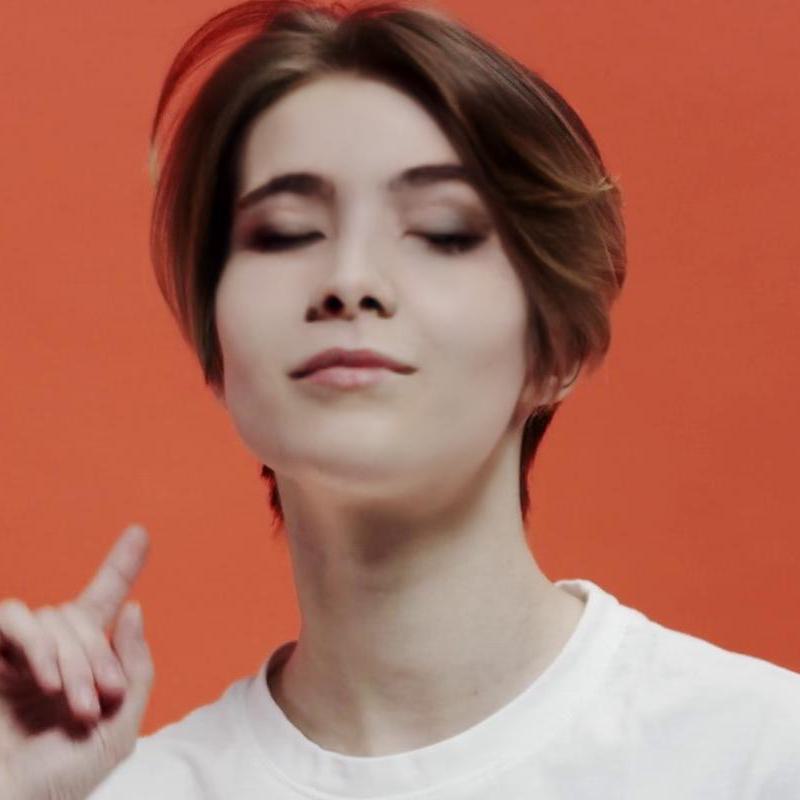}
     \includegraphics[width=\linewidth]{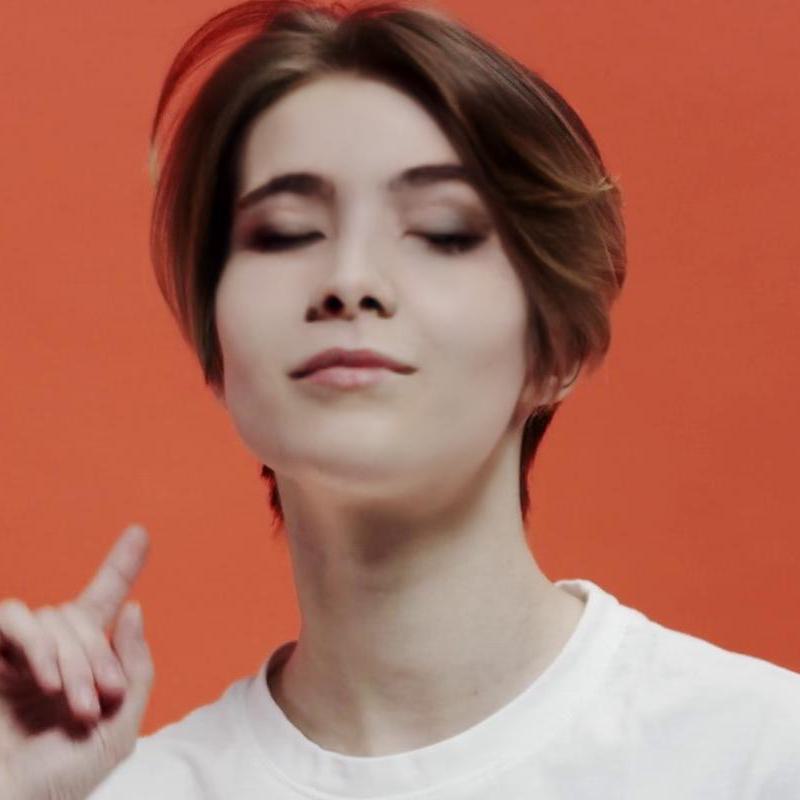}
     \includegraphics[width=\linewidth]{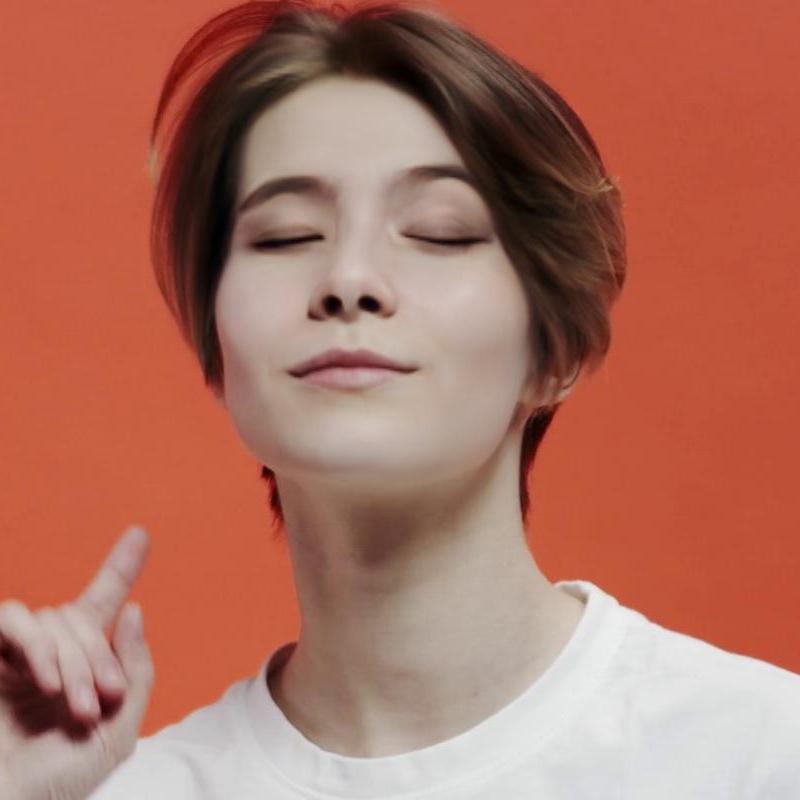}
     \end{minipage}
     }
    \hspace{-2.8mm}
    \subfloat[]{
     \begin{minipage}{0.105\linewidth}
     \includegraphics[width=\linewidth]{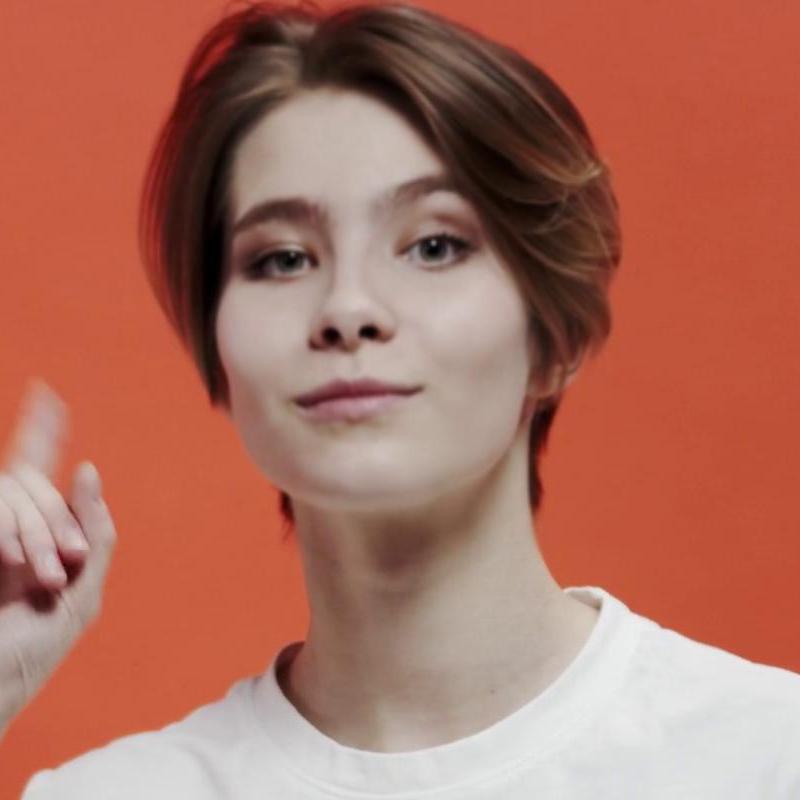}
     \includegraphics[width=\linewidth]{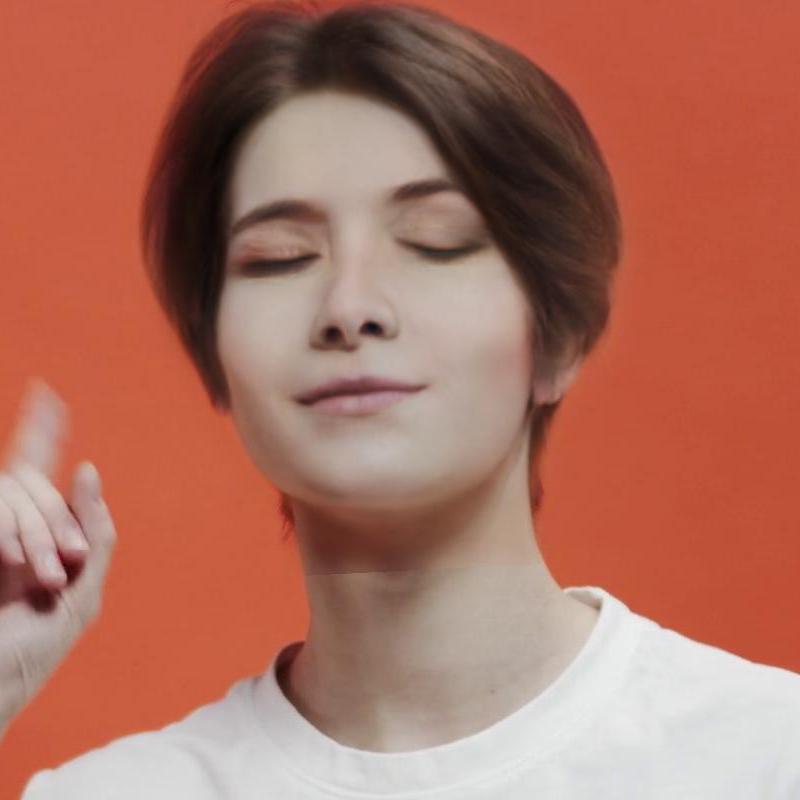}
     \includegraphics[width=\linewidth]{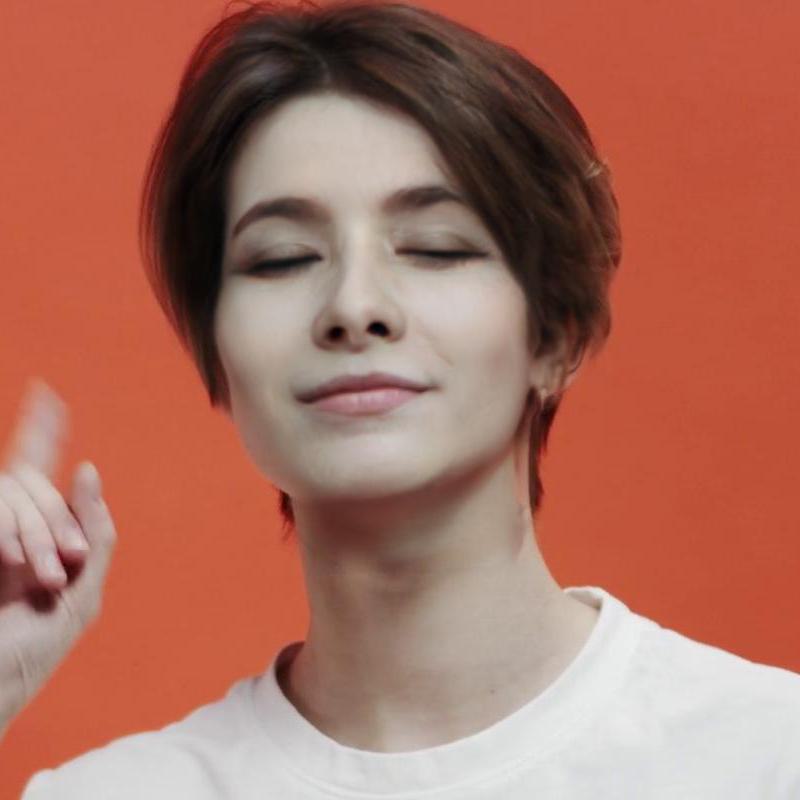}
     \includegraphics[width=\linewidth]{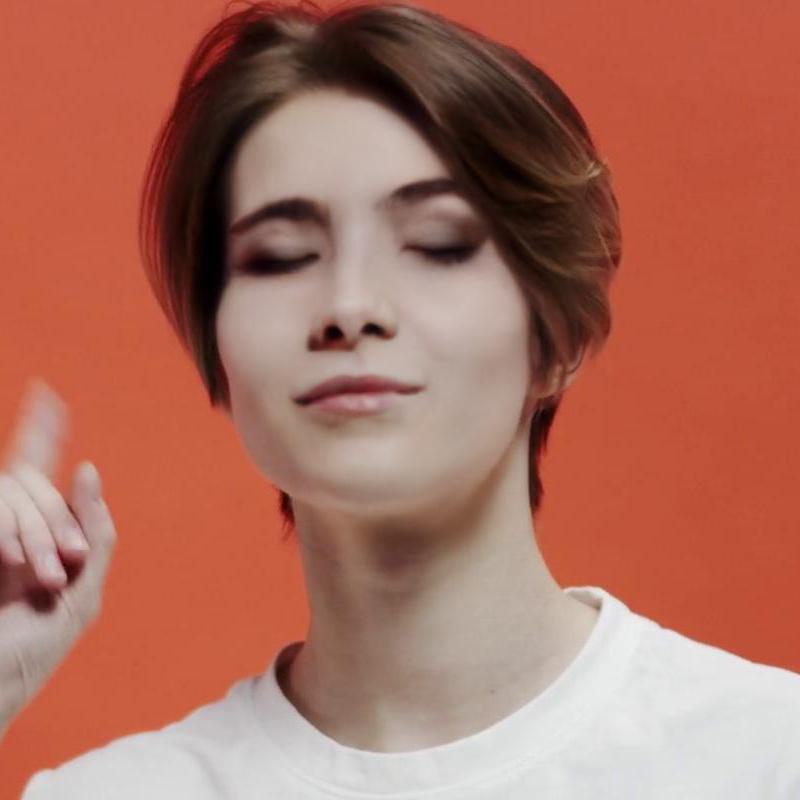}
     \includegraphics[width=\linewidth]{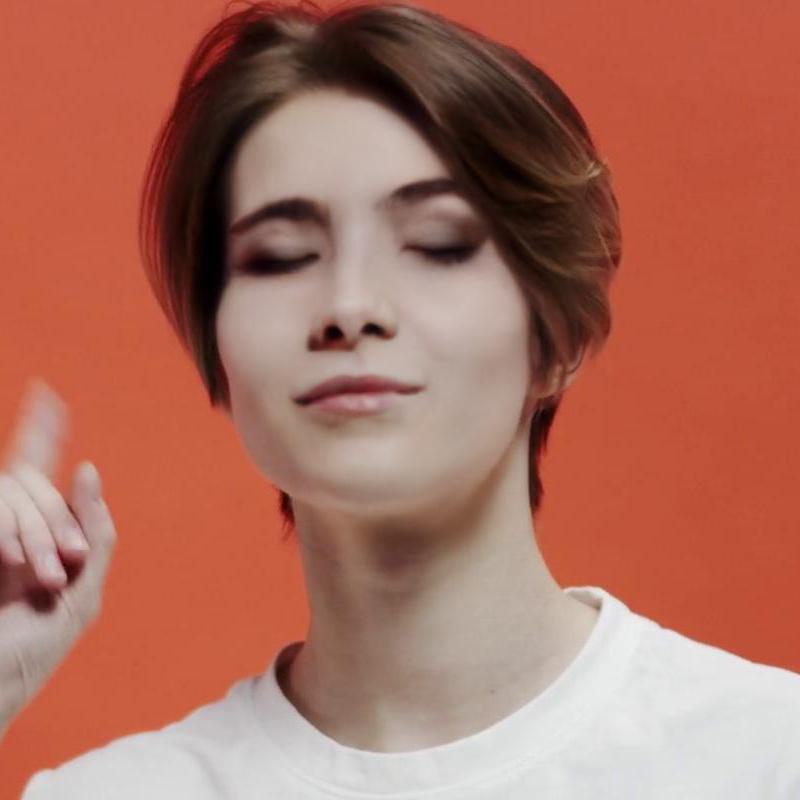}
     \includegraphics[width=\linewidth]{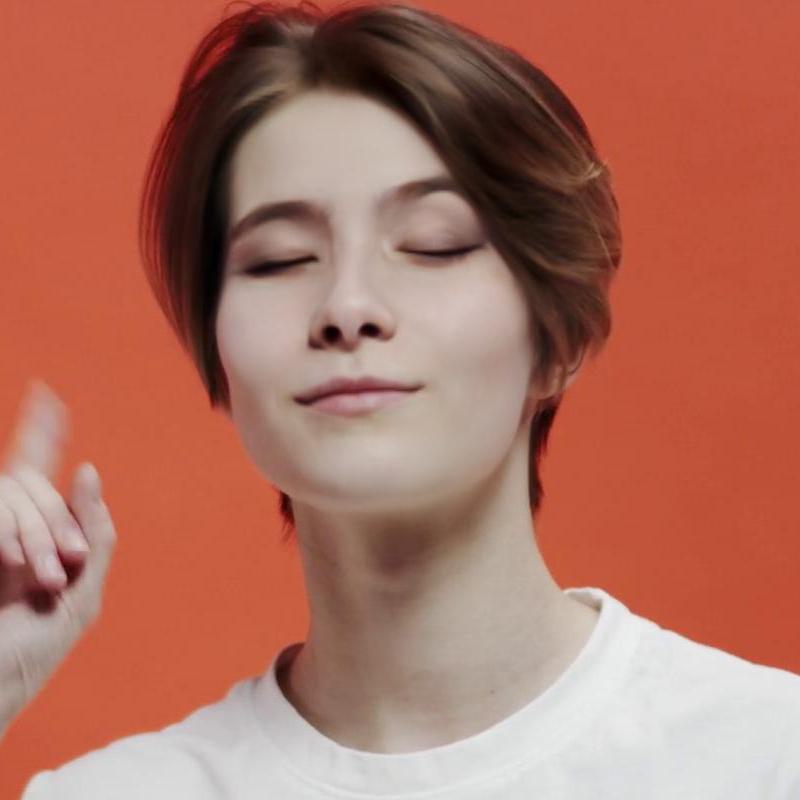}
     \end{minipage}
     }
\rotatebox[origin=c]{270}{\hspace{-0cm} \footnotesize{Original} \hspace{9mm} \footnotesize{IGCI} \hspace{10mm} \footnotesize{Latent-T.} \hspace{10mm} \footnotesize{STIT} \hspace{10mm} \footnotesize{TCSVE} \hspace{10mm} \footnotesize{RIGID} }
\vspace{-2mm}
\caption{Qualitative comparison on video editing. RIGID uses the same post processing as IGCI and Latent-Transformer, but the edited faces can be better blended with the original background. Besides, compared with STIT and TCSVE, our RIGID supports shape editing on the face boundary (\eg, ``Chubby'').}
\label{fig:edit_exp}
\end{figure*}

\textbf{Competitors and Evaluation Metrics.} We compare RIGID with three works, including IGCI~\cite{Xu2021ICCV}, Latent-Transformer~\cite{yao2021latent}, STIT~\cite{tzaban2022stitch}, and TCSVE~\cite{xu2022temporally}. Note that STIT~\cite{tzaban2022stitch} and TCSVE~\cite{xu2022temporally} use the same  method for inversion. We use several metrics for evaluating different methods. For the video inversion task, we use the pixel-wise Mean Square Error (MSE) and Learned Perceptual Image Patch Similarity (LPIPS)~\cite{zhang2018unreasonable} that evaluates the reconstruction quality. For evaluating the temporal stability, we use the flow-based Warp Error (WE) metric. For the edited task, we follow STIT~\cite{tzaban2022stitch} that use Temporally-Local (TL-ID) and Temporally-Global (TG-ID) identity preservation metrics for evaluating the temporal coherence of edited videos. We also use Fréchet Video Distance (FVD)~\cite{unterthiner2019fvd} metric on both edited and inverted videos. Inference Times (IT) over 100 frames is also reported. Please refer to supplementary for more details of competitors and evaluation metrics.


\begin{table*}[h]
   \caption{Video editing comparisons on two datasets. $\downarrow$ denotes the lower the better and vice versa, the best results are marked in \textbf{bold}.}
       \vspace{-6mm}
       \begin{center}
       \setlength{\tabcolsep}{0.35cm}{
       \begin{tabular}{c|c|c|c|c|c|c|c}
           \hline
           \multirow{1}{*}{\diagbox{\small{Methods}}{\small{Metrics}}}       & \multicolumn{3}{c|}{RAVDESS-72} & \multicolumn{3}{c|}{In-the-Wild-36}  &\multirow{2}{*}{IT$\downarrow$(s)}\\
           \cline{2-7}
          &{TL-ID$\uparrow$}   &{TG-ID$\uparrow$}  &{FVD$\downarrow$}      &{TL-ID$\uparrow$}   &{TG-ID$\uparrow$}  &{FVD$\downarrow$}  \\
           \hline
           IGCI~\cite{Xu2021ICCV}         &0.94            &0.79           &834.33           &0.93         &0.71          &499.22          &1.2$\times$e5     \\
           Latent-T.~\cite{yao2021latent}         &\textbf{0.99}   &0.93           &311.84           &\textbf{0.99}         &0.87          &267.42          &\textbf{49.2}      \\
           STIT~\cite{tzaban2022stitch}         &\textbf{0.99}   &\textbf{0.97}           &212.04           &\textbf{0.99}         &0.91          &211.34          &1.6$\times$e3        \\
           TCSVE~\cite{xu2022temporally}          &\textbf{0.99}   &\textbf{0.97}           &201.32           &\textbf{0.99}         &0.92          &198.46          &3.4$\times$e3        \\
           \hline
           RIGID        &\textbf{0.99}   &\textbf{0.97}  &\textbf{198.54}  &\textbf{0.99}&\textbf{0.93} &\textbf{183.36} &54.5 \\
           \hline
       \end{tabular}
       }
      \end{center}
      \label{table:edit}
      \vspace{-5mm}
\end{table*} 

\subsection{Evaluation on Video Inversion}

\textbf{Quantitative Evaluation.} We first evaluate the fidelity and temporal coherence of inverted videos. Quantitative comparison can be seen in Tab.~\ref{table:reconstruction}. We can see that both STIT, IGCI, and TCSVE have lower MSE and LPIPIS values, they optimize latent codes or generator specifically according to the video frames, which can reconstruct the target frames faithfully. IGCI presents the worse performance on WE and FVD metrics, since it optimizes the latent code for each image but fails to consider the temporal coherence of consecutive frames, making the inverted latent codes less consistent. In addition, optimization-based methods cost a lot of time during the inference, especially for IGCI, which takes about 20 minutes for a single frame. Latent-Transformer uses a learning-based encoder that accelerates the inference speed, but it processes each frame individually, both the temporal coherence and reconstruction quality cannot be guaranteed. With about 15$\times$ faster inference than STIT, RIGID achieves a comparable result on the RAVDESS-72 and In-the-Wild-36 datasets. RIGID not only inverts the frame faithfully but also preserves the original temporal relations across frames. Thanks to temporal compensated inversion, RIGID guarantees the fidelity of inverted faces. 

\textbf{Qualitative Evaluation.} The quantitative comparison of video inversion can be seen in Fig.~\ref{fig:reconstruction_exp}. We can see that two optimization-based works, IGCI and STIT reconstruct the frames well. Latent-Transformer utilizes the pSp encoder for inversion~\cite{richardson2021encoding}, and the results present different skin colors with original frames. Thanks to the temporal compensation inversion, our learning-based RIGID presents the competitive results with optimization-based IGCI and STIT on pixel-wise reconstruction. The video comparison can be seen in Fig.~\ref{fig:teaser}. Latent-Transformer cannot provide accurate reconstruction. IGCI inverts each frame faithfully, but it cannot guarantee the temporal coherence across frames, resulting in serious temporal flickering. STIT achieves high-quality inversion on a specific video at the cost of long computational times. In contrast, RIGID builds temporal relations of inverted videos by the temporal compensated inversion, yielding temporal coherent inverted videos with much less time cost.

\begin{table*}[h]
   \caption{Quantitative comparisons with various variants on video inversion and editing. $\downarrow$ denotes the lower the better and vice visa, the best results are marked in \textbf{bold}. The values of MSE are magnified 100 times.}
   \vspace{-6mm}
       \begin{center}
       \setlength{\tabcolsep}{0.35cm}{
       \begin{tabular}{c|c|c|c|c|c|c|c}
       \hline
       \multirow{1}{*}{\diagbox{\small{Variants}}{\small{Metrics}}}      & \multicolumn{4}{c|}{Video Inversion} & \multicolumn{3}{c}{Video Editing}  \\
        \cline{2-8}
        &MSE$\downarrow$ &{LPIPS$\downarrow$}   &{WE$\downarrow$}    &{FVD$\downarrow$}  &{TL-ID$\uparrow$} &{TG-ID$\uparrow$} &{FVD$\downarrow$}\\
           \hline
           $w/o$ TCC                  &1.31  &0.07  &106.52  &208.74   &0.98                &0.94      &321.98   \\
           $w/o$ NM                   &2.28  &0.08  &96.52   &201.22   &0.99                &0.92      &334.98   \\
           $w/o$ LFD                  &1.10  &0.06  &103.61  &215.96   &0.99                &0.94      &309.04   \\
           $w/o$ $\mathcal{L}_{ibfcc}$&\textbf{0.93}  &0.05  &705.22   &358.04              &0.98     &0.90     &299.12   \\
           $w/o$ $\mathcal{L}_{tc}$   &1.08  &0.05  &122.08  &203.54   &0.99                &0.97      &299.34   \\
           \hline
           RIGID                      &1.04 &\textbf{0.05}   &\textbf{84.62}   &\textbf{174.55}              &\textbf{0.99}       &\textbf{0.97}   &\textbf{198.54} \\
           \hline
       \end{tabular}
       }
      \end{center}
      \label{table:ablation}
      \vspace{-5mm}
\end{table*}

\subsection{Evaluation on Video Editing}

\textbf{Quantitative Evaluation.} The quantitative comparison of video editing can be seen in Tab.~\ref{table:edit}. We can see that IGCI~\cite{Xu2021ICCV} has high FVD values on both two datasets. As discussed above, it cannot produce consistent latent codes, making the edited frames discontinuous. In addition, IGCI has lower values on both TL-ID and TG-ID, which evidences that identity information cannot be preserved locally and globally. Compared with IGCI, Latent-Transformer~\cite{yao2021latent} uses an encoder for producing latent codes, and the edited frames are more consistent. STIT~\cite{tzaban2022stitch} optimizes the generator for each video hence requires many inference times. RIGID achieves a comparable result on the RAVDESS-72 dataset with 30$\times$ faster during the inference. As for the In-the-Wild-36 Dataset, RIGID outperforms all competitors with three metrics. This should contribute to our \textit{in-between frame composition constraint}. It enforces the smoothness of edited frames and brings temporal coherence into edited videos.

\textbf{Qualitative Evaluation.} We present the qualitative comparison on video editing in Fig.~\ref{fig:teaser} and Fig.~\ref{fig:edit_exp}. We can see that ICGI presents blurry edited faces with temporal flickering, and Latent-Transformer loses temporal coherence on local details (see bangs in Fig.~\ref{fig_lt}). In addition, though they use the same post processing as our RIGID, their edited faces cannot be well blended with the original background. Without considering the temporal coherence of edited frames, their edited faces have large structure deformations from the original faces. STIT proposes a ``stitching tuning'' strategy for the seamless blending, it enforces the edit frames have similar transitions around the face boundary with originals. However, when the target face's boundary is close to the background, this method does not support its shape-related editing (\eg, ``Chubby'' or ``Double Chins''). As shown in the $1_{st}$ sample in Fig.~\ref{fig:edit_exp}, STIT fails on the ``Chubby'' editing. TCSVE uses the same strategy, hence it fails on this editing. More video comparisons can be found in the supplementary materials.

\subsection{Ablation Study}

In this section, we perform an ablation study to evaluate our RIGID on the RAVDESS-72 dataset. We develop five variants with the modification of the modules and the loss functions: 1) $w/o$ TCC, by removing temporal compensated code $w_t^{\prime}$. 2) $w/o$ NM, by removing noise map $n_t$. 3) $w/o$ LFD, by removing the latent frequency disentanglement in the framework. In this variant, frames in the same video have different latter codes $w^l_t$. 4) $w/o$ $L_{ibfcc}$, by removing \textit{in-between frame composition constraint}. 5) $w/o$ $\mathcal{L}_{tc}$, by removing the temporal consistent loss $\mathcal{L}_{tc}$.

The quantitative comparisons with various variants can be seen in Tab.~\ref{table:ablation}. Compared with RIGID, variant $w/o$ TCC has a large WE value. Code $w_t^{\prime}$ introduces the temporal compensation to the image-based latent code, and brings the temporal coherence in inverted videos. Variant $w/o$ NM has large MSE and LPIPS values. Since noise map $n_t$ injects the spatial information to the generator, and improves the inversion accuracy. Both variant $w/o$ LFD and $w/o$ $\mathcal{L}_{tc}$ have large WE numbers. Our latent frequency disentanglement strategy unifies the high-frequency information in a video, and $\mathcal{L}_{tc}$ preserves the temporal consistency from the original to the inverted video. They learn the temporal relations effectively. We observe that variant $w/o$ $\mathcal{L}_{ibfcc}$ has the largest WE and FVD values on inverted videos, though it is applied to the edited frames. That is because the constraint enforces the smoothness of edited frames, and can be propagated to the latent codes that control the coherence of inverted frames. Meanwhile, variant $w/o$ $\mathcal{L}_{ibfcc}$ has the worse performance on metrics TL-ID and TG-ID. Without the in-between constraint, the framework cannot guarantee the identity similarity across frames. By learning the temporal correlations in inverted and edited videos, the final RIGID achieves the best performance on two tasks.

\begin{figure}
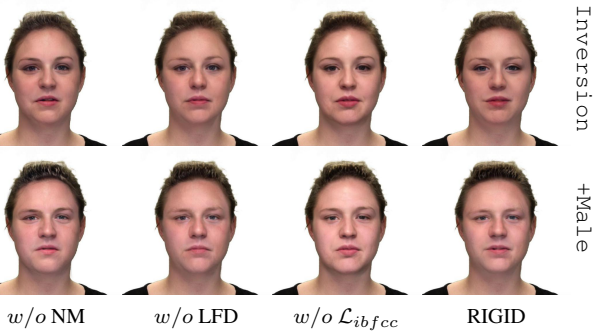

    \centering
    \captionsetup[subfloat]{labelformat=empty,justification=centering}
    \subfloat[\footnotesize{$w/o$ NM}]{
    \begin{minipage}{0.235\linewidth}
    \animategraphics[autoplay,loop,width=\textwidth]{15}{figure/rec/RAVDESS/WO_Noise_32/01-02-02-02-01-01-10/}{0000}{0048}\\
    \animategraphics[autoplay,loop,width=\textwidth]{15}{figure/edit/RAVDESS/WO_Noise_32/01-02-02-02-01-01-10_Male_rep/}{0000}{0048}
    \end{minipage}
    }
    \hspace{-2.25mm}
    \subfloat[\footnotesize{$w/o$ LFD}]{
    \begin{minipage}{0.235\linewidth}
    \animategraphics[autoplay,loop,width=\textwidth]{15}{figure/rec/RAVDESS/WO_LCSS_32/01-02-02-02-01-01-10/}{0000}{0048}\\
    \animategraphics[autoplay,loop,width=\textwidth]{15}{figure/edit/RAVDESS/WO_LCSS_32/01-02-02-02-01-01-10_Male_rep/}{0000}{0048}
    \end{minipage}
    }
    \hspace{-2.25mm}
    \subfloat[\footnotesize{$w/o$ $\mathcal{L}_{ibfcc}$}]{
    \begin{minipage}{0.235\linewidth}
    \animategraphics[autoplay,loop,width=\textwidth]{15}{figure/rec/RAVDESS/WO_IFSC_32/01-02-02-02-01-01-10/}{0000}{0048}\\
    \animategraphics[autoplay,loop,width=\textwidth]{15}{figure/edit/RAVDESS/WO_IFSC_32/01-02-02-02-01-01-10_Male_rep/}{0000}{0048}
    \end{minipage}
    }
    \hspace{-2.25mm}
    \subfloat[\footnotesize{RIGID}]{
    \begin{minipage}{0.235\linewidth}
    \animategraphics[autoplay,loop,width=\textwidth]{15}{figure/rec/RAVDESS/FULL_32/01-02-02-02-01-01-10/}{0000}{0048}\\
    \animategraphics[autoplay,loop,width=\textwidth]{15}{figure/edit/RAVDESS/FULL_32/01-02-02-02-01-01-10_Male_rep/}{0000}{0048}
    \end{minipage}
    }
    \rotatebox[origin=c]{270}{\hspace{-4.5mm}\small{\texttt{Inversion}} \hspace{5mm} \small{\texttt{+Male}}}\\
\caption{Qualitative comparison with different variants. Variant $w/o$ LFD and $w/o$ $\mathcal{L}_{ibfcc}$ present noticeable temporal flickering. This figure contains \emph{animated videos}, which are best viewed using Adobe Acrobat.}
\label{fig:abation}
\vspace{-0.5cm}
\end{figure}

Qualitative comparison with different variants can be seen in Fig.~\ref{fig:abation}. We can see variant $w/o$ $\mathcal{L}_{ibfcc}$ presents annoying flickering both on the inverted and edited videos, which evidences the effectiveness of \textit{in-between frame composition constraint}. As talked about in the main paper, it builds the temporal relations of video frames, which guarantees the smoothness of generated videos. Variant $w/o$ NM cannot invert the video faithfully. Variant $w/o$ LFD may present temporal flickering. The final RIGID presents temporal coherent inversion and editing videos.

\section{Conclusion and Discussions}
\label{sec:conclusion}

In this paper, we propose RIGID, a novel recurrent framework that learns the temporal correlations between successive frames for both inversion and editing. RIGID first learn the temporal correlation of inverted videos by a temporal-compensated reconstruction. It takes previous and current faces as input, and produces a temporal compensated code with a spatial noise map. They, together with the image-based latent code, are fed to the StyleGAN generator to reconstruct the video frames. Meanwhile, RIGID disentangles high-frequency artifacts in the latent space for eliminating the temporal flickering. Furthermore, for learning the temporal correlation of edited video, we enforce an edited frame can be composed by the flow-based warping of its neighbor frames. RIGID achieves attribute-agnostic editing with much less time cost. Extensive experiments demonstrate the superiority of RIGID.

\textbf{Limitations.}
As cropping and alignment of faces are needed before the inversion, those hair portions outside the cropped region cannot be edited by our method, which leads to unnatural conflicts after blending. It may be addressed by using StyleGAN3~\cite{karras2021alias}~since it can generate unaligned faces. However, as indicated in~\cite{alaluf2022third}, StyleGAN3 has a worse edit ability than StyleGAN2, which is essential in GAN inversion. 

\textbf{Broader Impact.}
RIGID can potentially be used for academic research and commercial applications. It provides a pipeline for video-based GAN inversion, and benefits the design of temporal coherent video GANs. Besides, RIGID can create more facial videos in the media industry. On the other hand, although not the purpose of this work, it can potentially be misused for the violation of portrait rights and privacy. The risk can be mitigated by face forgery detection methods, \ie, we can detecte the blending boundary using~\cite{li2020face}.

\textbf{Acknowledgement.}
This paper is partially supported by the National Key R$\&$D Program of China No.2022ZD0161000; the General Research Fund of Hong Kong No.17200622; National Natural Science Foundation of China (No. 61972162); Guangdong Natural Science Funds for Distinguished Young Scholars (No. 2023B1515020097); and Singapore Ministry of Education Academic Research Fund Tier 1 (MSS23C002).

{\small
\bibliographystyle{ieee_fullname}
\bibliography{egbib}
}

\newpage

\title{Supplement to RIGID: Recurrent GAN Inversion and Editing of Real Face Videos}

%


\setcounter{figure}{6}
\setcounter{table}{3}
\setcounter{section}{0}

\author{Yangyang Xu$^1$,
       Shengfeng He$^2$,
       Kwan-Yee K. Wong$^1$,
       and Ping Luo$^{1,3}$
       \\
       {$^1$Deapartment of Computer Science, The University of Hong Kong} \\
       {$^2$School of Computing and Information Systems, Singapore Management University}\\
       {$^3$Shanghai AI Laboratory}\\
      }

\teaser{
}

\maketitle

In this supplement, we provide more experimental details, which includes the introduction of competitors, the introduction of evaluation metrics, and the qualitative comparisons. More results can be found in \url{https://cnnlstm.github.io/RIGID}.

\section{Introduction of Competitors}
We compare RIGID with three works, including IGCI~\cite{Xu2021ICCV}, Latent-Transformer~\cite{yao2021latent} and STIT~\cite{tzaban2022stitch}. IGCI~\etal~\cite{Xu2021ICCV} introduces consecutive images into GAN inversion for improving the quality of the reconstruction and editability. Latent-Transformer~\cite{yao2021latent} inverts each  individually using pSp~\cite{richardson2021encoding} encoder. STIT~\cite{tzaban2022stitch} optimizes the generator for each video based on the ``pivot'' latent codes in the first step then uses a ``stitching tuning'' strategy on edited frames for the seamless, both two steps cost many times. TCSVE~\cite{xu2022temporally} extends this idea by proposing a temporal consistency loss that applies on the edited frames. Note their seamless blending is only applied to the edited frames, and we use the official code to obtain their inverted frames. Besides, we use the same post processing as RIGID on IGCI and Latent-Transformer for a fair comparison.

\begin{figure*}
    \centering
    \subfloat[\footnotesize{Original}]{
    \begin{minipage}{0.19\linewidth}
    \animategraphics[autoplay,loop,width=\textwidth]{24}{./figure/rebuttal/jim_ori/}{0000}{0036}
    \end{minipage}
    }
    \hspace{-2.25mm}
    \subfloat[\footnotesize{\texttt{+Wearing Lipstick}}]{
    \begin{minipage}{0.19\linewidth}
    \animategraphics[autoplay,loop,width=\textwidth]{24}{./figure/rebuttal/jim_Wearing_Lipstick_rep/}{0000}{0036}
    \end{minipage}
    }
    \hspace{-2.25mm}
    \subfloat[\footnotesize{\texttt{+Narrow Eyes}}]{
    \begin{minipage}{0.19\linewidth}
    \animategraphics[autoplay,loop,width=\textwidth]{24}{./figure/rebuttal/jim_Narrow_Eyes_rep/}{0000}{0036}
    \end{minipage}
    }
    \hspace{-2.25mm}
    \subfloat[\footnotesize{\texttt{+Young}}]{
    \begin{minipage}{0.19\linewidth}
    \animategraphics[autoplay,loop,width=\textwidth]{24}{./figure/rebuttal/jim_Young_rep/}{0000}{0036}
    \end{minipage}
    }
    \hspace{-2.25mm}
    \subfloat[\footnotesize{\texttt{+Arched Eyebrows}}]{
    \begin{minipage}{0.19\linewidth}
    \animategraphics[autoplay,loop,width=\textwidth]{24}{./figure/rebuttal/jim_Arched_Eyebrows_rep/}{0000}{0036}
    \end{minipage}
    }
\caption{Our RIGID also works well without seeing any editings during training. This figure contains \emph{animated videos}, which are best viewed using Adobe Acrobat.}
\label{fig:new_edit}
\end{figure*}

\begin{table*}[h!]
   \caption{{Quantitative comparisons with various variants on video inversion and editing. $\downarrow$ denotes the lower the better and vice visa, the best results are marked in \textbf{bold}.}}
   \vspace{-6mm}
       \begin{center}
       \setlength{\tabcolsep}{0.2cm}{
       \begin{tabular}{c|c|c|c|c|c|c|c}
       \hline
       \multirow{1}{*}{\diagbox{Variants}{Metrics}}       & \multicolumn{4}{c|}{Video Inversion} & \multicolumn{3}{c}{Video Editing}  \\
        \cline{2-8}
        &MSE$\downarrow$($\times$e-2) &{LPIPS$\downarrow$}   &{WE$\downarrow$}    &{FVD$\downarrow$}  &{TL-ID$\uparrow$} &{TG-ID$\uparrow$} &{FVD$\downarrow$}\\
           \hline
           $E(I_t^a)$  &\textbf{1.018}  &\textbf{0.051} &90.252      &196.335   &0.996     &0.938      &278.619   \\
           $E(I_t^a,f^a_{t \Rightarrow t-1})$  &1.334  &0.066  &101.592   &210.044   &0.982     &0.934      &244.821   \\
           2 layers        &1.019  &\textbf{0.051}  &89.223   &183.455   &\textbf{0.999}     &0.962      &221.054   \\
           6 layers             &1.045  &0.053  &\textbf{83.923}   &185.476   &\textbf{0.999}     &0.967      &215.994   \\
           $w/o$ \texttt{editing}        &\textbf{1.025}  &\textbf{0.051}  &\textbf{79.923}   &\textbf{168.241}   &0.998     &0.932      &242.134   \\
           \texttt{WA}                    &1.45  &0.07  &119.29            &247.42                &0.97                &0.94      &275.45   \\
           \texttt{FI}                    &1.23  &0.06  &243.34           &263.22                &0.98                &0.94      &318.16  \\
           \hline
           RIGID     &1.040 &\textbf{0.051}   &{84.623}   &{174.551}  &\textbf{0.999} &\textbf{0.969}   &\textbf{198.536} \\
           \hline
       \end{tabular}
       }
      \end{center}
      \label{table:ablation}
      \vspace{-2.5mm}
\end{table*}

\begin{table*}[h!]
      \caption{{Quantitative comparisons with Latent-Transformer on video inversion and editing. $\downarrow$ denotes the lower the better and vice visa, the best results are marked in \textbf{bold}.}}
      \vspace{-6mm}
       \begin{center}
       \setlength{\tabcolsep}{0.2cm}{
       \begin{tabular}{c|c|c|c|c|c|c|c}
       \hline
       \multirow{1}{*}{\diagbox{Variants}{Metrics}}       & \multicolumn{4}{c|}{Video Inversion} & \multicolumn{3}{c}{Video Editing}  \\
        \cline{2-8}
        &MSE$\downarrow$($\times$e-2) &{LPIPS$\downarrow$}   &{WE$\downarrow$}    &{FVD$\downarrow$}  &{TL-ID$\uparrow$} &{TG-ID$\uparrow$} &{FVD$\downarrow$}\\
           \hline
           Latent-T.  &7.455  &0.155  &92.066   &229.066   &0.950     &0.921     &390.77   \\
           \hline
           RIGID    &\textbf{1.277} &\textbf{0.076}   &\textbf{89.068}   &\textbf{177.081}  &\textbf{0.978} &\textbf{0.953}   &\textbf{230.422} \\
           \hline
       \end{tabular}
       }
      \end{center}
      \label{table:large}
      \vspace{-7.5mm}
\end{table*}

\section{Introduction of Evaluation Metrics}

\textbf{MSE and LPIPS.}
For the video inversion task, we use the pixel-wise Mean Square Error (MSE) and Learned Perceptual Image Patch Similarity (LPIPS)~\cite{zhang2018unreasonable} that evaluates the reconstruction quality. Note that the above two metrics evaluate generated faces directly without post processing for avoiding the influence of the original background. 

\textbf{Warp Error.}
For evaluating the temporal stability of inverted videos, we follow Lai~\etal~\cite{lai2018learning} that uses the Warp Error (WE) evaluation. For a pair of consecutive frames $O_{i-1}$ and $O_{i}$, their warp error is computed as:

\begin{equation}
WE\left(O_{t}, O_{t+1}\right)= \sum_{i=1}^{N} M_{t}^{(i)}\left\|O_{t}^{(i)}-\hat{O}_{t+1}^{(i)}\right\|_{2}^{2},
\end{equation}

where $\hat{O}_{t+1}$ is the warped of ${O}_{t+1}$. $M_{t} \in\{0,1\}$ is a non-occlusion mask indicating non-occluded regions, we detect it using~\cite{ruder2016artistic}, $N$ is the pixel number on a frame.
The warp error of an inverted video $O$ is calculated as:
\begin{equation}
WE(O)=\frac{1}{T-1} \sum_{t=1}^{T-1} WE\left(O_{t}, O_{t+1}\right).
\end{equation}

\textbf{Temporally-Local and Temporally-Global Identity Preservation.}
For the edited task, we follow STIT~\cite{tzaban2022stitch} that use Temporally-Local (TL-ID) and Temporally-Global (TG-ID) identity preservation metrics for evaluating the temporal coherence of edited videos. We first employ an off-the-shelf identity detection network~\cite{deng2019arcface} for extracting the identities on all video frames. TL-ID evaluates the identity similarity between pairs of adjacent frames, which evaluates the consistency of a video at the local level.  TG-ID evaluates the similarity between all pairs of videos, it evaluates the consistency of a video globally. 

\textbf{FVD.}
Both edited and inverted videos are evaluated by Fréchet Video Distance (FVD)~\cite{unterthiner2019fvd} metric. Same as its counterpart FID~\cite{heusel2017gans}, FVD also aligns with human perceptual quality. Limited by the length of videos, we sample 100 frames to compute this statistic.

\section{More Ablation Studies}

For analyzing the our input, we present another two variants: 6) we take the use the concatenation between current frame $I_t^a$ and flow $f^a_{t->t-1}$ as input (E($I_t^a$, $f^a_{t->t-1}$)); 7) We only take the current frame $I_t$ as input (E($I_t^a$)). In addition, we also analysis the split between high and low frequency part of $\mathcal{W+}$ space. 8) We use the last 2 layers as high-frequency layers (2 layers); 9) We use the last 6 layers as high-frequency layers (6 layers). In addition, we also evaluate whether our RIGID still work well without seeing any edits in training. In this variant 10), we apply the \textit{in-between frame composition constraint} on the reconstructed frames, but not the edited frames ($w/o$ \texttt{editing}). In addition, we also replaced $\mathcal{L}_{ibfcc}$ with two alternatives: 11) window-averaging on latent codes based on Alaluf~\etal~\cite{alaluf2022third} (\texttt{WA}), and 12)pre-trained frame interpolation work~\cite{xu2019quadratic} for synthesizing intermediate frame $\hat{E}_{t}$ (\texttt{FI}).

The ablation studies can be seen in Tab.~\ref{table:ablation}. We can see that only taking the current frame as input ($E(I_t^a)$) has a slight advantage on pixel-wise reconstruction (MSE), but gets worse performance on temporal smoothness evaluation (WE and FVD). Besides, without taking the edited frames as input, this variant has worse performance on the video editing task (TL-ID and TG-ID). For the variant $E(I_t^a,f^a_{t \Rightarrow t-1})$, it gets a much worse performance, especially on the editing. Without taking the edited frame $E_t$ as input, the correlation in edited frames cannot be modeled successfully. In addition, variant (2 layers) has a slightly better performance on pixel-wise reconstruction, but it performs poorly in temporal consistency evaluation. In contrast, variant (6 layers) has a better performance on the temporal consistency evaluation (WE), but gets inferior results on pixel-wise reconstruction evaluation (MSE and LPIPS). We therefore choose 4 layers for a trade off between temporal consistency and pixel-wise reconstruction evaluation. Variant $w/o$ editing gets a better results on reconstruction, but the edited videos are slightly worse than RIGID, since in this variant, the \textit{in-between frame composition constraint} is applied on the reconstructed frames but not the edited frames. However, as shown in Fig.~\ref{fig:new_edit}, this variant also obtains the coherent edited videos. That is because the obtained latent codes lie in the latent spaces of StyeGAN2, which naturally support various attributes editings. Variant \texttt{WA} shows comparable performance on the WE metric, it exhibited higher MSE and LPIPS scores due to modified code trajectories. In video-based GAN inversion, prioritizing smoothness over reconstruction fidelity is inefficient. Variant \texttt{FI} performed poorly in both inversion and editing tasks. Compared to frame interpolation work, our method directly predicts visible masks, which proves to be an easier task for accurate predictions.

\section{Comparison on a Larger Experiment}

Same as our RIGID, Latent-Transformer~\cite{yao2021latent} is more efficient that can be evaluated in a larger experiment. Here we evaluate our model on 100 videos of the RAVDESS dataset and give the comparison with Latent-Transformer in the following table, we can see our RIGID outperforms Latent-Transformer both on video reconstruction and editing. The comparison can be seen in Tab.~\ref{table:large}. We can see our RIGID outperforms Latent-Transformer both on video reconstruction and editing on this large experiment.


\end{document}